\documentclass{article}

\usepackage[preprint]{neurips_2025}

\usepackage{microtype}
\usepackage{graphicx}
\usepackage{subfigure}
\usepackage{booktabs} 
\usepackage[utf8]{inputenc} 
\usepackage[T1]{fontenc}    
\usepackage{hyperref}       
\usepackage{url}            
\usepackage{booktabs}       
\usepackage{amsfonts}       
\usepackage{nicefrac}       
\usepackage{microtype}      
\usepackage[table]{xcolor}         
\usepackage{enumitem}

\usepackage{graphicx}    
\usepackage{pgffor}      
\usepackage{ifthen}      
\usepackage{comment}

\usepackage{wrapfig}
\usepackage{booktabs}   
\usepackage{colortbl}   

\usepackage{amsmath}
\usepackage{amssymb}
\usepackage{mathtools}
\usepackage{amsthm}

\usepackage[capitalize,noabbrev]{cleveref}

\theoremstyle{plain}

\theoremstyle{definition}

\theoremstyle{remark}

\usepackage[textsize=tiny]{todonotes}

\title{Deep Reinforcement Learning Agents \\ are not even close to Human Intelligence\thanks{In this paper, ``\textit{intelligence}'' specifically refers to the ability to adapt to structural simplifications in relational reasoning tasks, which humans handle with ease but deep RL agents systematically fail to generalize to.}} 

\author{%
  Quentin Delfosse$^{1,2}$\thanks{equal contribution} \\
  \texttt{quentin.delfosse@tu-darmstadt.de} \\
  \And
  Jannis Blüml$^{1,3\dagger}$ \\
  \texttt{jannis.blueml@tu-darmstadt.de}
  \AND
  Fabian Tatai$^{4,5}$ \\
  \And
  Théo Vincent$^{1,6}$ \\
  \And
  Bjarne Gregori$^{1}$ \\
  \And
  Elisabeth Dillies$^{7}$ \\
  \AND
  Jan Peters$^{1,3,4,6}$ \\
  \And
  Constantin Rothkopf~$^{1,3,4,5}$ \\
  \And
  Kristian Kersting$^{1,3,4,6}$ \\
  \AND \\
  $^1$Department of Computer Science, Technical University Darmstadt, Germany \\
  $^2$National Research Center for Applied Cybersecurity (ATHENE), Germany \\
  $^3$Hessian Center for Artificial Intelligence (hessian.AI) \\
  $^4$Centre for Cognitive Science, Darmstadt \\
  $^5$Institute of Psychology, Technical University Darmstadt, Germany \\
  $^6$German Research Center for Artificial Intelligence (DFKI) \\
  $^7$Sorbonne Université, Paris, France \\
}

\newcommand{\jannis}[1]{\textcolor{orange}{Jannis: #1}}

\definecolor{maroon}{cmyk}{0.2, 1, 0, 0.1}

\newcommand{\eg}{\emph{e.g.}~} 

\newcommand{\ie}{\emph{i.e.}~} 

\newcommand{\cf}{\emph{cf.}~}

\newcommand{\ci}[1]{\scriptstyle\ [#1] }

\begin{document}

\maketitle

\begin{abstract}
Deep reinforcement learning (RL) agents achieve impressive results in a wide variety of tasks, but they lack zero-shot adaptation capabilities. 
While most robustness evaluations focus on tasks complexifications, for which human also struggle to maintain performances, no evaluation has been performed on tasks simplifications. 
To tackle this issue, we  introduce HackAtari, a set of task variations of the Arcade Learning Environments. We use it to demonstrate that, contrary to humans, RL agents systematically exhibit huge performance drops on simpler versions of their training tasks, uncovering agents' consistent reliance on shortcuts.\linebreak
Our analysis across multiple algorithms and architectures highlights the persistent gap between RL agents and human behavioral intelligence, underscoring the need for new benchmarks and methodologies that enforce systematic generalization testing beyond static evaluation protocols. 
Training and testing in the same environment is not enough to obtain agents equipped with human-like intelligence.

\end{abstract}

\section{Introduction}

Deep reinforcement learning (RL) has become a key technique for training agents to solve relational reasoning tasks directly from high-dimensional sensory inputs~\citep{zambaldi2018logicrelational}.
In these tasks, agents must identify distinct entities, infer their relationships, and model their dynamics to derive effective decision-making policies.
The Arcade Learning Environment (ALE)~\citep{bellemare2013arcade} is the most widely used benchmark for evaluating RL algorithms in this setting, offering a diverse collection of Atari 2600 games that span a broad range of perceptual and strategic challenges, including spatial reasoning, long-term planning, and real-time reaction.
In their seminal work,\linebreak
~\citet{Mnih2015dqn} introduced the first deep RL method able to solve many ALE games and claimed that when
``\textit{agents are confronted with a difficult task: they must derive efficient representations of the environment from high-dimensional sensory inputs, and use these to generalize past experience to new situations}''. Inspired by the brain’s visual processing mechanisms, they incorporated convolutional neural networks to extract spatial features from raw pixels. 
Deep RL algorithms have then been improved and finally achieved \textit{superhuman performance} in all Atari games~\citep{Badia2020agent57, Badia2020ngu}.
However, the capacity of these agents to ``\textit{generalize experiences from past situations to novel scenarios}'', core to relational reasoning, is rarely assessed by RL practitioners. 
Traditional benchmarks typically involve training and evaluating agents within identical environments, thereby masking their reliance on shortcuts and spurious correlations -- a particularly troubling issue, given recent evidence that RL agents often exploit superficial shortcuts rather than learning robust, causal strategies\linebreak~\citep{ilyas2019adversarial, GeirhosJMZBBW20, ChanFKCG20, koch2021objective, delfosse2024interpretable}.

This reliance on shortcuts has lately been uncovered in the simplest Atari Pong game (depicted in Figure~\ref{fig:changeRAM}). 
In this game, the agent's enemy follows a deterministic behavior based on the position of the ball.
~\citet{delfosse2024interpretable} have thus exposed that deep and symbolic RL agents learn to rely on the enemy's position to catch and return the ball optimally. 
Hiding the enemy or altering its behavior thus leads to a performance drop.
While the inability of DQN agents to zero-shot generalize to complexified versions of their training environment has been showcased~\citep{Farebrother18generalization}, there has not yet been any systematic evaluation of the potential misalignments of RL agents. 
To assess whether RL agents learn the right behaviors, we evaluate them on simplified task variations that humans can easily adapt to - such as color changes or gameplay simplifications - ensuring that performance drops reveal flawed reasoning rather than increased task difficulty.

In this article, we demonstrate that current deep and symbolic RL agents systematically fail to solve simplified variants of their training tasks, supporting that \textbf{human-level performances in training settings does not imply human-like reasoning capabilities}.
We present a systematic investigation of generalization failures in RL using controlled environment modifications, most of which simplify the original tasks. 
We release HackAtari\footnote{HackAtari available at 
\href{https://github.com/k4ntz/HackAtari}{https://github.com/k4ntz/HackAtari}
.}, a suite of environment variations for the most widely used RL benchmark: the Arcade Learning Environments~\citep{bellemare2013arcade}.
We argue that RL agents should be evaluated on held-out task variations -- common practice in machine learning -- to enable meaningful comparisons with human intelligence, particularly in relational reasoning domains.

Our main contributions can be summarized as follows: 
\begin{description}[leftmargin=20pt, itemsep=0pt,parsep=0pt,topsep=-3pt,partopsep=0pt]
    \item[\ \ (i)] We show that deep RL agents fail to generalize to task simplifications and consistently rely on shortcut learning -- selecting the right actions for the wrong reasons, regardless of the algorithm.
    \item[\ (ii)] We show that symbolic/object-centric agents exhibit better adaptation capabilities, but are not yet able to match humans' one.
    \item[(iii)] We conduct a user study to assert users' abilities to maintain their performances within many of these variations, and thus provide performance references for these simplifications.
    \item[(iv)] We open-source a wide range of variations for the Arcade Learning Environments, opening up new possibilities for designing and testing RL agents with supposed  human-like intelligence.
\end{description}

In the following, we present the different RL agents' architectures and algorithms used in this article. We then introduce our HackAtari framework and use it to evaluate RL agents' generalization ability.

\section{The Quest for General RL algorithms}
The objective of relational reasoning RL is to develop general agents with human-comparable problem-solving skills.
This goal encompasses two key subgoals: (1) designing a single, versatile algorithm that can be applied across a wide range of tasks, and (2) enabling trained agents to generalize effectively to variations of their training tasks. 
This work focuses on the second aspect of generality. We here evaluate two prominent classes of agents: deep RL agents and neurosymbolic agents. Task agnostic deep agents often struggle with overfitting and generalization to task variations~\citep{Farebrother18generalization}. 
In contrast, neurosymbolic agents introduce inductive biases by representing environments in terms of objects and their interactions, supporting abstract and transferable representations. 

\textbf{Task agnostic deep RL agents} have been introduced with the Deep Q-Networks (DQN)~\citep{Mnih2015dqn} algorithm, demonstrating that a single convolutional architecture could learn to play a variety of Atari games directly from pixel inputs, establishing a foundation for general-purpose deep RL. 
Subsequent improvements extended DQN along several dimensions: C51~\citep{BellemareDM17} returns distributions to capture uncertainty better. M-DQN~\citep{vieillard2020munchausen} introduces entropy-regularized rewards to improve training stability. Rainbow~\citep{hessel2018rainbow} combines several of these enhancements into a widely used baseline. 
i-DQN~\citep{vincent2025iterated} enables multiple consecutive Bellman updates through a sequence of learned action-value functions.
On-policy methods like PPO~\citep{Schulman2017ProximalPO} and distributed frameworks such as IMPALA~\citep{espeholt2018impala} offer scalable alternatives with strong empirical performance. 
More recently, model-based agents like Dreamer~\citep{Hafner2020dreamer} learn latent dynamics models and perform planning in imagination, yielding greater sample efficiency and improved extrapolation in high-dimensional visual environments. 
The ease of adaptation of deep RL algorithms to novel tasks, requiring no expert knowledge injection, explains their widespread adoption in relational RL~\citep{shaheen2025reinforcement}.

\textbf{Object-centric and neurosymbolic agents} incorporate stronger inductive biases through structured representations but require specific domain adaption. 
Early work on object-oriented MDPs~\citep{diuk2008object} introduced state decompositions based on object attributes and relations, enabling policies to reason abstractly over entities rather than raw pixel arrays. Recent methods build on this idea to improve both interpretability and generalization. 
Unsupervised object extraction methods~\citep{lin2020space,delfosse2021moc} have allowed RL practitioners to develop symbolic policies.
~\citet{JiangL19NLRL} and ~\citet{Delfosse2023InterpretableAE} represents policies as sets of logic rules over symbolic object states, showing transfer to environments with varying object types and counts.~\citet{shindo2024blendrl} leverages large language models to autonomously generate predicates from visual inputs and integrate symbolic reasoning with deep learning in a hybrid policy architecture. Other approaches constrain object-centric policy representations to interpretable structures such as polynomials ~\citep{luo2024insight}, decision trees ~\citep{viper, delfosse2024interpretable, marton2024sympol}, or programs~\citep{cao2022galois, liu2023hierarchical, kohler2024interpretable}. 
These methods are usually also tested in tasks that require extrapolating to unseen object configurations or recombining learned behaviors across novel layouts. 
Finally, object-centric deep agents have been developed. They make use of object-centric masking~\citep{Davidson20, bluml2025deep}, removing background information to force attention on dynamic entities without relying on explicit symbolic conversion. 

\section{Creating ALE variations}
\label{sec:HackAtari}


Our main objective is to show that relational reasoning RL agents consistently learn misaligned policies, \ie that they rely on shortcuts. 
As most of the RL agents trained on such tasks encode their policies within black-box neural networks, we cannot explicit the reason behind their action selections. 
While existing explainable techniques, such as importance maps, help identifying the decisive input zones, they do not explain the core reasoning. 
As outlined by~\citet{delfosse2024interpretable}, deep PPO agents trained on Pong highlight the ball, as well as both paddles. 
This lures external viewers into thinking that the agent attends to all the relevant objects. In contrast, it uses the position of the enemy paddle to estimate the vertical position of the ball. 
The only way to explicitly exhibit these agents reliance on shortcuts is by altering the environment. 
Evaluations on tasks variations have been conducted in the past on more complex environments~\citep{Farebrother18generalization}. 
However, the eventual performance drops do not imply the agent's misalignment, as it could stem from bad adaptation capabilities. 

To assert that relational reasoning agents learn incorrect policies, and have not mastered the desired set of skills necessary to solve the game, we need to test them on tasks simplifications, \ie tasks that will lead humans to \textit{increase or maintain} their overall performances.

We here introduce HackAtari, an extension of the Arcade Learning Environment (ALE), that provides different tasks modifications (mainly simplifications) to evaluate RL agents' \textit{true} relational reasoning abilities. 
Since the source code of the Atari games used in the ALE is proprietary and not publicly available, direct modifications to game logic or mechanics at the code level are not feasible. As a result, the only practical method to alter or simplify these environments for experimental purposes is through direct manipulation of their Random Access Memory (RAM) states.
Thus, modifications are implemented through direct alterations of the game's RAM, allowing diverse controlled perturbations.
Specifically, we created visual alterations, such as changing or obscuring object colors, forcing agents to rely less on superficial visual features. 
Additionally, we adjusted gameplay dynamics, including modifying the speed or presence of objects and enemies, or introducing new mechanics such as gravity effects. 
\begin{wrapfigure}{r}{0.45\linewidth}
\vspace{-3.7mm}
    \centering
    \includegraphics[width=.98\linewidth]{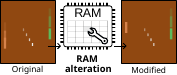}
    \vspace{-1mm}
    \caption{\textbf{RAM alteration allows for modified environments}, here exemplified on Pong. Altering specific RAM cells leads to an enemy remaining static after it returned the ball.}  
    \vspace{-3mm}
    \label{fig:changeRAM}
\end{wrapfigure}
This is exemplified in Figure~\ref{fig:changeRAM} on Pong. 
In the default game, the brown enemy is programmed to go down if the ball is bellow its paddle and up if the ball is above. 
There is thus a high correlation between the enemy's and the ball's vertical position, hence, a misalignment opportunity. 
To create the LazyEnemy variation, we first identify the RAM cell that controls the ball’s horizontal speed. When this value is positive --- indicating that the ball is moving toward the green agent --- we overwrite the enemy’s vertical position with its previous value. This makes the enemy remain static whenever the ball approaches the agent.
We can thus evaluate potential RL agents misalignments.
Let us now provide further examples of tasks variations included in HackAtari, most of which are illustrated in Figure~\ref{fig:multiple-examples}.

\begin{figure}[t]
    \centering    \includegraphics[width=\linewidth]{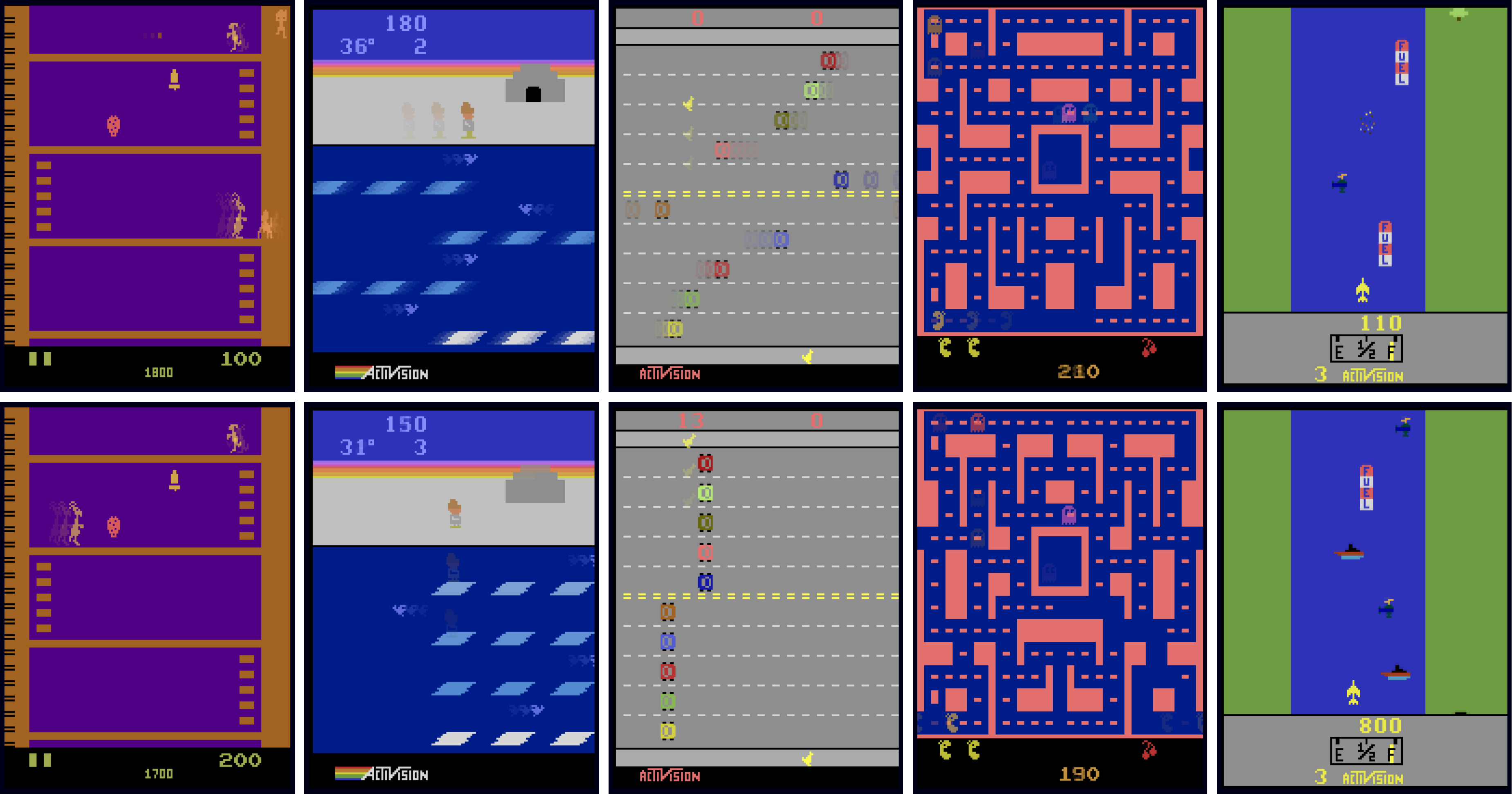}
    \caption{\textbf{Examples of HackAtari simple tasks variations.} Top: the original Atari games used to trained RL agents. Bottom: simplifications (\ie variations for which human performances do not drop). These include color changes and gameplay shifts. Superposed frames show the game dynamics. Descriptions of more environments and their variations are provided in Appendix~\ref{appendix:environments}.}
    \label{fig:multiple-examples}
\end{figure} 

\textbf{NoDanger (Kangaroo).} In Kangaroo, the player controls a mother Kangaroo that needs to reach her joey (top left). Monkeys will come to punch her and throw coconuts.
To collect points, mother Kangaroo can punch the enemies, collect fruits and reach her baby. 
In the NoDanger variations, deadly monkeys and coconuts are deactivated. The player can safely navigate to the joey.

\textbf{StableBlocks (Frostbite).}
In Frostbite, the player builds an igloo by jumping on moving floating ice blocks while avoiding deadly predators. In the modified task, the ice blocks are aligned and static.

\textbf{StoppedCars (Freeway).} In Freeway, the agent's goal is to have the (left) controlled chicken cross the highway. If the chicken collides with a car, it is sent down and immobilized for a few seconds.
In this variation, cars are completely stopped, making crossing trivial.

\textbf{Maze2 (MsPacman).}
In MsPacman, Pacman needs to navigate within a maze to consuming all the dots while avoiding the four pursuing ghosts. In this variation, the maze layout is modified.

\textbf{RestrictedFire (RiverRaid).}
Players here pilot a fighter jet over a river, aiming to destroy enemy targets while avoiding collisions with riverbanks and obstacles. In this variation, the agent can only shoot in case of unavoidable obstacles. It can collects points by dodging and overcoming the enemies.

\textbf{ShiftedShields (SpaceInvaders).}
In SpaceInvaders, we shift the protective shields horizontally by $1$ or by $3$ pixels to the right. This type of change adjusts the environment in a very minimal way, which is often not even noticeable by humans. It does not significantly alter the core gameplay or difficulty.

At the time of writing, HackAtari incorporates more than $224$ variations in total, spanning over more than $33$ games of the original Arcade Learning Environments.
We provide illustrations and descriptions of the variations in Appendix~\ref{appendix:environments}. 
As HackAtari creates tasks variations by altering the RAM, these modify or ablate a specific part of the gameplay. 
Thus, most variations simplify (\ie should lead humans to at least maintain their performances) the original tasks, which is necessary for showing that RL agents learn misaligned behaviors (using shortcuts). 
Further, modifying the RAM values adds a negligible time overhead on top of the Atari emulator, that makes training and testing on the variations as fast as on the original games.
Finally, ALE already incorporates some variations that augment some games' difficulty levels, that are of course integrated in HackAtari.

Overall, these capabilities collectively position HackAtari as a valuable resource for developing relational reasoning RL agents that are robust, adaptable, and capable of generalizing their policies beyond their initial training conditions.

\section{Experimental Evaluation}
\label{sec:evaluation}

Let us now empirically evaluate several aspects of deep and object-centric RL agents' generalization capabilities, focusing specifically on their ability to handle simplified variations of their original training tasks. We here aim to answer the following research questions:

\begin{enumerate}[leftmargin=25pt,itemsep=0pt,parsep=2pt,topsep=1pt,partopsep=2pt]
    \item[\textbf{(Q1)}] {Do RL agents' performance drop on HackAtari tasks variations?}
    \item[\textbf{(Q2)}] {Can human easily adapt to such tasks variations?}
    \item[\textbf{(Q3)}] {Are deep agents systematically learning shortcuts on relational reasoning tasks?}
    \item[\textbf{(Q4)}] {Do human inductive biases help aligning RL agents?}    
    \end{enumerate}

\textbf{Experimental Setup.}
We evaluate a diverse set of RL agents, including standard value-based methods: DQN~\citep{Mnih2015dqn}, C51~\citep{BellemareDM17} and i-DQN~\citep{vincent2025iterated}, policy-gradient methods: PPO~\citep{Schulman2017ProximalPO} and IMPALA~\citep{espeholt2018impala}). 
All agents use the nature-CNN neural network, except for IMPALA, whose authors have come up with a more complex neural network using neural architecture search. 
We also evaluate several object-centric variants: OCCAM~\citep{bluml2025deep}, Object-centric agents using neural networks~\citep{delfosse2024ocatari}, SCoBots~\citep{delfosse2024interpretable}. 
Most tested agents have been collected from their official repository, with only deep PPO and OCNN variants trained in-house using the CleanRL framework~\citep{huang2022cleanrl}. 
Detailed training settings, architectures, and source references for each agent are provided in Appendix~\ref{appendix:agents}.
As commonly done, all agents are trained for $200$ million frames on the Atari environments (v$5$) using a fixed frameskip of $4$ and a repeat action probability of $0.25$, following established evaluation protocols~\citep{machado2018revisiting}. A complete hyperparameters list can be found in the extended experimental setup (\cf Appendix~\ref{appendix:results}). 

To assess generalization, we evaluate each agent HackAtari controlled environment variations, with some described in Section~\ref{sec:HackAtari}.
The evaluation is conducted over $30$ episodes per game per agent, using at least $3$ different evaluation seeds per agent. This resource-limited evaluation suffices to assert the consistent performance drops of the agents (as shown by confidence intervals (CI) in Appendix~\ref{appendix:Q1}).
For metrics, we report Expert-Humans Normalized Score (E-HNS), with expert scores borrowed from~\citet{Badia2020agent57}, and the random scores evaluated in house both on the original and the task variations (\cf Table~\ref{appendix:Q2}), following~\citet{agarwal2021deep}, as well as Performance Drop, aggregated using the inter-quartile Mean (IQM) and $95\%$ CI. 
The Performance Drop is positive if the agent's performance increases, null if the performance does not change, and negative if the performance drops. 
Note that in the computation of these metrics, the average score, obtained by random agents (on the original or on the tasks variation) is ablated.
These metrics enable robust comparisons of the ability of agents to generalize to task variations. For further details on the metrics, \cf Appendix~\ref{appendix:metrics}.\linebreak

\begin{figure}
    \centering
    \includegraphics[width=\linewidth]{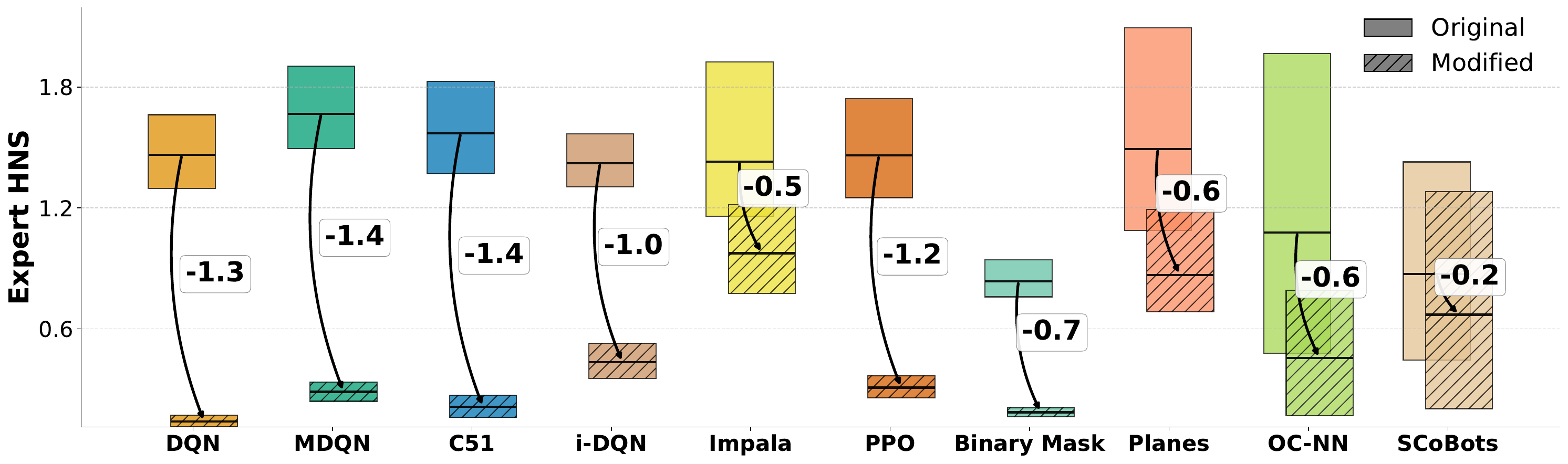}
    \caption{\textbf{Deep and symbolic RL agents performances drop on HackAtari variations}, illustrated by the IQM (following reliable~\citep {agarwal2021deep}) over the human normalized scores (HNS) of various RL agents on a total set of $32$ task variations (over $17$ games). IQMs are computed over $3$ seeded trained agents ($30$ evaluations each). Expert-human scores are borrowed from~\citet{Badia2020agent57}. Performance in the original environment is plotted filled, while the performance in the modified environment is plotted hatched.
    Raw IQM scores (with CIs) for each agent on each game (original and variations) and extended results are provided in Appendix~\ref{appendix:Q1} and \ref{appendix:Q3}.
    }
    \label{fig:all_agents_ehns}
\end{figure}

To evaluate the effect of the game variations on human performance, we conducted an online study with $128$ subjects on $15$ games and a variation for each (on the Prolific platform). 
We included English speaking participants from around the world and all age groups (Mean Age: $33$, Range:[$18$-$73$]). \linebreak
The subjects were allowed to train for $10$ to $15$ minutes on the original game. 
They were then evaluated for $15$ minutes on the original task, and finally for $15$ minutes on a HackAtari variation (for the game and variation list, \cf Appendix~\ref{appendix:Q2}). 
Each subject was only trained and evaluated on a single game and its variation.
Unlike to~\citet{Badia2020agent57}, we have not selected professional gamers, to avoid biased evaluations, potentially stemming from such users' ability to perform well on \textit{any} video game task (thus not directly measuring their ability to adapt to the variations), but there overall zero-shot (or no training) performances.
For more details on the evaluation protocol and demographics studies, \cf Appendix~\ref{appendix:human_study_extended}. 
Let us now demonstrate that current RL agents consistently learn shortcuts, and fail to adapt to task simplifications.

\textbf{RL agents' performance drop on HackAtari most tasks variations (Q1).}
We first evaluate all the available RL algorithms on a subset of $17$ ALE games (the ones with publicly available agents; C51 was only available for $12$ games), with $1$ to $4$ variations per game. 
Figure~\ref{fig:all_agents_ehns} depicts the expert-human normalized scores, averaged using IQMs, for each algorithm, on the $17$ original games and on the $30$ modifications. Every RL agent exhibits a significant performance drop over the complete set of tested variations. 
Remarkably, SCoBots agents exhibit the lowest E-HNS performance drop ($20\%$ overall), and IMPALA maintains an average superhuman performance in the game variations. 
This outlines that changing the architecture or inference paradigm has a bigger impact on the ability of agents to adapt to task variations than varying the algorithm. 
However, we still need to demonstrate that these agents learn misaligned policies that prevent them from adapting to simplifications, and do not fail on the variations because of potential increased game complexities.

\begin{figure}
    \centering
    \includegraphics[width=\linewidth]{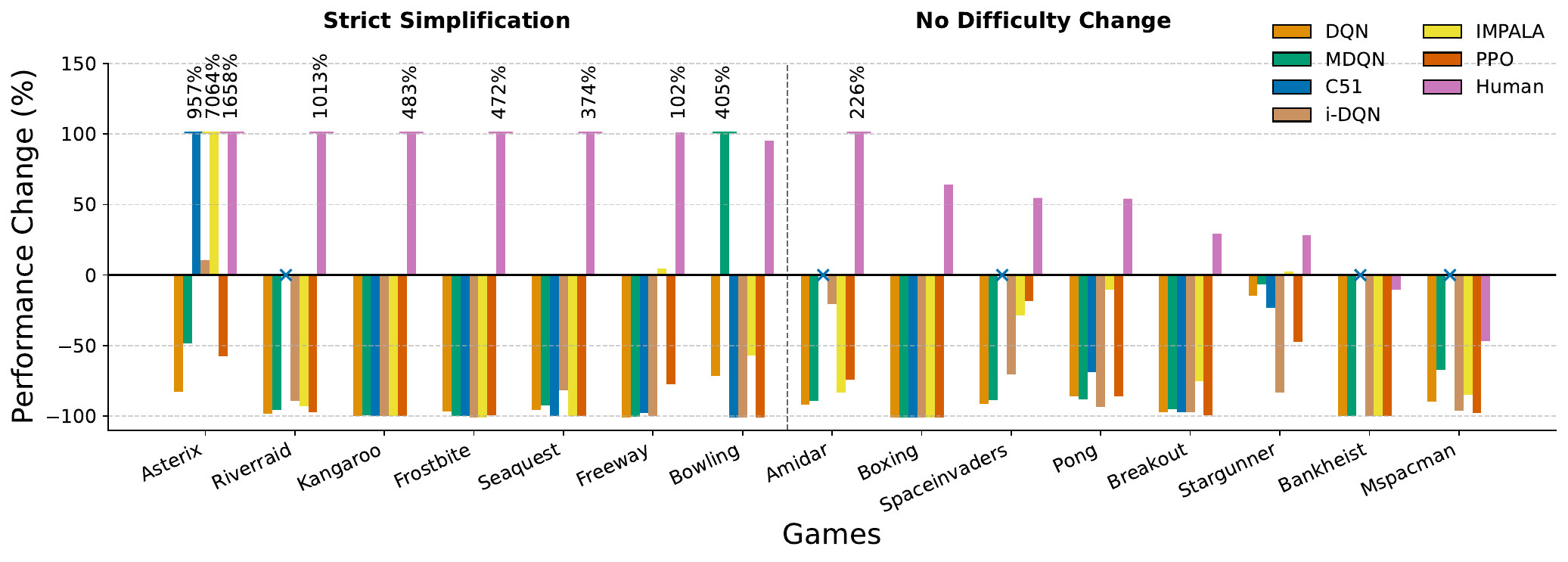}
    \vspace{-6mm}
    \caption{\textbf{While humans easily adapt to task simplifications, deep agents' performances drop,} 
    illustrated on $15$ ALE games. Non-expert users and deep RL agents are trained and evaluated on the original ALE environment, then presented with a variation of the task. Left: Variations considered as task simplifications by design. Right: Variations for which little or no performance increase is expected. Games for which no C51 agent is publicly available are marked with {\color{blue} $\boldsymbol{\times}$}.
    For the exact performances of humans and deep agents, \cf Appendix~\ref{appendix:Q1} and \ref{appendix:Q2}.
    }
    \label{fig:deep_agents_pc}
\end{figure}

\textbf{Human adaptation to simplification far exceeds RL agents' ones (Q2).} 
Our claim that RL agents consistently learn misaligned policies when trained on relational reasoning tasks relies on the fact that, contrary to previous work, we evaluate these agents on tasks \textit{simplifications} (here meant as variations for which human performances do not deteriorate). 
We thus first need to evaluate human users' ability to adapt their learned relational reasoning policies to the HackAtari task variations. 
We thus selected $15$ games, with one variation for each, for which most agents exhibit performance drops while we expect humans to improve or maintain their performances. 
We randomly selected $134$ users (at least $8$ per game), trained on the original ALE games, then evaluated them on the original version and on the variations. 
We evaluate the performance changes of each agent type on each game. Note that the performance change depicts the change in variation (between the original ALE task and the HackAtari selected variation) of each agent. This metric does not allow for comparing the performances of the different agents, neither on the original task, nor on the variation, but measures individual performance variations. 
Figure~\ref{fig:deep_agents_pc} shows that humans performances drastically increase on $11$ games, notably increases on $2$ games, slightly decreases on Bankheist, and notably decreases on MsPacman.\linebreak 
The modifications used for Bankheist spawn police cars that constantly chase the player, but increase the reward obtained by successfully robbing a bank. 
Depending on the ability of humans to escape the chasing police, this can lead to either a performance increase or a drop (\cf Appendix~\ref{appendix:human_results_figures}). \linebreak
For MsPacman, humans exhibit performance drops on the maze layout changes. This is likely due to some fatigue faced by the participants (MsPacman is one of the most complex and stressful games), as most users' performance already slightly decreases between the training phase and the first evaluation. This could also stem from users' overfitting strategies. 
They could be learning to favor a secure path that is not transferable between the maze layout changes. 
Further investigation is needed to determine how much each feature impacts the performance changes. 
However, users' performance drop is still significantly lower than all the available deep agents ones on MsPacman.
Overall, the descriptions of the applied change provided in Section~\ref{sec:HackAtari} and Appendix~\ref{appendix:environments} and the user study theoretically and experimentally support that these $15$ tasks variations are simplifications. 

\textbf{Deep agents fail to adapt to task simplifications (Q3).} 
Let us now investigate whether deep agents can adapt to task simplifications. 
We evaluate several widely-used algorithms, including DQN, PPO, C51 (not available on all games), IMPALA, i-DQN, and MQN, across many original Atari environments and some simplified variations included in HackAtari.
Figure~\ref{fig:deep_agents_pc} shows that, contrary to humans, all agents exhibit severe performance drops in most of the task variations. 
Even IMPALA's performances drop by more than $50\%$ on $10$ out of $15$ games.
All other algorithms exhibit even higher performance drops (also confirmed by Figure~\ref{fig:all_agents_ehns}). 
Deep RL agents can thus not maintain their performances on most task simplifications, demonstrating their inability to learn meaningful, correctly aligned policies. 

\begin{figure}[t]
    \centering
    \includegraphics[width=\linewidth]{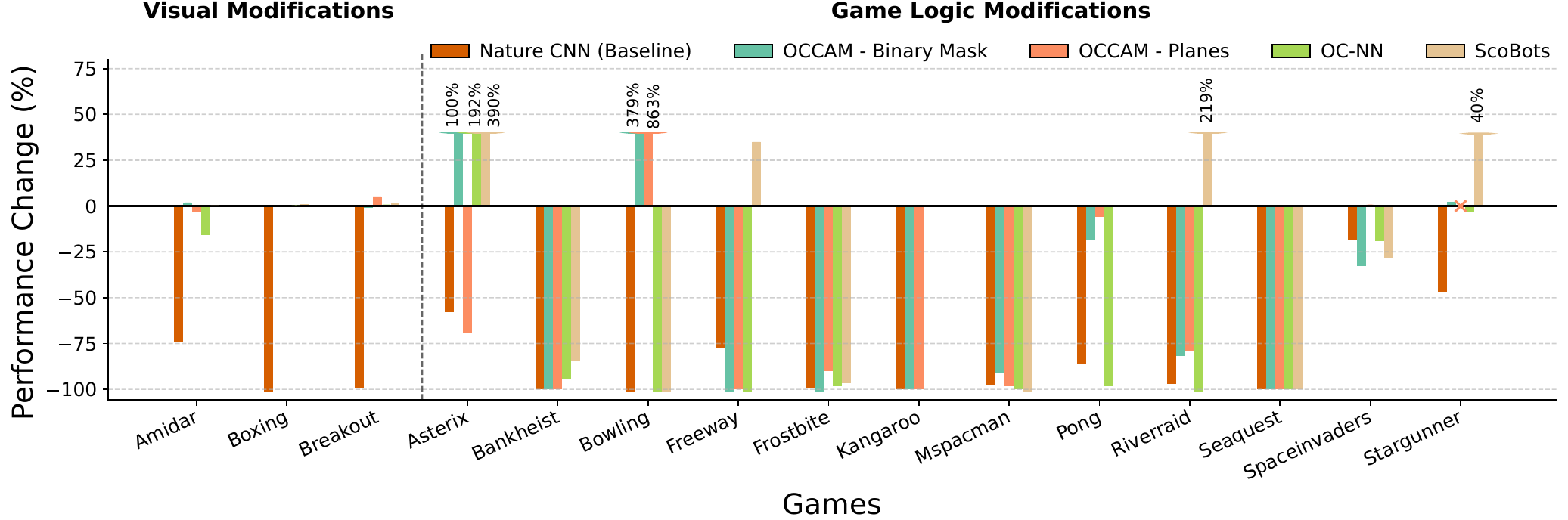}
    \caption{\textbf{Object-centric RL agents also fail to adapt to simplified environments.} 
    Different object-centric approaches (all using PPO) are here compared to the classical CNN baseline on the same $15$ variations (as Figure~\ref{fig:deep_agents_pc}). Visual perturbations (\eg color change, left) have a very limited impact on the symbolic agents, while most gameplay modifications (left) still cannot be solved by these agents. 
    Extended results are available in Appendix~\ref{appendix:Q3}.}
    \label{fig:oc_agents_pc}
\end{figure}

\textbf{The object-centric inductive bias is not enough (Q4).} RL agents relying on more human inductive biases could help narrow the gap between deep agents and humans. 
The most common bias is object-centricity, with agents that first extract the depicted objects and their properties (such as position, color, size) from the pixel states. 
Such representations have been shown to improve transferability and interpretability in structured environments~\citep{shindo2024blendrl}.
We thus evaluate whether introducing object-centric representations enhance the agents' ability to generalize to simplifications. 
Figure~\ref{fig:oc_agents_pc} shows Performance Change for object-centric agents. 
As these agents are all trained using PPO, we also provide also include deep (\ie using Nature-CNN) PPO baseline. 
As expected, these agents are insensitive to visual perturbations (\eg color changes). 
Notably, decision tree based SCoBots agents have limited performance drops (<$30$\%), on $9$ out of $15$ games.
However, even on most tasks with gameplay alterations, all object-centric agents exhibit high performance drops. 
Overall, even if these symbolic architectures generally adapt better than the deep pixel-based PPO baseline, they still exhibit notable performance drops. These findings suggest that while object-centricity helps, it alone is insufficient for robust generalization to simpler tasks.

Overall, our analysis demonstrates the global incapacity of both deep and symbolic agents to adapt to task simplifications, as they often completely overfit to their training tasks. \textbf{RL agents do not} ``\textit{derive efficient representations allowing to generalize past experience to new situations}''~\citep{Mnih2015dqn}.

\newpage
\section{Related Work}

\paragraph{Generalization Benchmarks and Failures.}
As shown here, generalization remains a core unsolved problem in reinforcement learning. Traditional evaluations often focus on performance in fixed environments, leading to overfitting and a poor understanding of agents' robustness. Extensions to the Arcade Learning Environment (ALE) such as sticky actions~\citep{machado2018revisiting}, unseen modes~\citep{cobbe2019quantifying} and action noise~\citep{koch2021objective} introduced limited variability to probe robustness.~\citet{zhang2018natural} also focus on perturbations on the observation space of the ALE by evaluating trained agents against abrupt background changes.
More ambitious frameworks like CoinRun~\citep{cobbe2019quantifying} and Procgen~\citep{cobbe2020procgen} procedurally generate diverse levels, but this does not prevent misalignment~\cite{Langosco2022goal}. 
Crafter~\citep{hafner2022benchmarking} and MineRL~\citep{guss2019minerl, milani2020retrospective} further emphasize compositionality and long-term credit assignment. 
However, these benchmarks increase perceptual or structural complexity, making it difficult to disentangle poor adaptation from underlying misalignment. 
In contrast, our work mainly introduces \emph{simplifications} that preserve task semantics while reducing difficulty, revealing performance failures even in settings where humans readily adapt.

\paragraph{Misalignment and Shortcut Learning.}
We show that RL agents often succeed by exploiting spurious correlations or shallow heuristics, rather than learning task-aligned policies. 
This phenomenon, known as shortcut learning~\citep{GeirhosJMZBBW20}, has been documented in supervised vision~\citep{ilyas2019adversarial,stammer2021right} and increasingly in RL~\citep{zhang2018natural, cobbe2020procgen,koch2021objective,delfosse2024interpretable}. 
Recent interpretability efforts, such as saliency maps and attention overlays, suggest that agents attend to task-relevant regions~\citep{Greydanus2018-atariheatmaps}, yet fail to capture the latent reasoning process. 
In Pong, for instance, agents may appear to track the ball but instead use opponent behavior as a proxy~\citep{delfosse2024interpretable}.
While robustness techniques such as adversarial regularization~\citep{pinto2017robust} and domain randomization~\citep{tobin2017domain} offer partial defenses, they do not prevent agents from failing when irrelevant cues are removed. 
By evaluating agents on simplified tasks, where human performance is stable or improves, we isolate misalignment not as failure to adapt, but as failure to \emph{learn to select actions for the right reasons}.

\section{Towards RL policies aligned with the true task goals} 

\textbf{Evaluating agents' robustness through simplifications to better reflect human-like reasoning.}
Robust reinforcement learning is often framed as worst-case optimization~\citep{pmlr-v32-tamar14,moos2022robust} or adversarial training~\citep{pinto2017robust}, with recent work addressing observation perturbations through network sparsity~\citep{grooten2023automatic} or Lipschitz constraints~\citep{barbara2024robust}. However, RL agents often fail on simplified versions of their training tasks -- settings where humans adapt effortlessly. 
This reveals a reliance on superficial cues rather than robust relational reasoning. 
We argue that testing agents on logically simpler variations is essential for evaluating true generalization. 
HackAtari allows for testing whether agents truly understand the task, challenging the common assumption --- reflected in metrics like the Human Normalized Score --- that achieving human-level performance implies human-like reasoning.
We advocate for benchmarks that explicitly test relational understanding, as the ones appearing in supervised learning setting~\citep{helff2025v, wust2025bongard} and for integrating human inductive biases to align RL agents with task structure.

\textbf{Incorporating human inductive bias is crucial for the rise of aligned RL agents.}
Our work demonstrates that relying on extensive compute, large number of parameters and long training does not guarantee that deep RL agents learn coherent representations of the tasks, aligned with the intended task goals.
While the dopaminergic system in humans partly resembles model-free reinforcement learning~\citep{schultz1997neural}, 
humans make use of use rich model-based representations of their environment~\citep{tenenbaum2011abstraction, daw2011model}.
Several studies provide guidance how these findings from cognitive science about human intelligence should be incorporated in intelligent agents~\citep{lake2017building, zhu2020dark}.
Besides using structured representation and planning in navigation~\citep{kessler2024human} humans have also been shown to have coherent but noisy models of their physical environment~\citep{kubricht2017intuitive, battaglia2013simulation}, 
aiding with inference over object properties~\citep{neupartl2020intuitive}, learning tool use ~\citep{allen2020rapid}, and choosing actions ~\citep{tatai2025intuitive}.
Further, humans also decompose complex tasks into a sequence of high-level actions, and learn skills that correspond to the different sub-goals of such tasks. 
Hierarchical RL has been shown to improve adaptation to task variations by learning reusable sub-policies or skill hierarchies~\citep{BaconHP17, hausman2018learning, li2020sub, cannon2025accelerating}.
Humans also model causal object interaction~\citep{michotte1963perception} and can employ abstract and qualitative reasoning about physical situations~\citep{forbus1988qualitative}. 
Latest approaches try to incorporate causal world models within RL agents~\citep{yang2024towards, dillies2025better} applicable to multi-objective tasks~\citep{bhatija2025multi}.
Additionally, humans also represent the beliefs and desires of other agents~\citep{dennett1989intentional}, mathematically formalizable as inverse reinforcement learning (IRL)~\citep{baker2009action}.
IRL can infer uncertainties about others' behaviors and the consequential reward signals~\citep{dimitrakakis2011bayesian, straub2023probabilistic}, relevant in \eg Kangaroo, where agents prioritize enemy-killing over child-rescue, showcasing misalignment between RL agents’ and human goals. Incorporating the common reasoning abilities of language models to extract reward signals from object-centric states could help tackle this problem~\citep{kaufmann2024ocalm}.

\section{Limitations and Future Work}
While HackAtari offers valuable insights into RL generalization, it is limited by its reliance on ALE, which restricts evaluation to visually stylized environments that may not transfer to more complex or realistic domains. 
As the benchmark gains popularity, there is also a risk that the agents will overfit to its specific perturbations, undermining its diagnostic value, even if we believe that with enough variations for a given tasks, agents able to solve all variations while learning on the original tasks are likely to be aligned with the primary desired task objectives. 
Finally, the absence of standardized difficulty scaling across variations can make it difficult to interpret the significance of observed performance drops.
We hope that overall, this analysis and the release of the HackAtari variations will motivate RL practitioners to develop agents can maintain performance on task simplifications, thus learning correctly aligned policies.\linebreak
We believe that systematic evaluations should also be run on held-out test sets also for other, potentially more complex RL benchmarks.
To obtain correctly aligned agents, IMPALA stands out for maintaining superhuman performance under task variations, likely due to its off-policy V-trace correction, distributed data collection, and smoother policy updates. 
Understanding how these elements support generalization may guide the development of more task-aligned deep RL agents. \linebreak
However, we believe that object-centricity, and in general the inclusion of human inductive biases, represent a better avenue towards RL agents that learn to select the right action for the right reasons.

\section{Conclusion}

While~\citet{Mnih2015dqn} emphasized that ``\textit{reinforcement learning agents must learn efficient representations from high-dimensional sensory inputs and generalize from past experience to new situations}'', our results demonstrate that current deep and symbolic RL agents largely fail to meet this objective. To investigate this issue systematically, we introduce HackAtari, an open-source benchmark comprising over 200 task variations based on the Arcade Learning Environment, the most widely used evaluation suite in deep RL. HackAtari is designed to go beyond in-distribution evaluation by allowing researchers to test agents on slight but targeted modifications of familiar tasks. These variations, such as changes in color schemes or simplified game dynamics, are typically trivial for humans to adapt to but often reveal the brittleness of learned policies. We hope that HackAtari will serve as a diagnostic tool for evaluating whether RL agents have truly internalized the relational structure of their environments and whether they select actions for the right, generalizable reasons.\linebreak
Finally, ~\citet{silver2021reward} have hypothesized that \textit{reward is enough}, but our results suggest that without the right biases and evaluations, reward alone is not enough to build agents that truly understand.\linebreak

\section{Impact statement}
Our work aims at developing correctly aligned RL agents, who base their action selection on the correct intended goal. We believe that such algorithms are critical to uncover and mitigate potential misalignments of AI systems. 
A malicious user can, however, utilize such approaches for aligning agents in a harmful way, thereby potentially leading to a negative impact on further users or society as a whole. 
Even so, we believe that the checking that RL agents follow the intended goal could greatly help preventing the deployment of agents whose behavior could unexpectedly diverge from the one intended by their creators, thus reducing potential harm caused by such agents.

\section*{Acknowledgments}
We would like to express our sincere gratitude to our students for their dedication and feedback as well as all people participating in our study.

This research work has been primarily funded by the German Federal Ministry of Education and Research, the Hessian Ministry of Higher Education, Research, Science and the Arts (HMWK) within their joint support of the National Research Center
for Applied Cybersecurity ATHENE as well as the Excellence Program of the Hessian Ministry of Higher Education through their cluster projects within the Hessian Center for AI (hessian.AI) ``The Third Wave of Artificial Intelligence - 3AI'' and ``The Adaptive Mind''. Further, we acknowledge support and funding by the German Research Center for AI (DFKI).

\bibliography{bib}
\bibliographystyle{icml2025}

\newpage
\appendix
\onecolumn


\section*{Appendix Overview}
\vspace{-0.5em}
\begin{itemize}
    \item[\textbullet] \textbf{Appendix~\ref{appendix:metrics}:} \textit{Metrics}\\
    Definitions of Human-Normalized Score (HNS), Interquartile Mean (IQM), and performance change metrics.

    \item[\textbullet] \textbf{Appendix~\ref{appendix:agents}:} \textit{Agent Architectures and Training Setup}\\
    Describes all agents used in the study, their implementation details, training protocols, and sources.

    \item[\textbullet] \textbf{Appendix~\ref{appendix:compute}:} \textit{Computational Resources}\\
    Details hardware setup, GPU usage, training time, and total inference budget required for evaluation.

    \item[\textbullet] \textbf{Appendix~\ref{appendix:results}:} \textit{Evaluation Setup and Extended Results}\\
    Provides extended results and tables for deep agents, object-centric models, and human baselines.

    \item[\textbullet] \textbf{Appendix~\ref{appendix:hackatari_extended}:} \textit{Code and Data}\\
    Describes access to HackAtari environments, RAM modifications, and data logging pipeline.

    \item[\textbullet] \textbf{Appendix~\ref{appendix:human_study_extended}:} \textit{Human Study Details}\\
    Includes full study protocol, consent materials, payment info, interface screenshots, and questionnaire.

    \item[\textbullet] \textbf{Appendix~\ref{appendix:environments}:} \textit{HackAtari Game Descriptions and Variants}\\
    Full documentation of game environments and all implemented task simplifications used for evaluation.

\end{itemize}
\vspace{1em}

\newpage

\newpage

\section{Metrics}
\label{appendix:metrics}
Throughout this article, we have used two complementary metrics to evaluate agents in Atari environments: the Interquartile Mean (IQM) over \textit{Human-Normalized Scores (HNS)}, which measures absolute performance of an agent, normalized by the performance of a human. Metrics are aggregated using the \texttt{rliable} library~\citep{agarwal2021deep} to provide statistically reliable evaluations across diverse environments.

\subsection*{(Expert) Human-Normalized Score (HNS)}

To assess agent performance relative to both naive and expert baselines, we report the Human-Normalized Score (HNS), a standard evaluation metric in Atari benchmarks~\citep{Mnih2015dqn, machado2018revisiting}. Let $A$ denote the average score achieved by the agent, $H$ the score of a human expert, and $R$ the score of a random policy. The HNS is defined as:

\begin{equation}
\text{HNS} = \frac{A - R}{|H - R|}
\end{equation}

An HNS of 1.0 (or 100\%) indicates human-level performance, while values above 1.0 suggest superhuman ability. Values near 0 imply the agent performs comparably to a random policy, and negative values indicate performance worse than random. This metric normalizes across games with different score scales and dynamics, enabling fair comparison and aggregation across tasks. HNS is computed using raw episodic returns averaged over multiple seeds. We are using the human and random reference scores from~\citet{Badia2020agent57}. Since~\citet{Badia2020agent57} were using ``\textit{professional human game testers}'', we call this Expert HNS. We report HNS alongside other metrics such as interquartile mean (IQM) and performance change to provide a robust and nuanced picture of agent behavior across original and modified environments.

\subsection*{IQM over Multiple Games and Modifications}

To evaluate the robustness of each model across multiple environments and their corresponding modifications, we compute the Expert HNS over all available games, \cf \autoref{fig:all_agents_ehns}. For modified environments, each game's (multiple) variants are first aggregated by computing the mean of HNS across modifications, ensuring that games with more variants do not exert disproportionate influence on the result. To aggregate the games, we follow~\citet{agarwal2021deep}, using \texttt{rliable}. This means we calculate the IQM over all results of all seeds for all games.

\subsection*{Using the Interquantile Mean (IQM) + 95\% Confidence Intervall (CI) over Mean + StdErr.}
Following~\citet{agarwal2021deep}, we report the interquartile mean (IQM) along with 95\% stratified bootstrap confidence intervals (CIs) as our primary performance metric. Return distributions in Atari are frequently highly variable, skewed, and heavy-tailed, often due to a small number of runs achieving unusually high scores. These outliers can inflate the sample mean, resulting in a metric that overstates the agent’s typical performance. Additionally, pairing the mean with standard error (SE) implicitly assumes independent, identically distributed (i.i.d.) samples from a light-tailed, symmetric distribution, assumptions that are frequently violated in RL settings.

The IQM addresses these issues by computing the mean over the interquartile range (25th to 75th percentiles), discarding both the lowest and highest quartiles. This yields a robust point estimate that emphasizes consistent behavior across seeds and reduces the influence of extreme values. To quantify uncertainty, we use non-parametric bootstrap resampling to construct 95\% confidence intervals, which make no distributional assumptions and better reflect the empirical variability of the data.

This combination—IQM with bootstrap CIs—provides a statistically sound and robust summary of performance, particularly well-suited for environments like Atari where outcomes can vary substantially across random seeds. It enables more reliable comparisons between agents and guards against misleading conclusions driven by a few atypical trajectories.

\subsection*{Aggregating Human Study Results Using Mean with Per-Participant IQMs}

To evaluate human performance in a robust and statistically sound manner, we report the mean over per-participant IQM as our primary summary statistic. Each participant completed multiple episodes per evaluation condition (e.g., original vs. modified environment), resulting in a distribution of scores that may be highly variable due to factors such as learning effects, momentary lapses in attention, or individual familiarity with similar tasks.

To reduce the impact of intra-participant variance and episodic outliers, we first compute the IQM across each participant’s episodes. This yields a robust estimate of central performance for each individual, effectively filtering out unusually high or low scores that may not be representative of typical behavior.

After obtaining one IQM per participant per condition, we compute the mean across participants to summarize overall group-level performance. The goal of this is to obtain a more stable and representative estimate of average human performance, where all humans share the same influence on the results, independent of their performance or how many episodes they play. 

\subsection*{Performance Change}
To quantify the impact of environment modifications on agent performance, we compute the performance change (PC) as the relative degradation in performance between the original and modified environments. For a given metric $M$ (e.g., HNS or raw return), let $M_{\text{original}}$ denote the score in the original environment and $M_{\text{modif}}$ the corresponding score in the modified one. In this work, we used the raw scores for the computations; as such, we subtract the random scores from each metric to make sure. The performance drop is defined as:

\begin{equation}
\text{PC} = \frac{(M_{\text{modif}}-R_\text{modif})-(M_{\text{original}}-R_\text{original})}{|M_{\text{original}}-R_\text{original}|}
\end{equation}

A value of 0 indicates no performance change, while values closer to 100\% represent a user reaching twice the amount of points. A value close to -100\% means the agent performs similar or worse than a random agent. This normalized form facilitates comparison across tasks with different score ranges and highlights robustness failures that may be masked by absolute performance metrics.

\clearpage
\section{Agents}
\label{appendix:agents}

We evaluate a diverse set of reinforcement learning (RL) agents, covering both standard baselines and object-centric architectures. Most agents are based on publicly available implementations or pretrained models, while \textbf{PPO} was trained by us specifically for this study. Below, we summarize the agents used, their foundational works, and where to find their implementations or pretrained models.

\begin{itemize}
\item \textbf{DQN, MDQN} – Standard value-based deep RL models. DQN is the canonical deep Q-learning agent introduced by~\citet{Mnih2015dqn}, while MDQN is its Munchausen variant~\citep{vieillard2020munchausen} that incorporates entropy regularization into Q-learning. We use the pretrained models from~\citet{gogianu2022agents}\footnote{\url{https://github.com/floringogianu/atari-agents}}, originally built for analyzing agent robustness.

\item \textbf{C51} – A distributional RL method that models return distributions instead of expected values~\citep{BellemareDM17}. C51 models were also taken from~\citet{gogianu2022agents}, though pretrained models are not available for all games included in our benchmark.

\item \textbf{i-DQN} –  Another strong and relatively new baseline, based on the idea of iterated Q-Networks (i-QN)~\citep{vincent2025iterated}, enabling multiple consecutive Bellman updates by learning a tailored sequence of action-value functions where each serves as the target for the next one. We used the models by~\citet{vincent2025iterated}, which can be found on Huggingface\footnote{\url{https://huggingface.co/TheoVincent/Atari_i-QN}}.

\item \textbf{IMPALA} – A scalable actor-critic architecture designed for distributed training, introduced by~\citet{espeholt2018impala}. We evaluate pretrained models by the cleanRL team, available on Huggingface\footnote{\url{https://huggingface.co/cleanrl}} .

\item \textbf{PPO} – A widely used policy-gradient baseline introduced by~\citet{Schulman2017ProximalPO}. We trained our own PPO models using the CleanRL framework~\citep{huang2022cleanrl}, following their recommended hyperparameters and reproducibility practices. Training details and hyperparameters are provided in Table~\ref{tab:hyperparams}. Models will be made available after acceptance.

\item \textbf{Binary Mask PPO, Planes PPO} – Object-centric PPO variants introduced by~\citet{bluml2025deep}, incorporating structured visual representations (e.g., binary object masks, spatial planes) instead of raw pixels. These models were trained by us using modified observation wrappers. Code and trained models can be found on GitHub\footnote{\url{https://github.com/VanillaWhey/OCAtariWrappers}} or were provided by the authors. Note, these models were trained only for 40 million frames instead of 200 million. 

\item \textbf{Semantic Vector PPO, ScoBots Vector} – Agents that operate on symbolic or vector-based representations of objects in the environment, pretrained using the OCAtari and ScoBots frameworks~\citep{delfosse2024ocatari, delfosse2024interpretable}. Both Semantic Vector PPO and ScoBots Vector were provided by the authors. Final model checkpoints will be made available after acceptance.

\end{itemize}

\clearpage
\subsection{Training PPO Agents}
\label{appendix:training}

We train PPO agents for 200 million environment frames using the \texttt{v5} version of the ALE, following the evaluation protocol established by~\citet{machado2018revisiting}. A fixed frameskip of 4 and a repeat action probability of 0.25 introduce environment stochasticity and align with established Atari benchmarks. Observations are preprocessed by converting RGB frames to grayscale, resizing to $84 \times 84$ pixels, and stacking the last four frames to capture short-term temporal dynamics. Each agent is trained using 3 random seeds to ensure robustness across initialization.

Our PPO policy uses an IMPALA-style convolutional encoder~\citep{espeholt2018impala} with separate heads for the policy and value function. Training is conducted with CleanRL~\citep{huang2022cleanrl}, with full training settings provided in Table~\ref{tab:hyperparams}. The agent is trained to maximize the sum of undiscounted episodic returns.

In addition to pixel-based PPO, we train a Semantic Vector agent using object-centric observations derived from OCAtari~\citep{delfosse2024ocatari}. Instead of raw images, this agent receives a structured vector representation extracted from two consecutive frames. Each observation encodes object-level information, including positions, bounding boxes, and class labels for entities present in the scene. These vector inputs are passed through a multilayer perceptron encoder before being processed by a PPO policy. Training follows the same protocol as our pixel-based PPO agents, using CleanRL with adjusted hyperparameters for vector inputs.

This setup enables direct comparison between pixel-level and object-centric agents under identical training conditions and stochastic environment settings, allowing us to assess how input structure affects generalization and robustness.

\begin{table}[h]
\centering
\setlength{\tabcolsep}{3pt}
\renewcommand{\arraystretch}{1.1}
\caption{Hyperparameters used in our PPO training to ensure reproducibility and consistency.}
\label{tab:hyperparams}
\begin{tabular}{|l|c||l|c|}
\toprule
\textbf{Hyperparameter} & \textbf{Value} & \textbf{Hyperparameter} & \textbf{Value} \\
\midrule
Seeds & \{0,1,2\} & Learning Rate ($\alpha$) & $2.5 \times 10^{-4}$ \\
Total Timesteps & $2 \times 10^8$ & Number of Environments & 10 \\
Batch Size ($B$) & 1280 & Minibatch Size ($b$) & 320 \\
Update Epochs & 4 & GAE Lambda ($\lambda$) & 0.95 \\
Discount Factor ($\gamma$) & 0.99 & Value Loss Coefficient ($c_v$) & 0.5 \\
Entropy Coefficient ($c_e$) & 0.01 & Clipping Coefficient ($\epsilon$) & 0.1 \\
Clip Value Loss & \texttt{True} & Max Gradient Norm ($\|g\|_{\text{max}}$) & 0.5 \\
\bottomrule
\end{tabular}
\end{table}

\clearpage

\section{Computational Resources}
\label{appendix:compute}

Although many of the agents used in this study were obtained from publicly available repositories and did not require retraining (cf. Appendix~\ref{appendix:agents}), substantial computing resources were required to conduct large-scale evaluations across our proposed HackAtari benchmark.

\paragraph{Evaluation Scale.} Each of our $11-12$ agents was evaluated on $17$ games with $1-4$ simplified or modified variations per game (in total 50 game configurations), across $3$ random seeds and $10$ episodes per configuration. This resulted in around $1,500$ gameplay rollouts per agent. While these are not expensive, the amount still leads to non-trivial inference costs. The total evaluation workload included both deep RL baselines (e.g., DQN, PPO, IMPALA, C51, i-DQN,...) and object-centric agents (e.g., binary masks, planes, OC-NN, ScoBots,...), for a total of $12$ unique agent configurations. This results in around $18,000$ evaluation runs. The results can be seen in Appendix~\ref{appendix:results}. Evaluation used deterministic seeds and fixed emulator settings (e.g., sticky actions) to ensure reproducibility (also \cf Appendix~\ref{appendix:setup}).

\paragraph{Training Resources.} A subset of agents, namely, PPO (pixel-based) were trained in-house using the CleanRL framework (cf. Appendix~\ref{appendix:training}). Most PPO agents took around $4-6$h per instance In total, we trained around $120$ PPO models. Total training time for these agents amounted to approximately {$500-700$ GPU hours.

\paragraph{Hardware.} All training was conducted on a cluster of NVIDIA DGX systems, each equipped with A100 GPUs and sufficient RAM (greater than 40 GB per process) to support parallelized Atari rollouts, see Table~\ref{tab:hardware}. All evaluations, except IMPALA, were conducted using Python 3.11, Gym 0.26.2, and the ALE-py Atari interface (v0.8.1), on macOS 15.3.1. IMPALA, due to their requirement of the envpool package, was evaluated on a separate device, using Python 3.9 and Arch 2025.02.01.

\begin{table*}[tbh!]
    \centering
    \begin{tabular}{lr} \toprule
        \textbf{Hardware/Software} & \textbf{Description}  \\ \midrule
        GPU & 8 $\times$ NVIDIA® Tesla A100 \\
        NGC Container & \url{nvcr.io/nvidia/pytorch:23.05-py3} \\
        GPU-Driver & CUDA 12.2 \\
        CPU & Dual Intel Xeon Platinum 8168 \\
        Operating System & Ubuntu 23.02 LTS \\
        \bottomrule
    \end{tabular}
    \caption{Hard- and software configuration for our experimental section. }
    \label{tab:hardware}
\end{table*}

\clearpage

\section{Extended Results}
\label{appendix:results}

\subsection*{Evaluation Setup}
\label{appendix:setup}

Our evaluation benchmarks RL agents across \textit{17} Atari environments, each tested under $1$ to $4$ environment modifications, depending on the environment. All agents are trained, not always by ourselves, exclusively on the original, unmodified versions of the games using $200$ million environment frames and the hyperparameters listed in Table~\ref{tab:hyperparams}, unless otherwise specified in the figure or table captions. Evaluation is performed using our own evaluation setup, based on HackAtari.

Each experiment is conducted over $3$ seeds ($0,1,2$) with $10$ episodes per game per seed, totaling $30$ episodes per configuration. We follow the evaluation guidelines by \citet{machado2018revisiting}, e.g., using a repeat action probability of $0.25$ and a maximum of 30 NOOP actions after each reset, meaning initial states are sampled by taking a random number of no-ops on reset. No-op is assumed to be action 0.

The evaluated agent architectures and training settings are described in detail in Appendix~\ref{appendix:agents}. For performance metrics, we report raw episodic returns, interquartile mean (IQM), and performance change, as defined in Section~\ref{appendix:metrics}. The selected environments and the corresponding modifications applied to each are introduced in Section~\ref{appendix:environments}.

\begin{table}[h]
\centering
\caption{Evaluation Parameters}
\setlength{\tabcolsep}{3pt}  
\renewcommand{\arraystretch}{1.1}  

\begin{tabular}{|l|c||l|c|}
\midrule
\textbf{Hyperparameter} & \textbf{Value} & \textbf{Hyperparameter} & \textbf{Value} \\
\midrule
\#Episodes & 30 & 
\#Seeds & 3 \\
Epsilon & 0.001 &
Frameskip & 4 \\
Frames stacked & 4 &
Repeat Action Probability & 0.25 \\
Full Actionspace & False &
Max. NOOP Actions after Reset & 30 \\
\midrule
\end{tabular}
\label{tab:hyperparams2}
\end{table}

\clearpage
\subsection*{Mapping Results to Figure}

\begin{table}[h]
\centering
\caption{Mapping of HackAtari game variations to figures in the main paper. Each entry indicates which figure(s) include evaluation results for the given game and modification pair. Games in {\color{maroon!60}pink} are used in the human study (cf. Appendix~\ref{appendix:human_study_extended}).}
\label{tab:figure_usage_map}
\begin{tabular}{lll}
\toprule
\textbf{Game} & \textbf{Variant} & \textbf{Appears In} \\
\midrule
\rowcolor{maroon!30} Amidar & paint roller player & Fig. 3, 4, 5 \\
Amidar & pig enemies & Fig. 3 \\
\rowcolor{maroon!30} Asterix & obelix & Fig. 3, 4, 5 \\
\rowcolor{maroon!30} BankHeist & two police cars & Fig. 3, 4, 5 \\
Bowling & shift player & Fig. 3\\
\rowcolor{maroon!30} Bowling & top pins & Fig. 3, 4, 5 \\
\rowcolor{maroon!30} Boxing & color player red & Fig. 3, 4, 5 \\
Boxing & switch positions & Fig. 3 \\
Breakout & color all blocks red & Fig. 3 \\
\rowcolor{maroon!30} Breakout & color player and ball red & Fig. 3, 4, 5 \\
FishingDerby & fish on different sides & Fig. 3 \\
Freeway & all black cars & Fig. 3 \\
Freeway & stop all cars edge & Fig. 3 \\
\rowcolor{maroon!30} Freeway & stop all cars & Fig. 3, 4, 5 \\
Freeway & stop random car & Fig. 3 \\
\rowcolor{maroon!30} Frostbite & reposition floes easy & Fig. 3, 4, 5 \\
Jamesbond & straight shots & Fig. 3 \\
\rowcolor{maroon!30} Kangaroo & no danger & Fig. 3, 4, 5 \\
\rowcolor{maroon!30} MsPacman & set level 1 & Fig. 3, 4, 5 \\
MsPacman & set level 2 & Fig. 3 \\
MsPacman & set level 3 & Fig. 3 \\
\rowcolor{maroon!30} Pong & lazy enemy & Fig. 3, 4, 5 \\
RiverRaid & exploding fuels & Fig. 3 \\
RiverRaid & game color change01 & Fig. 3 \\
\rowcolor{maroon!30} RiverRaid & restricted firing & Fig. 3, 4, 5 \\
\rowcolor{maroon!30} Seaquest & disable enemies & Fig. 3, 4, 5 \\
SpaceInvaders & relocate shields off by one & Fig. 3 \\
\rowcolor{maroon!30} SpaceInvaders & relocate shields off by three & Fig. 3, 4, 5 \\
\rowcolor{maroon!30} StarGunner & remove mountains & Fig. 3, 4, 5 \\
StarGunner & static bomber & Fig. 3 \\
StarGunner & static flyers & Fig. 3 \\
StarGunner & static mountains & Fig. 3 \\
\bottomrule
\end{tabular}
\end{table}

\clearpage
\subsection{Can Deep Agents solve simplifications? }
\label{appendix:Q1}
We present the raw scores obtained by all evaluated deep RL agents across a set of Atari games and their corresponding HackAtari variations. Each agent was trained solely on the original environment and evaluated on both the unmodified and modified versions without any fine-tuning or adaptation. Performance is measured over 30 episodes per configuration, averaged across 3 random seeds, and reported as interquartile means (IQM) with 95\% confidence intervals.

The task variations are primarily designed as \textit{simplifications}—modifications that preserve core game mechanics while reducing visual or strategic complexity or modifications that should not influence the performance of an agent, e.g., color changes. They provide a strong test for determining whether agents have learned task-aligned policies or are relying on superficial cues. A reliable agent should maintain or improve performance under these conditions.

The table also indicates which task variations were used in our human study (cf. Appendix E). These entries are highlighted in {\color{maroon!60}pink}. 

\begin{table}[h]
    \centering
    \setlength{\tabcolsep}{4pt}
    \renewcommand{\arraystretch}{1.1}
    \caption{Raw episodic scores for all evaluated deep RL agents across selected Atari games and their HackAtari task variations (cf. \autoref{fig:all_agents_ehns}). Each entry shows the interquartile mean (IQM) and 95\% confidence interval across 30 episodes and 3 random seeds. Variants marked in {\color{maroon!60}pink} were used in the human study (cf. Appendix~\ref{appendix:human_study_extended}) and in \autoref{fig:deep_agents_pc}.}
    \label{appendix:c1-drl}
    \resizebox{\linewidth}{!}{
\begin{tabular}{l|cccccc}
\midrule
Game (Variant) & DQN & C51 & PPO & IMPALA & i-DQN & MDQN \\ \midrule
\rowcolor{maroon!30} \underline{Amidar} & $867\ci{771,~956}$ & -- & $1052\ci{1017,~1111}$ & $1215\ci{1131, 1298}$ & $666\ci{452, 901}$ & $1481\ci{1225,~1928}$ \\
\rowcolor[gray]{0.95} \rowcolor{maroon!20} paint roller player & $71\ci{54,~94}$ & -- & $271\ci{189,~377}$ & $205\ci{166, 255}$ & $529\ci{439, 630}$ & $160\ci{94,~300}$ \\
\rowcolor[gray]{0.95} pig enemies & $480\ci{316,~664}$ & -- & $1313\ci{1177,~1454}$ & $1122\ci{890, 1369}$ & $70\ci{53, 86}$ & $708\ci{555,~1248}$ \\ \midrule
\rowcolor{maroon!30} \underline{Asterix} & $10212\ci{6734,~17884}$ & $7766\ci{5262,~17937}$ & $8588\ci{6915,~11412}$ & $3543\ci{2084, 7491}$ & $4017\ci{3194, 4888}$ & $5116\ci{3046,~9275}$ \\
\rowcolor[gray]{0.95} \rowcolor{maroon!20} obelix & $3812\ci{3062,~4593}$ & $81812\ci{52250,~121593}$ & $5594\ci{4687,~6750}$ & $240218\ci{136250, 346501}$ & $6269\ci{5288, 7365}$ & $4594\ci{3437,~5906}$ \\ \midrule
\rowcolor{maroon!30} \underline{BankHeist} & $1064\ci{1011,~1113}$ & -- & $1062\ci{1028,~1092}$ & $388\ci{222, 636}$ & $1291\ci{1226, 1366}$ & $1319\ci{1245,~1413}$ \\
\rowcolor[gray]{0.95} \rowcolor{maroon!20} two police cars & $0\ci{0,~0}$ & -- & $0\ci{0,~0}$ & $0\ci{0, 0}$ & $0\ci{0, 0}$ & $0\ci{0,~0}$ \\ \midrule
\rowcolor{maroon!30} \underline{Bowling} & $34\ci{30,~39}$ & $46\ci{40,~53}$ & $67\ci{65,~69}$ & $45\ci{42, 49}$ & $42\ci{40, 45}$ & $33\ci{33,~35}$ \\
\rowcolor[gray]{0.95} shift player & $36\ci{31,~41}$ & $45\ci{39,~50}$ & $63\ci{54,~65}$ & $39\ci{35, 44}$ & $43\ci{39, 48}$ & $37\ci{34,~40}$ \\
\rowcolor[gray]{0.95} \rowcolor{maroon!20} top pins & $40.88\ci{9, 94}$ & $37.38\ci{23, 56}$ & $80.00\ci{60, 96}$ & $62\ci{38, 87}$ & $26\ci{21, 41}$ & $181.38\ci{139, 213}$ \\ \midrule
\rowcolor{maroon!30} \underline{Boxing} & $92\ci{88,~94}$ & $77\ci{68,~85}$ & $99\ci{97,~99}$ & $94\ci{92, 97}$ & $96\ci{95, 98}$ & $95\ci{93,~97}$ \\
\rowcolor[gray]{0.95} \rowcolor{maroon!20} color player red & $-2\ci{-5,~0}$ & $-7\ci{-12,~-2}$ & $-1\ci{-2,~0}$ & $-3\ci{-8, 0}$ & $-2\ci{-6, -0}$ & $-2\ci{-3,~0}$ \\
\rowcolor[gray]{0.95} switch positions & $47\ci{21,~68}$ & $-16\ci{-29,~-1}$ & $57\ci{34,~68}$ & $94\ci{92, 97}$ & $67\ci{60, 72}$ & $46\ci{23,~68}$ \\ \midrule
\rowcolor{maroon!30} \underline{Breakout} & $123\ci{85,~178}$ & $41\ci{27,~64}$ & $391\ci{366,~409}$ & $303\ci{245, 350}$ & $259\ci{219, 295}$ & $284\ci{202,~349}$ \\
\rowcolor[gray]{0.95} color all blocks red & $144\ci{91,~215}$ & $23\ci{12,~37}$ & $413\ci{396,~424}$ & $247\ci{175, 322}$ & $103\ci{63, 173}$ & $261\ci{176,~331}$ \\
\rowcolor[gray]{0.95} \rowcolor{maroon!20} color player and ball red & $4\ci{2,~6}$ & $2\ci{1,~3}$ & $4\ci{3,~6}$ & $76\ci{59, 121}$ & $8\ci{7, 10}$ & $15\ci{11,~19}$ \\ \midrule
\underline{FishingDerby} & $32\ci{24,~40}$ & $5\ci{0,~13}$ & $41\ci{37,~44}$ & $34\ci{31, 38}$ & $32\ci{28, 38}$ & $42\ci{32,~50}$ \\
\rowcolor[gray]{0.95} fish on different sides & $-90\ci{-92,~-88}$ & $-84\ci{-87,~-82}$ & $-98\ci{-98,~-97}$ & $27\ci{25, 30}$ & $20\ci{16, 24}$ & $-94\ci{-96,~-90}$ \\ \midrule
\rowcolor{maroon!30} \underline{Freeway} & $27\ci{16,~33}$ & $32\ci{31,~32}$ & $31\ci{30,~31}$ & $33\ci{33, 33}$ & $33\ci{33, 33}$ & $34\ci{33,~34}$ \\
\rowcolor[gray]{0.95} all black cars & $10\ci{7,~13}$ & $18\ci{16,~20}$ & $25\ci{23,~26}$ & $28\ci{27, 29}$ & $26\ci{24, 29}$  & $25\ci{24,~26}$ \\
\rowcolor[gray]{0.95} stop all cars edge & $6\ci{1,~13}$ & $34\ci{22,~40}$ & $39\ci{37,~39}$ & $11\ci{6, 19}$ & $0\ci{0, 6}$ & $40\ci{39,~40}$ \\
\rowcolor[gray]{0.95} \rowcolor{maroon!20} stop all cars & $0\ci{0,~0}$ & $0\ci{0,~0}$ & $7.56\ci{0, 20}$ & $35\ci{29, 38}$ & $0\ci{0, 1}$ & $0.38\ci{0, 1}$ \\
\rowcolor[gray]{0.95} stop random car & $16\ci{9,~20}$ & $22\ci{20,~22}$ & $21\ci{20,~21}$ & $22\ci{21, 22}$ & $16\ci{16, 18}$ & $24\ci{22,~24}$ \\ \midrule
\rowcolor{maroon!30} \underline{Frostbite} & $5431\ci{4411,~6624}$ & $5165\ci{4466,~6055}$ & $301\ci{288,~311}$ & $281\ci{262, 299}$ & $6198\ci{5196, 7193}$ & $8349\ci{6961,~8951}$ \\
\rowcolor[gray]{0.95} \rowcolor{maroon!20} reposition floes easy & $198\ci{43,~461}$ & $22\ci{0,~63}$ & $9\ci{0,~28}$ & $0\ci{0, 2}$ & $0\ci{0, 0}$  & $28\ci{12,~58}$ \\ \midrule
\underline{Jamesbond} & $972\ci{887,~1046}$ & $3056\ci{1099,~5134}$ & $2216\ci{1850,~2596}$ & $11778\ci{9775, 13891}$ & $600\ci{571, 637}$ & $722\ci{662,~778}$ \\
\rowcolor[gray]{0.95} straight shots & $22\ci{6,~46}$ & $94\ci{53,~762}$ & $481\ci{215,~987}$ & $1772\ci{734, 3319}$ & $63\ci{46, 77}$ & $91\ci{65,~118}$ \\ \midrule
\rowcolor{maroon!30} \underline{Kangaroo} & $9669\ci{7962,~11487}$ & $7456\ci{5231,~10012}$ & $13525\ci{12275,~14243}$ & $6244\ci{4200, 8181}$ & $14269\ci{13581, 14550}$ & $13588\ci{12581,~14012}$ \\
\rowcolor[gray]{0.95} \rowcolor{maroon!20} no danger & $19\ci{0,~50}$ & $6\ci{0,~37}$ & $0\ci{0,~0}$ & $0\ci{0, 0}$ & $15\ci{0, 54}$ & $81\ci{50,~100}$ \\ \midrule
\rowcolor{maroon!30} \underline{MsPacman} & $4232\ci{3632,~4636}$ & -- & $7225\ci{6684,~7600}$ & $3652\ci{3089, 4212}$  & $3667\ci{3142, 4180}$ & $3961\ci{3208,~4579}$ \\
\rowcolor[gray]{0.95} \rowcolor{maroon!20} set level 1 & $602\ci{500,~718}$ & -- & $357\ci{291,~438}$ & $631\ci{527, 744}$ & $328\ci{283, 414}$ & $1412\ci{1216,~1555}$ \\
\rowcolor[gray]{0.95} set level 2 & $333\ci{240,~426}$ & -- & $326\ci{229,~428}$ & $786\ci{641, 979}$ & $408\ci{368, 465}$ & $432\ci{307,~513}$ \\
\rowcolor[gray]{0.95} set level 3 & $356\ci{306,~412}$ & -- & $347\ci{295,~419}$ & $738\ci{477, 986}$ & $176\ci{139, 214}$ & $231\ci{158,~302}$ \\ \midrule
\rowcolor{maroon!30} \underline{Pong} & $19\ci{17,~19}$ & $5\ci{4,~6}$ & $18\ci{14,~19}$ & $9\ci{7, 11}$ & $19\ci{19, 20}$ & $20\ci{18,~20}$ \\
\rowcolor[gray]{0.95} \rowcolor{maroon!20} lazy enemy & $-15\ci{-17,~-12}$ & $-13\ci{-15,~-8}$ & $-15\ci{-17,~-13}$ & $6\ci{1, 10}$ & $-18\ci{-19, -17}$ & $-16\ci{-18,~-12}$ \\ \midrule
\rowcolor{maroon!30} \underline{Riverraid} & $14030\ci{12366,~14900}$ & -- & $16186\ci{15134,~16920}$ & $17421\ci{14467, 20599}$ & $11251\ci{10166, 12375}$ & $15340\ci{14739,~15953}$ \\
\rowcolor[gray]{0.95} exploding fuels & $5530\ci{5129,~5847}$ & -- & $7972\ci{7710,~8233}$ & $6756\ci{5249, 8446}$ & $4763\ci{4229, 5383}$ & $6094\ci{5454,~6671}$ \\
\rowcolor[gray]{0.95} game color change01 & $439\ci{405,~475}$ & -- & $368\ci{256,~480}$ & $721\ci{592, 888}$ & $379\ci{302, 461}$ & $234\ci{138,~427}$ \\
\rowcolor[gray]{0.95} \rowcolor{maroon!20} restricted firing & $757\ci{576,~984}$ & -- & $933\ci{520,~1820}$ & $1656\ci{1313, 2114}$ & $1573\ci{1227, 1928}$ & $1155\ci{883,~1553}$ \\ \midrule
\rowcolor{maroon!30} \underline{Seaquest} & $5916\ci{5281,~6750}$ & $123622.50\ci{38904,~222808}$ & $1836\ci{1826, 1840}$ & $951\ci{945, 958}$ & $6336\ci{4995, 7672}$ & $18170\ci{15821,~21457}$ \\
\rowcolor[gray]{0.95} \rowcolor{maroon!20} disable enemies & $268\ci{0,~717}$ & $0\ci{0,~0}$ & $0\ci{0,~0}$ & $0\ci{0, 0}$ & $1134\ci{409, 2030}$ & $1392\ci{676,~5015}$ \\ \midrule
\rowcolor{maroon!30} \underline{SpaceInvaders} & $4173\ci{2305,~7020}$ & -- & $1406\ci{1179,~1668}$ & $11514\ci{5439, 19878}$ & $3225\ci{2222, 4798}$ & $6183\ci{3195,~10738}$ \\
\rowcolor[gray]{0.95} relocate shields off by one & $1314\ci{949,~2302}$ & -- & $1533\ci{1320,~1725}$ & $10415\ci{6222, 15419}$ & $3853\ci{2041, 6291}$ & $2863\ci{1983,~6178}$ \\
\rowcolor[gray]{0.95} \rowcolor{maroon!20} relocate shields off by three & $475\ci{322,~684}$ & -- & $1158\ci{919,~1385}$ & $8252\ci{4306, 13849}$ & $1037\ci{864, 1454}$ & $820\ci{578,~1344}$ \\ \midrule
\rowcolor{maroon!30} \underline{StarGunner} & $59269\ci{55750,~61300}$ & $31894\ci{22831,~42906}$ & $29288\ci{15737,~42162}$ & $166881\ci{159112, 177275}$ & $55411\ci{51977, 58388}$ & $63181\ci{61068,~64781}$ \\
\rowcolor[gray]{0.95} \rowcolor{maroon!20} remove mountains & $50838\ci{43012,~56243}$ & $24688\ci{15562,~35025}$ & $15781\ci{8693,~23206}$ & $171169\ci{162438, 180062}$ & $9726\ci{7177, 12427}$ & $58931\ci{56487,~60975}$ \\
\rowcolor[gray]{0.95} static bomber & $64506\ci{62362,~66925}$ & $25250\ci{17175,~36750}$ & $41475\ci{24381,~55043}$ & $172731\ci{163138, 181425}$ & $55146\ci{51323, 57812}$ & $66475\ci{65006,~68143}$ \\
\rowcolor[gray]{0.95} static flyers & $235031\ci{209081,~260543}$ & $24856\ci{14675,~36993}$ & $4244\ci{950,~13562}$ & $878106\ci{869850, 884031}$ & $68684\ci{40065, 112265}$ & $190569\ci{133318,~246019}$ \\
\rowcolor[gray]{0.95} static mountains & $46950\ci{37356,~54575}$ & $32769\ci{23600,~42875}$ & $19562\ci{10293,~29275}$ & $159169\ci{152875, 165269}$ & $53965\ci{49888, 56515}$ & $61769\ci{59487,~63681}$ \\ \midrule
\end{tabular}
}
\end{table}

\clearpage

\subsection{Can Object-centricity solve misalignment?}
\label{appendix:Q3}

This section presents the raw performance scores of object-centric agents evaluated on the same HackAtari task variations described in Appendix C.1. These agents incorporate structured inductive biases through symbolic, object-based, or mask-based representations. While such representations are often credited with improved interpretability and generalization in structured environments, our results assess whether they also support better alignment with simplified task variants.

All object-centric agents were trained using Proximal Policy Optimization (PPO) and receive object-based inputs—either through learned visual decompositions (e.g., binary masks or spatial planes) or symbolic representations (e.g., object class vectors from OCAtari and ScoBots). For comparison, we include the standard PPO baseline trained on raw pixel inputs.

Each entry reports the interquartile mean (IQM) and 95\% confidence interval across 30 episodes and 3 random seeds. A reliable object-centric agent should demonstrate resilience to both visual and gameplay simplifications, avoiding performance drops that indicate overreliance on spurious correlations. However, as our findings show, object-centricity improves robustness primarily to visual perturbations, but fails to fully eliminate shortcut behavior in more complex dynamics-altering variations.

\begin{table}[h]
    \centering
    \setlength{\tabcolsep}{4pt}
    \renewcommand{\arraystretch}{1.1}
    \caption{Episodic scores for object-centric RL agents on HackAtari task variations. All agents use PPO and differ only in their input representation (e.g., binary masks, planes, or symbolic vectors). Models were taken by~\citet{bluml2025deep} and~\citet{delfosse2024ocatari, delfosse2024interpretable} or self-trained. Scores are reported as interquartile means (IQM) with 95\% confidence intervals, evaluated over 30 episodes and 3 seeds. PPO with raw pixels, as done by~\citet{Mnih2015dqn}, is included as a baseline. Games and Modifications in {\color{maroon!60}pink} were used in \autoref{fig:oc_agents_pc} and are identical to the human study.}
    \label{appendix:c2-oc}
    \resizebox{\linewidth}{!}{
\begin{tabular}{@{}lccccccc@{}}
\toprule
Game (Variant) & PPO & Object Masks & Binary Masks & Class Masks & Planes & Semantic Vector & ScoBots \\ \midrule
\rowcolor{maroon!30} \underline{Amidar} & $1052\ci{1017,~1111}$ & $554 \ci{493,~615}$ & $525 \ci{430,~605}$ & $479 \ci{442,~513}$ & $527 \ci{509,~552}$ & $357\ci{325,~407}$ & $116\ci{94,~128}$ \\
\rowcolor[gray]{0.95} \rowcolor{maroon!20} paint roller player & $271\ci{189,~377}$ & $315 \ci{251,~369}$ & $535 \ci{435,~610}$ & $492 \ci{458,~526}$ & $510 \ci{491,~528}$ & $301\ci{252,~350}$ & $116\ci{91,~129}$ \\
\rowcolor[gray]{0.95} pig enemies & $1313\ci{1177,~1454}$ & $384 \ci{332,~426}$ & $513 \ci{415,~597}$ & $516 \ci{465,~554}$ & $108 \ci{91,~118}$ & $25\ci{16,~34}$ & $18\ci{17,~20}$ \\ \midrule
\rowcolor{maroon!30} \underline{Asterix} & $8588\ci{6915,~11412}$ & $4065\ci{3522, 4497}$ & $171\ci{116, 241}$ & $4181\ci{3528, 4900}$ & $91753\ci{83844, 98809}$ & $733\ci{358,~1066}$ & $883\ci{624,~1133}$ \\
\rowcolor[gray]{0.95} \rowcolor{maroon!20} obelix & $5594\ci{4687,~6750}$ & $4562\ci{3812, 5406}$ & $2062\ci{1719, 2594}$ & $3375\ci{2750, 4469}$ & $30343\ci{21312, 46312}$ & $4583\ci{3039,~7543}$ & $4000\ci{3333,~5500}$ \\ \midrule
\rowcolor{maroon!30} \underline{BankHeist} & $1062\ci{1028,~1092}$ & $781 \ci{769,~794}$ & $1192 \ci{1158,~1211}$ & $1181 \ci{1151,~1205}$ & $1154 \ci{1013,~1302}$ & $1163\ci{1155,~1173}$ & $720\ci{670,~741}$ \\
\rowcolor[gray]{0.95} \rowcolor{maroon!20} two police cars & $0\ci{0,~0}$ & $0 \ci{0,~0}$ & $0 \ci{0,~0}$ & $0 \ci{0,~0}$ & $0 \ci{0,~0}$ & $61\ci{42,~91}$ & $108\ci{80,~133}$ \\ \midrule
\rowcolor{maroon!30} \underline{Bowling} & $67\ci{65,~69}$ & $65 \ci{62,~67}$ & $70 \ci{67,~71}$ & $64 \ci{61,~67}$ & $99\ci{99,~99}$ & $99\ci{97,~99}$ & $51\ci{41,~60}$ \\
\rowcolor[gray]{0.95} shift player & $63\ci{54,~65}$ & $63 \ci{61,~66}$ & $67 \ci{64,~70}$ & $62 \ci{60,~66}$ & $84\ci{53,~85}$ & $0\ci{0,~0}$ & $51\ci{41,~61}$ \\
\rowcolor[gray]{0.95} \rowcolor{maroon!20} top pins & $80\ci{60,~96}$ & $58\ci{37, 78}$ & $98\ci{49, 159}$ & $184\ci{171, 195}$ & $98\ci{73, 124}$ & $8\ci{8,~8}$ & $0\ci{0,~0}$ \\ \midrule
\rowcolor{maroon!30} \underline{Boxing} & $99\ci{97,~99}$ & $94 \ci{92,~96}$ & $96 \ci{95,~97}$ & $94 \ci{93,~96}$ & $97 \ci{96,~98}$ & $87\ci{78,~96}$ & $51\ci{41,~60}$ \\
\rowcolor[gray]{0.95} \rowcolor{maroon!20} color player red & $-1\ci{-2,~0}$ & $80 \ci{76,~83}$ & $96 \ci{95,~97}$ & $95 \ci{93,~96}$ & $96 \ci{95,~98}$ & $87\ci{78,~95}$ & $51\ci{41,~61}$ \\
\rowcolor[gray]{0.95} switch positions & $57\ci{34,~68}$ & $-53 \ci{-77,~-11}$ & $-65 \ci{-97,~-18}$ & $29 \ci{-12,~65}$ & $37 \ci{-14,~77}$ & $-35\ci{-42,~-31}$ & $63\ci{45,~78}$ \\ \midrule
\rowcolor{maroon!30} \underline{Breakout} & $391\ci{366,~409}$ & $216 \ci{156,~279}$ & $259 \ci{212,~295}$ & $222 \ci{171,~267}$ & $371 \ci{348,~390}$ & $38\ci{25,~51}$ & $21\ci{14,~28}$ \\
\rowcolor[gray]{0.95} color all blocks red & $413\ci{396,~424}$ & $299 \ci{247,~328}$ & $278 \ci{229,~312}$ & $259 \ci{207,~290}$ & $372 \ci{330,~403}$ & $38\ci{25,~51}$ & $21\ci{14,~28}$ \\
\rowcolor[gray]{0.95} \rowcolor{maroon!20} color player and ball red & $4\ci{3,~6}$ & $99 \ci{72,~155}$ & $256 \ci{201,~287}$ & $216 \ci{165,~266}$ & $390 \ci{352,~407}$ & $38\ci{25,~51}$ & $21\ci{14,~29}$ \\ \midrule
\underline{FishingDerby} & $41\ci{37,~44}$ & $20\ci{13, 28}$ & $-81\ci{-84, -79}$ & $-62\ci{-70, -56}$ & $47\ci{42, 52}$ & $18\ci{11,~26}$ & $23\ci{18,~27}$ \\
\rowcolor[gray]{0.95} fish on different sides & $-98\ci{-98,~-97}$ & $23\ci{9, 30}$ & $-86\ci{-89, -83}$ & $-60\ci{-67, -53}$ & $30\ci{27, 32}$ & $3\ci{-4,~21}$ & $-4\ci{-12,~4}$ \\ \midrule
\rowcolor{maroon!30} \underline{Freeway} & $31\ci{30,~31}$ & $33 \ci{32,~33}$ & $33 \ci{33,~33}$ & $32 \ci{32,~32}$ & $33 \ci{33,~34}$ & $31\ci{31,~32}$ & $30\ci{28,~31}$ \\
\rowcolor[gray]{0.95} all black cars & $25\ci{23,~26}$ & $23 \ci{21,~25}$ & $33 \ci{32,~33}$ & $32 \ci{32,~33}$ & $33 \ci{33,~34}$ & $31\ci{31,~32}$ & $30\ci{28,~31}$ \\
\rowcolor[gray]{0.95} stop all cars edge & $39\ci{37,~39}$ & $0 \ci{0,~0}$ & $0 \ci{0,~0}$ & $7 \ci{0,~19}$ & $22 \ci{6,~37}$ & $0 \ci{0,~0}$ & $41\ci{41,~41}$ \\
\rowcolor[gray]{0.95} \rowcolor{maroon!20} stop all cars & $7\ci{0,~20}$ & $0\ci{0, 0}$ & $0\ci{0, 0}$ & $32\ci{20, 40}$ & $19\ci{4, 36}$ & $0 \ci{0,~0}$ & $41\ci{41,~41}$ \\
\rowcolor[gray]{0.95} stop random car & $21\ci{20,~21}$ & $18 \ci{17,~20}$ & $19 \ci{18,~20}$ & $19 \ci{18,~20}$ & $20 \ci{20,~21}$ & $15\ci{13,~16}$ & $21\ci{20,~22}$ \\ \midrule
\rowcolor{maroon!30} \underline{Frostbite} & $301\ci{288,~311}$ & $278 \ci{270,~290}$ & $290 \ci{278,~304}$ & $265 \ci{258,~270}$ & $278 \ci{271,~286}$ & $2821\ci{2488,~3134}$ & $1931\ci{1806,~2125}$ \\
\rowcolor[gray]{0.95} \rowcolor{maroon!20} reposition floes easy & $9\ci{0,~28}$ & $11 \ci{0,~32}$ & $4 \ci{0,~11}$ & $1 \ci{0,~24}$ & $28 \ci{0,~79}$ & $60\ci{43,~75}$ & $70\ci{60,~80}$ \\ \midrule
\underline{Jamesbond} & $2216\ci{1850,~2596}$ & -- & -- & -- & -- & $591\ci{525,~1583}$ & $175\ci{125,~241}$ \\
\rowcolor[gray]{0.95} straight shots & $481\ci{215,~987}$ & -- & -- & -- & -- & $108\ci{58,~200}$ & $0\ci{0,~21}$ \\ \midrule
\rowcolor{maroon!30} \underline{Kangaroo} & $13525\ci{12275,~14243}$ & $112\ci{0, 300}$ & $8425\ci{5550, 10625}$ & $112\ci{0, 300}$ & $1800\ci{1800, 1800}$ & $0\ci{0,~0}$ & $0\ci{0,~0}$ \\
\rowcolor[gray]{0.95} \rowcolor{maroon!20} no danger & $0\ci{0,~0}$ & $0\ci{0,~0}$ & $0\ci{0,~0}$ & $0\ci{0,~0}$ & $0\ci{0,~0}$ & $0\ci{0,~0}$ & $0\ci{0,~0}$ \\ \midrule
\rowcolor{maroon!30} \underline{MsPacman} & $7225\ci{6684,~7600}$ & $5195 \ci{4476,~5724}$ & $4227 \ci{3646,~4896}$ & $4313 \ci{3790,~4899}$ & $6211 \ci{5471,~6794}$ & $4583\ci{3650,~5298}$ & $3620\ci{3086,~3620}$ \\
\rowcolor[gray]{0.95} \rowcolor{maroon!20} set level 1 & $357\ci{291,~438}$ & $281 \ci{236,~319}$ & $548 \ci{400,~778}$ & $378 \ci{256,~553}$ & $294 \ci{234,~366}$ & $210\ci{210,~210}$ & $90\ci{90,~90}$ \\
\rowcolor[gray]{0.95} set level 2 & $326\ci{229,~428}$ & {$285\ci{258, 317}$} & {$376\ci{323, 439}$} & {$695\ci{446, 969}$} & {$126\ci{104, 161}$} & $170\ci{170,~170}$ & $90\ci{90,~90}$ \\
\rowcolor[gray]{0.95} set level 3 & $347\ci{295,~419}$ & {$259\ci{222, 301}$} & {$628\ci{512, 775}$} & {$228\ci{110, 497}$} & {$124\ci{98, 172}$} & $1610\ci{1610,~1610}$ & $90\ci{90,~90}$ \\ \midrule
\rowcolor{maroon!30} \underline{Pong} & $18\ci{14,~19}$ & $18 \ci{18,~19}$ & $19 \ci{18,~20}$ & $19 \ci{19,~20}$ & $19 \ci{19,~20}$ & $19\ci{19,~20}$ & $16\ci{15,~19}$ \\
\rowcolor[gray]{0.95} \rowcolor{maroon!20} lazy enemy & $-15\ci{-17,~-13}$ & $12 \ci{10,~15}$ & $12 \ci{5,~16}$ & $3 \ci{-4,~9}$ & $17 \ci{16,~18}$ & $-20\ci{-20,~-19}$ & $17\ci{15,~-19}$ \\ \midrule
\rowcolor{maroon!30} \underline{Riverraid} & $16186\ci{15134,~16920}$ & $7948 \ci{7814,~8092}$ & $7958 \ci{7739,~8204}$ & $7803 \ci{7552,~7993}$ & $7990 \ci{7824,~8198}$ & $2175\ci{2086,~2350}$ & $2230\ci{2074,~2368}$ \\
\rowcolor[gray]{0.95} exploding fuels & $7972\ci{7710,~8233}$ & $4111 \ci{3886,~4201}$ & $3651 \ci{3071,~4038}$ & $4039 \ci{3801,~4279}$ & $4136 \ci{3999,~4272}$ & $1068\ci{1010,~1241}$ & $1186\ci{952,~1221}$ \\
\rowcolor[gray]{0.95} game color change01 & $368\ci{256,~480}$ & $7638 \ci{7474,~7811}$ & $7850 \ci{7745,~8011}$ & $7867 \ci{7567,~8105}$ & $8042 \ci{7841,~8242}$ & $2175\ci{2091,~2353}$ & $2230\ci{2075,~2353}$ \\
\rowcolor[gray]{0.95} \rowcolor{maroon!20} restricted firing & $933\ci{520,~1820}$ & $523 \ci{511,~539}$ & $1677 \ci{1460,~1964}$ & $2261 \ci{1850,~2594}$ & $1864 \ci{1524,~2359}$ & $200\ci{200,~200}$ & $2741\ci{2216,~3058}$ \\ \midrule
\rowcolor{maroon!30} \underline{Seaquest} & $1836\ci{1826, 1840}$ & {$1656\ci{1388, 1820}$} & {$732\ci{465, 1141}$} & {$1832\ci{1815, 1840}$} & {$2013\ci{1840, 2308}$} & $293\ci{254,~353}$ & $293\ci{256,~360}$ \\
\rowcolor[gray]{0.95} \rowcolor{maroon!20} disable enemies & $0\ci{0,~0}$ & $0\ci{0,~0}$ & $0\ci{0,~0}$ & $0\ci{0,~0}$ & $0\ci{0,~0}$ & $0\ci{0,~0}$ & $0\ci{0,~0}$ \\ \midrule
\rowcolor{maroon!30} \underline{SpaceInvaders} & $1406\ci{1179,~1668}$ & $688 \ci{600,~806}$ & $826 \ci{763,~909}$ & $554 \ci{475,~634}$ & $1557 \ci{1247,~1851}$ & $278\ci{205,~321}$ & $335\ci{215,~451}$ \\
\rowcolor[gray]{0.95} relocate shields off by one & $1533\ci{1320,~1725}$ & $621 \ci{560,~693}$ & $803 \ci{730,~884}$ & $483 \ci{362,~588}$ & $1787 \ci{1416,~2217}$ & $290\ci{256,~358}$ & $241\ci{208,~318}$ \\
\rowcolor[gray]{0.95} \rowcolor{maroon!20} relocate shields off by three & $1158\ci{919,~1385}$ & $469 \ci{389,~571}$ & $589 \ci{474,~683}$ & $288 \ci{245,~325}$ & $1555 \ci{1189,~1899}$ & $239\ci{186,~347}$ & $266\ci{252,~330}$ \\ \midrule
\rowcolor{maroon!30} \underline{StarGunner} & $29288\ci{15737,~42162}$ & {$8406\ci{3938, 14106}$} & {$962\ci{900, 1000}$} & {$20812\ci{10075, 32231}$} & {$88787\ci{80131, 94862}$} & $10166\ci{5991,~18597}$ & $11550\ci{7357,~15600}$ \\
\rowcolor[gray]{0.95} \rowcolor{maroon!20} remove mountains & $15781\ci{8693,~23206}$ & {$9943\ci{5144, 15150}$} & {$1000\ci{1000, 1000}$} & {$23656\ci{13000, 33488}$} & {$87287\ci{78219, 95475}$} & $13950\ci{9640,~18069}$ & $11250\ci{7700,~14447}$ \\
\rowcolor[gray]{0.95} static bomber & $41475\ci{24381,~55043}$ & {$13375\ci{6512, 20819}$} & {$1000\ci{1000, 1000}$} & {$43031\ci{26325, 54312}$} & {$83268\ci{68862, 102025}$} & $24766\ci{21231,~28208}$ & $11550\ci{6815,~15821}$ \\
\rowcolor[gray]{0.95} static flyers & $4244\ci{950,~13562}$ & {$11625\ci{4881, 27369}$} & {$343\ci{275, 438}$} & {$83881\ci{27206, 178088}$} & {$577606\ci{505392, 646894}$} & $23950\ci{7666,~59957}$ & $11550\ci{7524,~15760}$ \\
\rowcolor[gray]{0.95} static mountains & $19562\ci{10293,~29275}$ & {$7325\ci{3888, 11600}$} & {$1000\ci{1000, 1000}$} & {$13718\ci{6506, 25269}$} & {$84487\ci{77275, 92850}$} & $13516\ci{8041,~17467}$ & $11550\ci{7323,~15592}$ \\ \bottomrule
\end{tabular}
}
\end{table}

\clearpage

\subsection{Can Humans Solve these Simplifications?}
\label{appendix:Q2}

To validate that our selected HackAtari variations constitute genuine simplifications rather than increased difficulty, we conducted a user study evaluating human adaptation to a subset of modified environment (cf. Appendix~\ref{appendix:human_study_extended}). 

Table~\ref{tab:c3-human-results} presents the raw scores for each game and its corresponding variation. We report the mean over IQM (cf. Appendix~\ref{appendix:metrics}) and 95\% confidence intervals across participants, alongside the random agent baseline and expert human scores (from~\citet{Badia2020agent57}) when available. This enables robust comparison between deep RL agents and non-expert human players.

\begin{table}[h]
    \centering
    \setlength{\tabcolsep}{4pt}
    \renewcommand{\arraystretch}{1.1}
    \caption{Raw scores for humans, random agents, and expert human baselines on original and modified tasks. Variants used in the human study are marked and match those highlighted in Table~\ref{appendix:c1-drl}. Scores are reported with 95\% confidence intervals and were used in \autoref{fig:deep_agents_pc}.}
    \label{tab:c3-human-results}
    \resizebox{.9\linewidth}{!}{
\begin{tabular}{@{}lccccc@{}}
\toprule
Game (Variant) & Random & Random (Badia et al.) & Human & Human (Badia et al.) \\ \midrule
\rowcolor{maroon!30} \underline{Amidar} & $1.62\ci{0,~3}$ & 5.8 & $114.58\ci{46,~220}$ & 7127.7 \\
\rowcolor[gray]{0.95} \rowcolor{maroon!20} paint roller player & $1.31\ci{0,~2}$ & -- & $360.43\ci{52,~930}$ & -- \\
\rowcolor[gray]{0.95} pig enemies & $1.06\ci{0,~3}$ & -- & -- & -- \\ \midrule
\rowcolor{maroon!30} \underline{Asterix} & $218.75\ci{162,~275}$ & 210.0 & $1357.54\ci{458,~2462}$ & 1719.5 \\
\rowcolor[gray]{0.95} \rowcolor{maroon!20} obelix & $2062.50\ci{1688,~2500}$ & -- & $20384.03\ci{4624,~39712}$ & -- \\ \midrule
\rowcolor{maroon!30} \underline{BankHeist} & $13.12\ci{10,~17}$ & 14.2 & $268.67\ci{98,~506}$ & 753.1 \\
\rowcolor[gray]{0.95} \rowcolor{maroon!20} two police cars & $0.00\ci{0,~29}$ & -- & $242.08\ci{72,~470}$ & -- \\ \midrule
\rowcolor{maroon!30} \underline{Bowling} & $63.12\ci{54,~65}$ & 23.1 & $109.74\ci{88,~129}$ & 160.7 \\
\rowcolor[gray]{0.95} shift player & $24.25\ci{22,~27}$ & -- & -- & -- \\
\rowcolor[gray]{0.95} \rowcolor{maroon!20} top pins & $90.44\ci{80,~104}$ & -- & $192.17\ci{157,~220}$ & -- \\ \midrule
\rowcolor{maroon!30} \underline{Boxing} & $1.50\ci{0,~3}$ & 0.1 & $-2.75ci{-5,~0}$ & 12.1 \\
\rowcolor[gray]{0.95} \rowcolor{maroon!20} color player red & $1.06\ci{-0,~3}$ & -- & $-0.93\ci{-7,~6}$ & -- \\
\rowcolor[gray]{0.95} switch positions & $1.06\ci{-0,~3}$ & -- & -- & -- \\ \midrule
\rowcolor{maroon!30} \underline{Breakout} & $1.31\ci{1,~2}$ & 1.7 & $8.35\ci{3,~14}$ & 30.5 \\
\rowcolor[gray]{0.95} color all blocks red & $0.94\ci{0,~1}$ & -- & -- & -- \\
\rowcolor[gray]{0.95} \rowcolor{maroon!20} color player and ball red & $1.00\ci{0,~2}$ & -- & $10.31\ci{4,~19}$ & -- \\ \midrule
\rowcolor{maroon!30} \underline{Freeway} & $0.00\ci{0,~0}$ & 0.0 & $17.74\ci{12,~22}$ & 29.6 \\
\rowcolor[gray]{0.95} all black cars & $0.00\ci{0,~0}$ & -- & -- & -- \\
\rowcolor[gray]{0.95} stop all cars edge & $0.31\ci{0,~1}$ & -- & -- & -- \\
\rowcolor[gray]{0.95} \rowcolor{maroon!20} stop all cars & $0.53\ci{0,~1}$ & -- & $35.83\ci{32,~39}$ & -- \\
\rowcolor[gray]{0.95} stop random car & $0.00\ci{0,~0}$ & -- & -- & -- \\ \midrule
\rowcolor{maroon!30} \underline{Frostbite} & $71.25\ci{58,~84}$ & 65.2 & $1372.81\ci{167,~2834}$ & 4334.7 \\
\rowcolor[gray]{0.95} \rowcolor{maroon!20} reposition floes easy & $7.50\ci{1,~28}$ & -- & $7540.21\ci{1369,~15306}$ & -- \\ \midrule
\rowcolor{maroon!30} \underline{Kangaroo} & $0.00\ci{0,~50}$ & 52.0 & $459.97\ci{200,~791}$ & 3035.0 \\
\rowcolor[gray]{0.95} \rowcolor{maroon!20} no danger & $0.00\ci{0,~0}$ & -- & $2431.48\ci{1616,~3414}$ & -- \\ \midrule
\rowcolor{maroon!30} \underline{MsPacman} & $243.75\ci{219,~270}$ & 307.3 & $996.9\ci{518,~1492}$ & 6951.6 \\
\rowcolor[gray]{0.95} \rowcolor{maroon!20} set level 1 & $201.88\ci{174,~229}$ & -- & $673.6\ci{305,~1206}$ & -- \\
\rowcolor[gray]{0.95} set level 2 & $231.88\ci{203,~262}$ & -- & -- & -- \\
\rowcolor[gray]{0.95} set level 3 & $187.50\ci{165,~219}$ & -- & -- & -- \\ \midrule
\rowcolor{maroon!30} \underline{Pong} & $-20.62\ci{-21,~-20}$ & -20.7 & $-15.19\ci{-18,~-10}$ & 14.6 \\
\rowcolor[gray]{0.95} \rowcolor{maroon!20} lazy enemy & $-20.62\ci{-21,~-20}$ & -- & $-12.2\ci{-17,~-6}$ & -- \\ \midrule
\rowcolor{maroon!30} \underline{Riverraid} & $1533.75\ci{1348,~1688}$ & 1338.5 & $2406.56\ci{1428,~3584}$ & 17118.0 \\
\rowcolor[gray]{0.95} exploding fuels & $860.62\ci{659,~946}$ & -- & -- & -- \\
\rowcolor[gray]{0.95} game color change01 & $1475.62\ci{1341,~1637}$ & -- & -- & -- \\
\rowcolor[gray]{0.95} \rowcolor{maroon!20} restricted firing & $520.00\ci{520,~579}$ & -- & $13222.81\ci{8391,~18407}$ & -- \\ \midrule
\rowcolor{maroon!30} \underline{Seaquest} & $70.00\ci{54,~88}$ & 68.4 & $2626.39\ci{331,~5397}$ & 42054.7 \\
\rowcolor[gray]{0.95} \rowcolor{maroon!20} disable enemies & $0.00\ci{0,~0}$ & -- & $12191.67\ci{6836,~17288}$ & -- \\ \midrule
\rowcolor{maroon!30} \underline{SpaceInvaders} & $137.81\ci{115,~169}$ & 148.0 & $237.8\ci{148,~338}$ & 1668.7 \\
\rowcolor[gray]{0.95} relocate shields off by one & $117.50\ci{92,~148}$ & -- & -- & -- \\
\rowcolor[gray]{0.95} \rowcolor{maroon!20} relocate shields off by three & $125.62\ci{96,~160}$ & -- & $286.96\ci{175,~436}$ & -- \\ \midrule
\rowcolor{maroon!30} \underline{StarGunner} & $675.00\ci{550,~800}$ & 664.0 & $2435.83\ci{846,~5056}$ & 10250.0 \\
\rowcolor[gray]{0.95} \rowcolor{maroon!20} remove mountains & $656.25\ci{550,~781}$ & -- & $2936.67\ci{1200,~5197}$ & -- \\
\rowcolor[gray]{0.95} static bomber & $468.75\ci{356,~594}$ & -- & -- & -- \\
\rowcolor[gray]{0.95} static flyers & $512.50\ci{419,~650}$ & -- & -- & -- \\
\rowcolor[gray]{0.95} static mountains & $706.25\ci{588,~831}$ & -- & -- & -- \\ \bottomrule
\end{tabular}
}
\end{table}

\clearpage

\section{Code and Data}
\label{appendix:hackatari_extended}

To support reproducibility and further research, we will release all code, task variations, evaluation scripts, and selected model checkpoints as part of the supplementary materials and the code base through an anonymized repository upon acceptance.

\paragraph{HackAtari Environment.}
The HackAtari benchmark is implemented as a lightweight extension of the Gymnasium Atari framework. All modifications are applied via deterministic RAM-based wrappers, allowing reproducible control over game dynamics, visuals, and semantics. The environment includes:
\begin{itemize}
    \item Using any of task variants by name (\cf Appendix~\ref{appendix:environments}).
    \item Supporting scripts to help visualize and change the RAM of the current state on the fly, as well as other tools helping to create new variants. 
    \item Compatibility with Gym-style RL pipelines, wrappers and vectorized rollouts.
    \item Compatibility with StableBaselines3 and CleanRL training frameworks
\end{itemize}

\paragraph{Agent Models and Evaluation.}
We include pretrained checkpoints for selected deep and object-centric agents in the supplementary materials, covering a subset of games and variants. 

\paragraph{Human Study Infrastructure.}
To facilitate future user studies and make it easy to reproduce and test it, we will release the full source code for the web-based evaluation interface used in our human experiments (see Appendix~\ref{appendix:human_study_extended}). 

\paragraph{Anonymized Access and Deanonymization.}
For the review phase, we include all evaluation results and selected models in the supplementary materials. Deanonymized versions of HackAtari, HackTari-Web, Models, and the Dataset will be provided upon paper acceptance.

\clearpage
\section{Human Study: Evaluating Human Performance of Modified Atari Environments}
\label{appendix:human_study_extended}

\subsection{Motivation and Research Question}

The primary motivation of this study is to understand how visual and semantic modifications to Atari game environments affect human gameplay performance. Our central research question is:

\textit{How do different types of modifications influence human players' performance in Atari games?}

By comparing performance across original and modified versions of the same game, we aim to identify which transformations preserve or degrade task-relevant understanding for humans. This helps evaluate the alignment between human representation of the games and machine abstractions. It also informs us if the proposed environmental changes are doable for a human, compared to the performance of our machine learning agents.

\subsection{Study Design}

\subsubsection*{Participants}
We recruited participants through Prolific\footnote{\url{https://www.prolific.com/}}. A total of \textbf{160} participants completed the study. All participants were at least 18 years old and reported fluent English skills. Participation was voluntary, and individuals were monetarily compensated. Communication with the participants, if required, was done over Prolific.

\subsubsection*{Procedure}
Each participant completed the study in three phases:
\begin{enumerate}
    \item \textbf{Free Training (10-15 minutes):} Participants played an unmodified version of a randomly assigned Atari game. This phase allowed them to familiarize themselves with the core mechanics. Participants could proceed early by clicking an ``I’m ready'' button after 10 min.
    \item \textbf{Evaluation Phase 1 (15 minutes):} Participants played the original version of the same game. 
    \item \textbf{Evaluation Phase 2 (15 minutes):} Participants then played a modified version of the game. The modification applied was determined by a token-to-variant mapping and involved visual or semantic changes.
\end{enumerate}

After completing the evaluation rounds, participants answered short questions about the task (e.g., reward identification) and reported their confidence.

Each participant was assigned to a specific game and one of several predefined modification conditions. Each participant could only participate once. We used 15 games (cf. Appendix~\ref{appendix:Q2}) with one modification each. 
Assignment of participants to conditions followed a round-robin strategy over available game-modification pairs.

\subsubsection*{Tasks}
Participants were asked to:
\begin{itemize}
    \item Play the game and maximize their score across both original and modified versions, further incentivized with a bonus payment based on performance.
    \item Identify what they believed was the modification.
    \item Rate their confidence in their understanding of the task.
    \item Provide free-text feedback when prompted.
\end{itemize}

\clearpage

\subsection{Implementation Details}

The study was conducted using a custom web-based platform called \textbf{HackTari}, which was built as a demonstration tool for HackAtari modification, enabling real-time human interaction with modified Atari game environments.

The system consisted of the following components:
\begin{itemize}
    \item A \textbf{Flask} backend that served the web interface and managed participant sessions.
    \item \textbf{Socket.IO} was used to support low-latency, bidirectional communication between the browser and game server, allowing responsive gameplay.
    \item Games were instantiated using \textbf{OpenAI Gym} environments, modified with custom wrappers to support HackAtari modifications. 
    \item Each participant was assigned a unique \textbf{token} that mapped to a specific game-modification variant. This ensured that gameplay sessions and logging were isolated per user.
    \item All gameplay logs, including actions, rewards, observations, and survey responses, were stored as session-specific \texttt{.csv} files for later analysis.
\end{itemize}

The full study workflow was hosted using AWS, requiring no software installation by participants. This setup allowed scalable, remote data collection while maintaining consistency across game conditions. Screenshots of the study can be found in Figures~\ref{fig:study_consent} to \ref{fig:post_questionnaire}.

\begin{figure}[h]
    \centering
    \includegraphics[width=\linewidth]{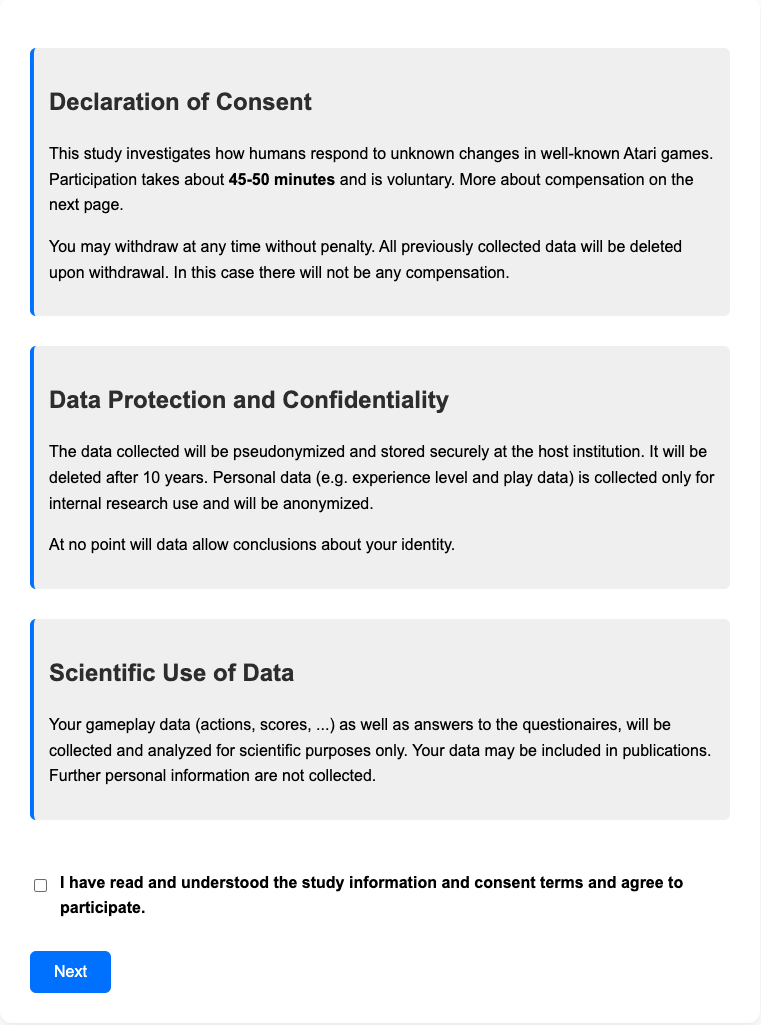}
    \caption{Consent form shown to participants at the beginning of the study. It includes a brief description of the study, use of data, confidentiality statement, and explicit agreement requirement before participation.}

    \label{fig:study_consent}
\end{figure}

\begin{figure}[h]
    \centering
    \includegraphics[width=\linewidth]{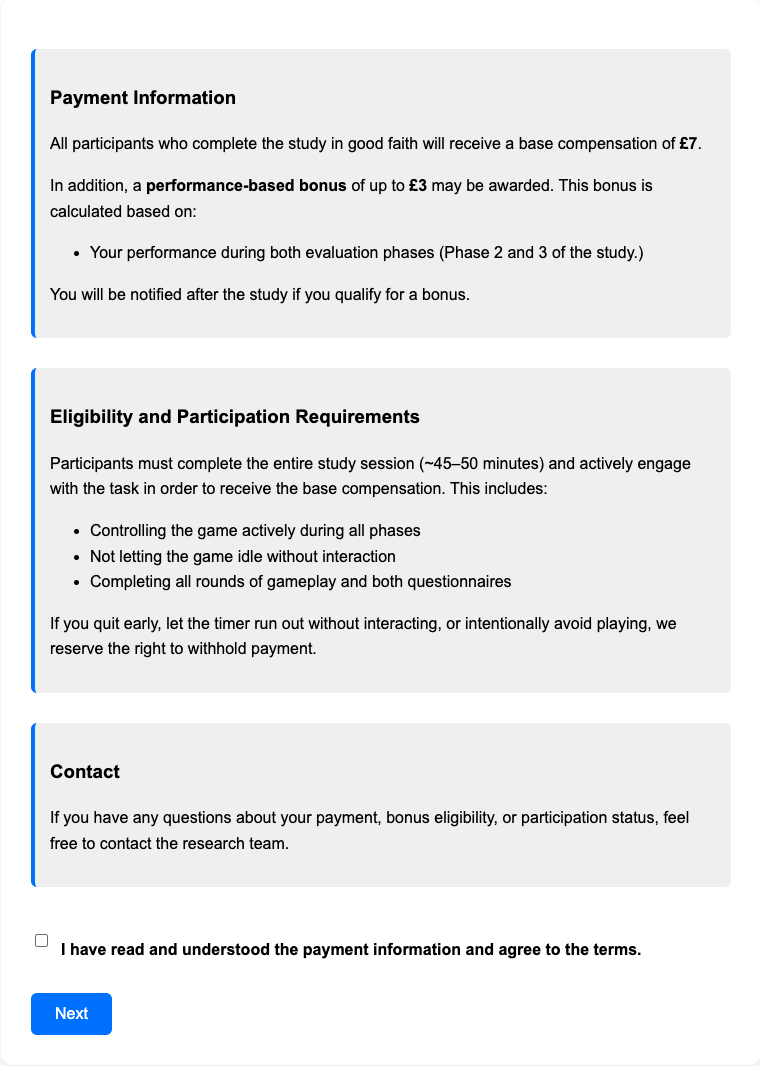}
    \caption{Payment information screen shown before the start of the task. It outlines base compensation and performance-based bonus structure, ensuring transparency around incentives and ethical compensation. Again an explicit agreement was required before paricipation.}
    \label{fig:study_payment}
\end{figure}

\begin{figure}[h]
    \centering
    \includegraphics[width=\linewidth]{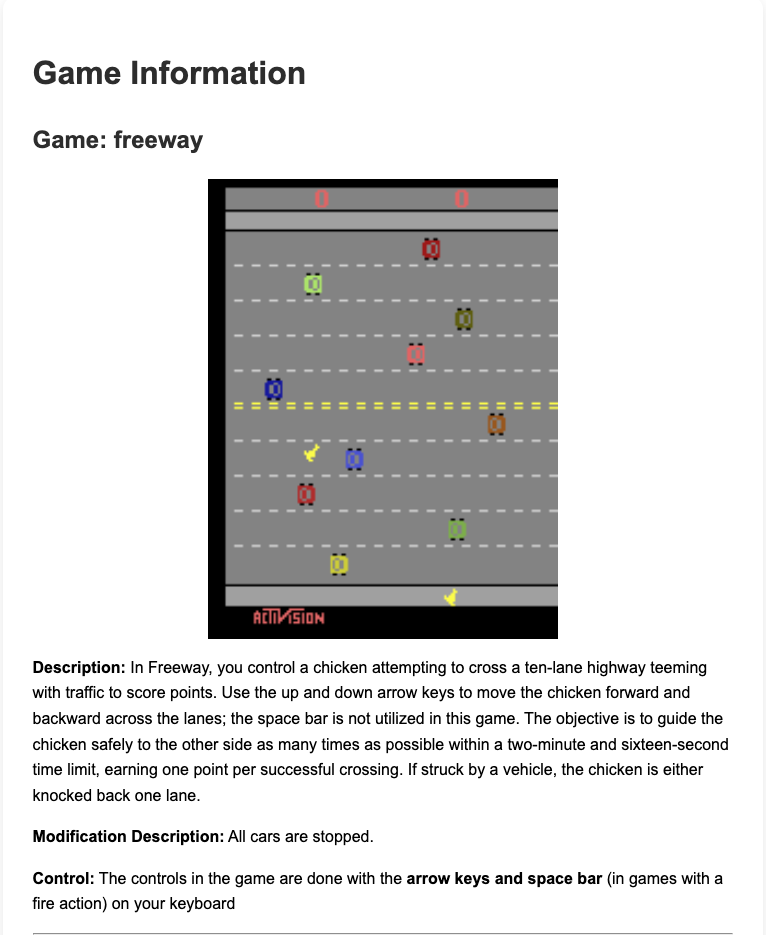}
    \caption{Example game description page shown to participants before gameplay. It provides an overview of the given Atari game, including the objective, scoring mechanics, and available controls.}
    \label{fig:study_game_info}
\end{figure}

\begin{figure}[h]
    \centering
    \includegraphics[width=\linewidth]{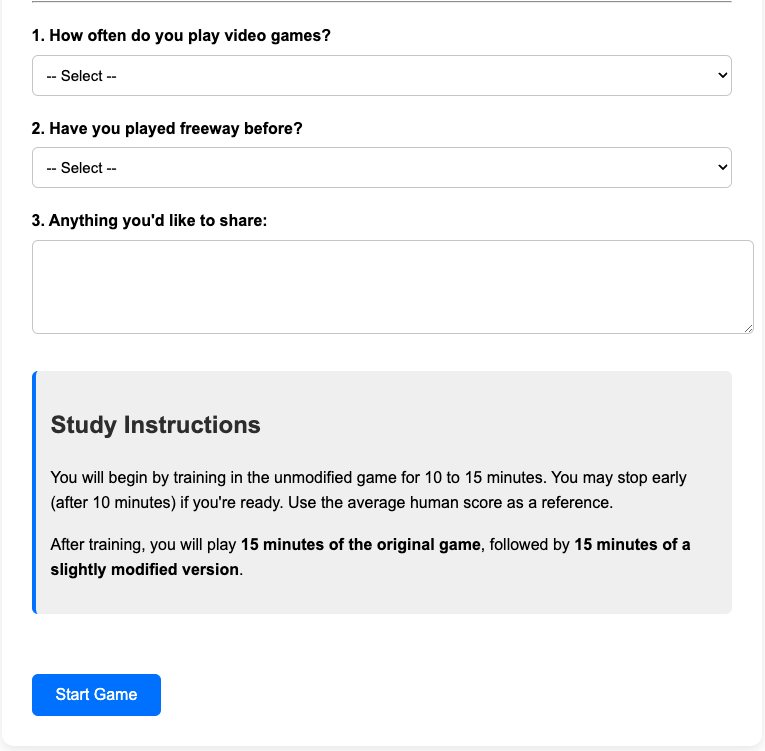}
    \caption{Pre-task questionnaire to get basic information about experience. This could support the quantitative evaluation in the next step. As well as a short reminder of how the study will be run before starting it.}
    \label{fig:study_questionaire1}
\end{figure}

\begin{figure}[h]
    \centering
    \includegraphics[width=\linewidth]{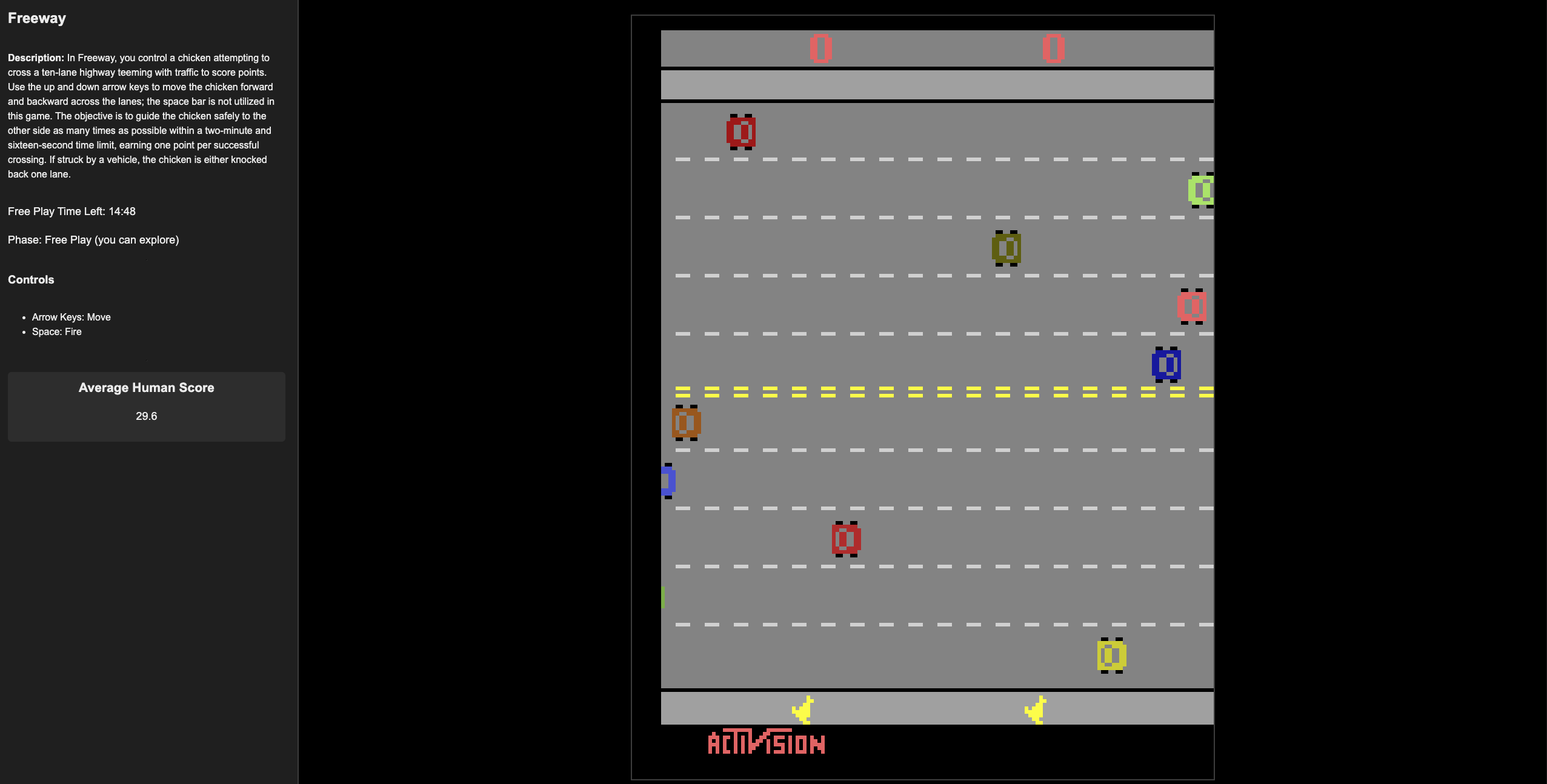}
    \caption{Screenshot of the interactive HackTari gameplay interface. Participants use this interface during Free Training, Evaluation Phase 1 (original game), and Evaluation Phase 2 (modified game). The interface includes the game window, control guide, and timer. In free play the average human score, taken from~\citet{Badia2020agent57}, is shown as reference. }
    \label{fig:study_interface}
\end{figure}

\begin{figure}[h]
    \centering
    \includegraphics[width=\linewidth]{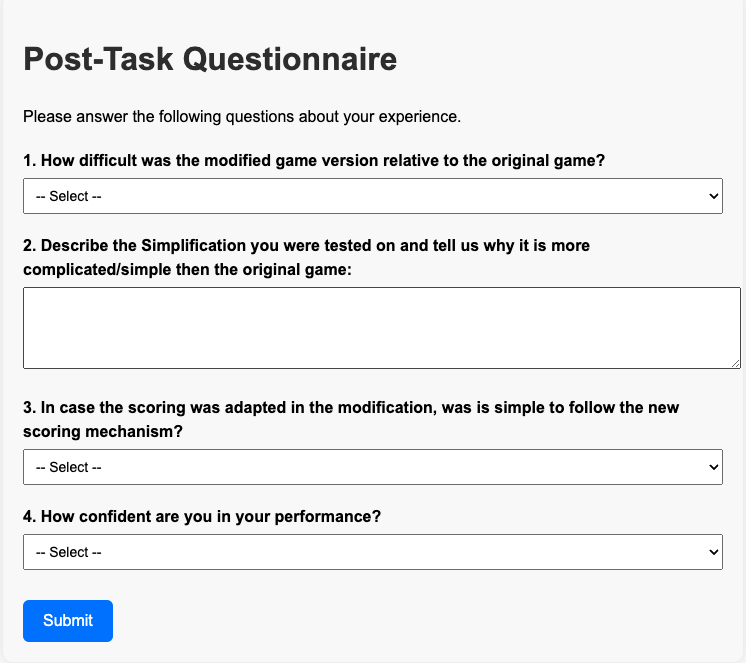}
    \caption{Post-task questionnaire to get feedback.}
    \label{fig:post_questionnaire}
\end{figure}

\subsection{Compensation}
Participants were compensated with a fixed payment of \textsterling7 (equal to \textasciitilde\textsterling8.50 per hour) for completing the study. Further, they were paid a bonus of up to \textsterling3 based on their performance. Importantly, this bonus was not about the performance drop but their actual performance in both phases, Eval1 and Eval2. This was done in accordance with institutional guidelines and ethical research standards.

\subsection{Ethical Considerations}

This study was conducted in accordance with ethical research standards. Participants provided informed consent prior to beginning the experiment. They were informed about the study design, the compensation, and general information about the game they would play. All data was anonymized, and no personally identifiable information was collected. Compensation was provided according to Prolific recommendations. The study was approved by the local ethics board.

\subsection{Results}
\label{appendix:human_results_figures}
Participants were excluded from our data analysis if they did not complete all phases of the experiment, if it was clearly apparent that they stopped actively playing the game, or if they reported technical difficulties.
Therefore, our final dataset includes \textbf{128} participants.
We analyzed the quantitative performance metrics across the original and modified game conditions. Findings can be seen in Figure~\ref{fig:human_analysis1} to \ref{fig:human_analysis3}.

\begin{figure}
    \centering
    \makebox[\textwidth][c]{\includegraphics[width=1.2\textwidth]{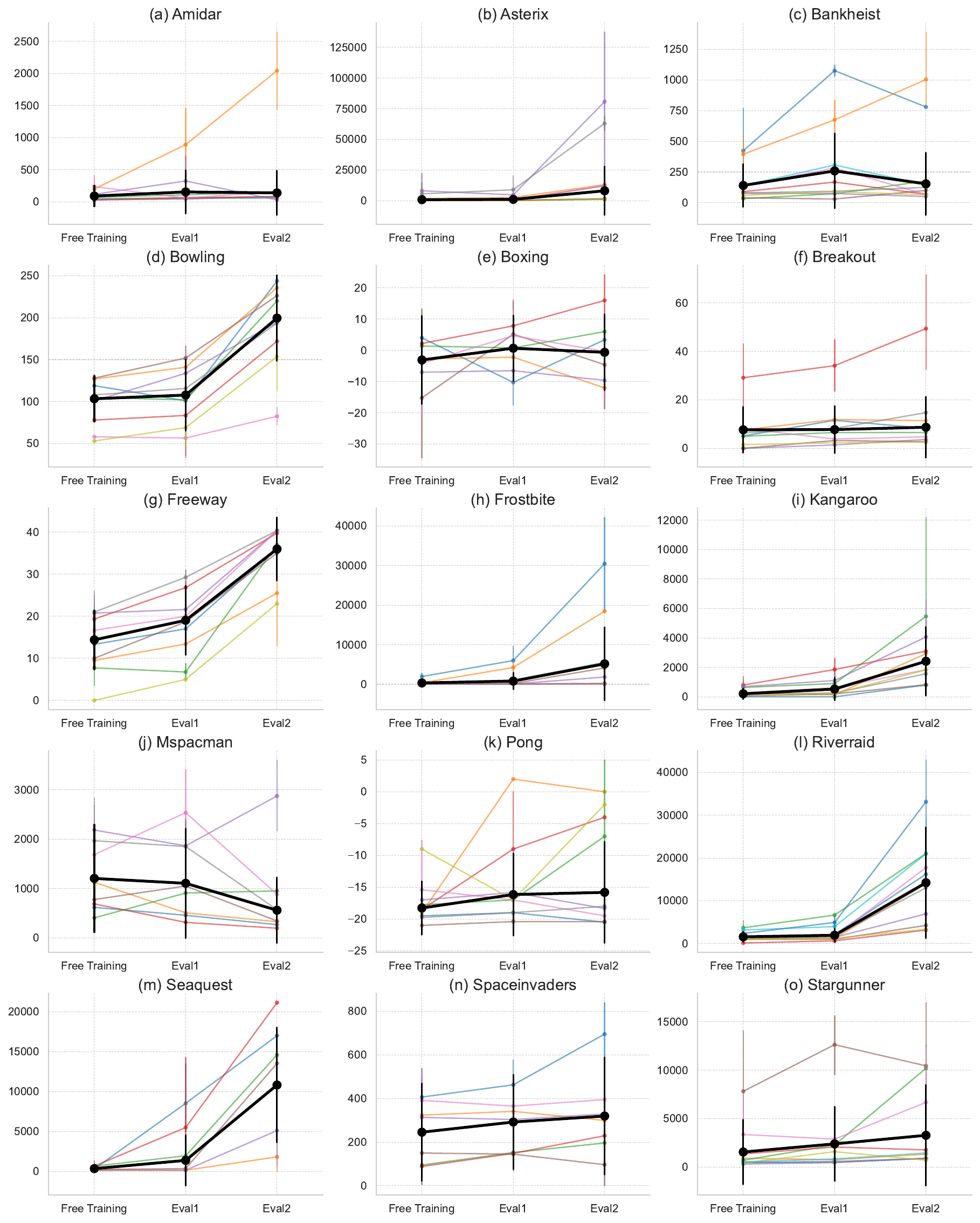}
    }
    \caption{Raw human scores across three phases: Free Training, Evaluation on the original task (Eval1), and Evaluation on the modified task (Eval2). Each subplot shows mean scores across all participants for a specific game. Most participants improved or maintained performance on Eval2, supporting the claim that the selected HackAtari variants are true task simplifications.}

    \label{fig:human_analysis1}
\end{figure}

\begin{figure}
    \centering
    \makebox[\textwidth][c]{\includegraphics[width=1.2\textwidth]{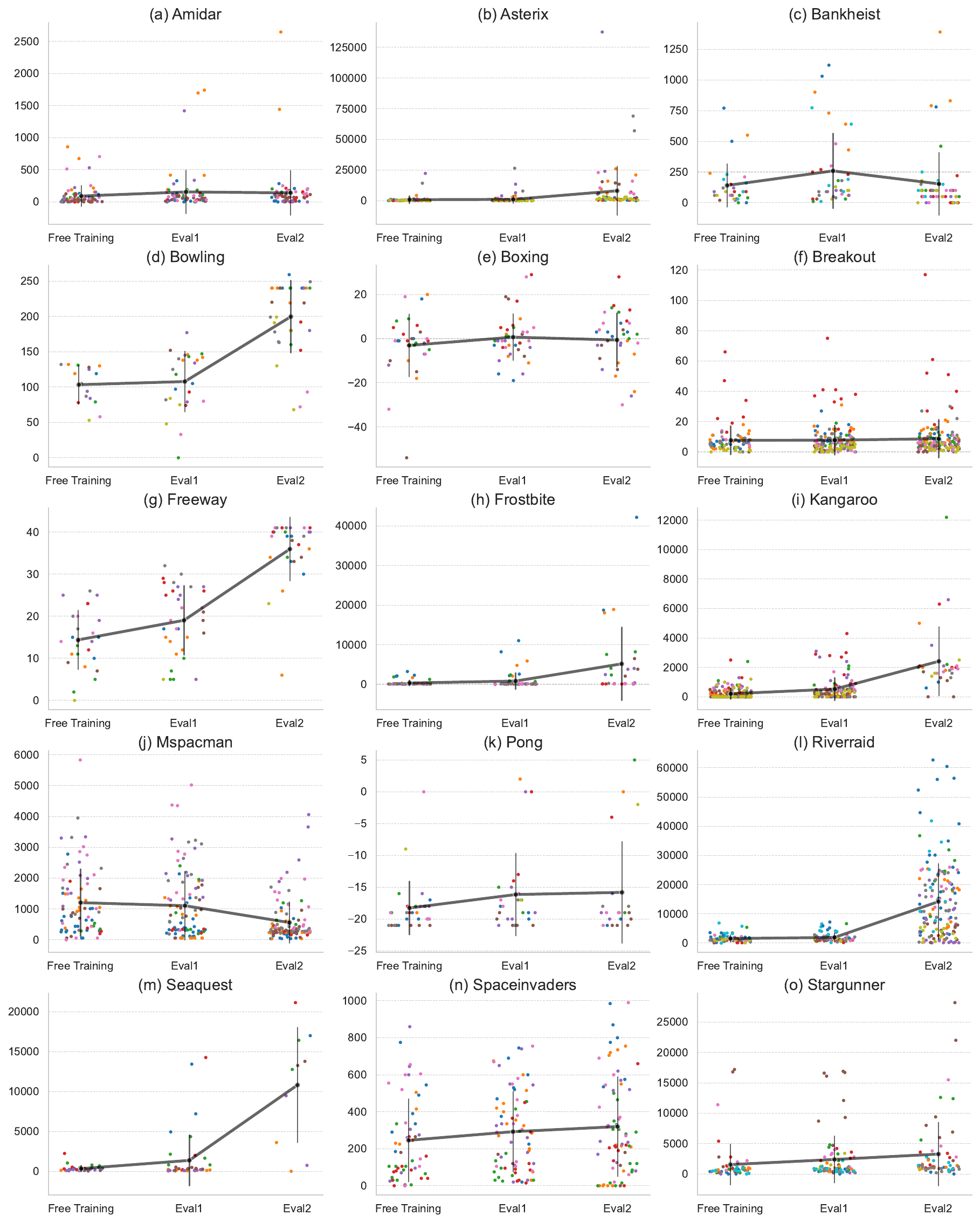}
    }
    \caption{Visualization of all results of all users.}

    \label{fig:human_analysis2}
\end{figure}

\begin{figure}
    \centering
    \makebox[\textwidth][c]{\includegraphics[width=1.2\textwidth]{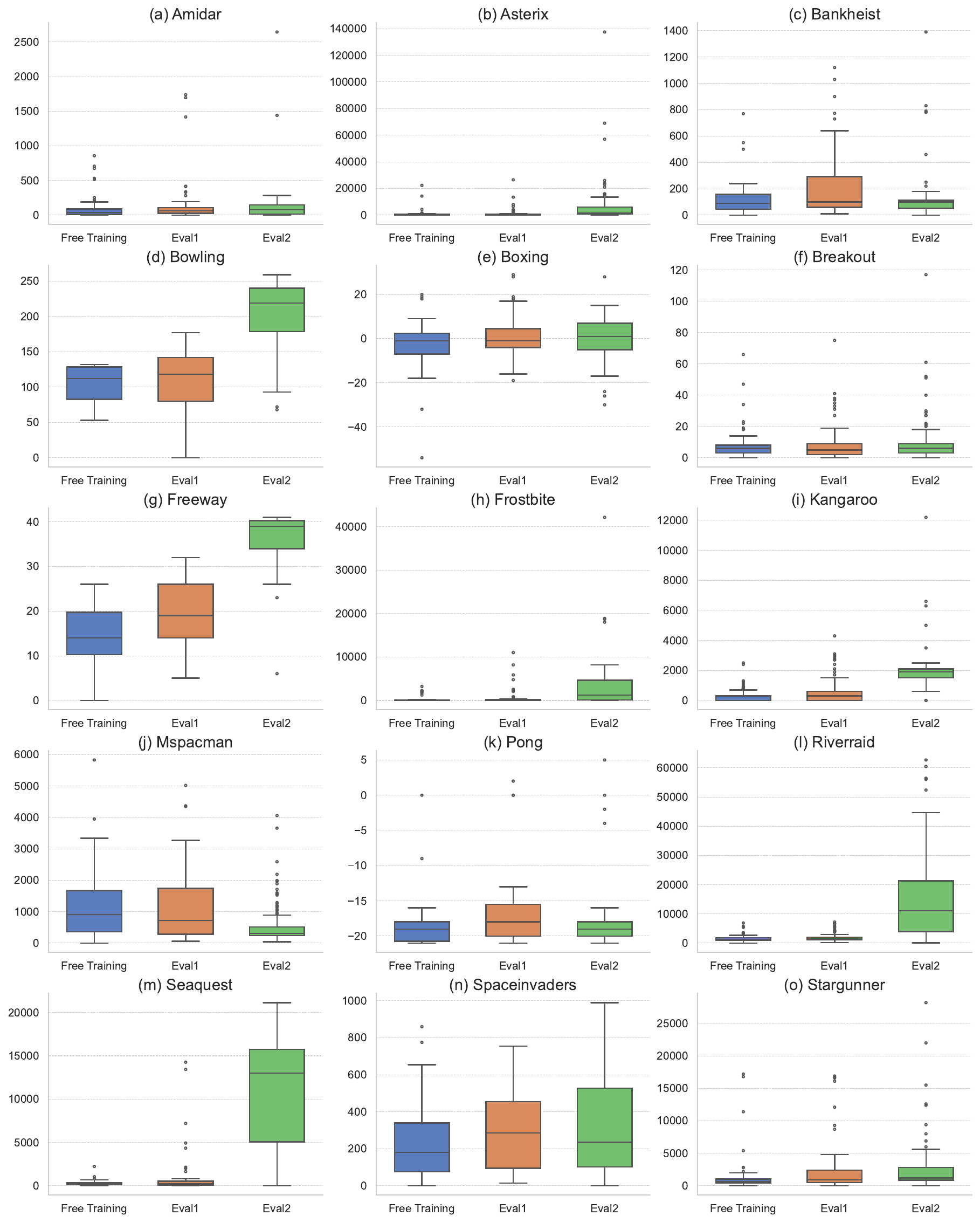}
    }
    \caption{This presentation emphasizes the robustness of human generalization to simplified modifications.}

    \label{fig:human_analysis3}
\end{figure}

\clearpage

\section{Game Descriptions and Modifications}
\label{appendix:environments}

  \subsection{Alien}
  \noindent
  \IfFileExists{environment_description/Alien.tex}{%
    \begin{minipage}{0.80\textwidth}
     \textbf{Description:} \\
      You are stuck in a maze-like space ship with three aliens. You goal is to destroy their eggs that are scattered all over the ship while simultaneously avoiding the aliens (they are trying to kill you). You have a flamethrower that can help you turn them away in tricky situations. Moreover, you can occasionally collect a power-up (pulsar) that gives you the temporary ability to kill aliens.
    \end{minipage}
  }{}
\IfFileExists{environment_images_new_pdfs/Alien.pdf}{%
    \begin{minipage}{0.18\textwidth}
      \raggedleft
      \includegraphics[scale=0.10]{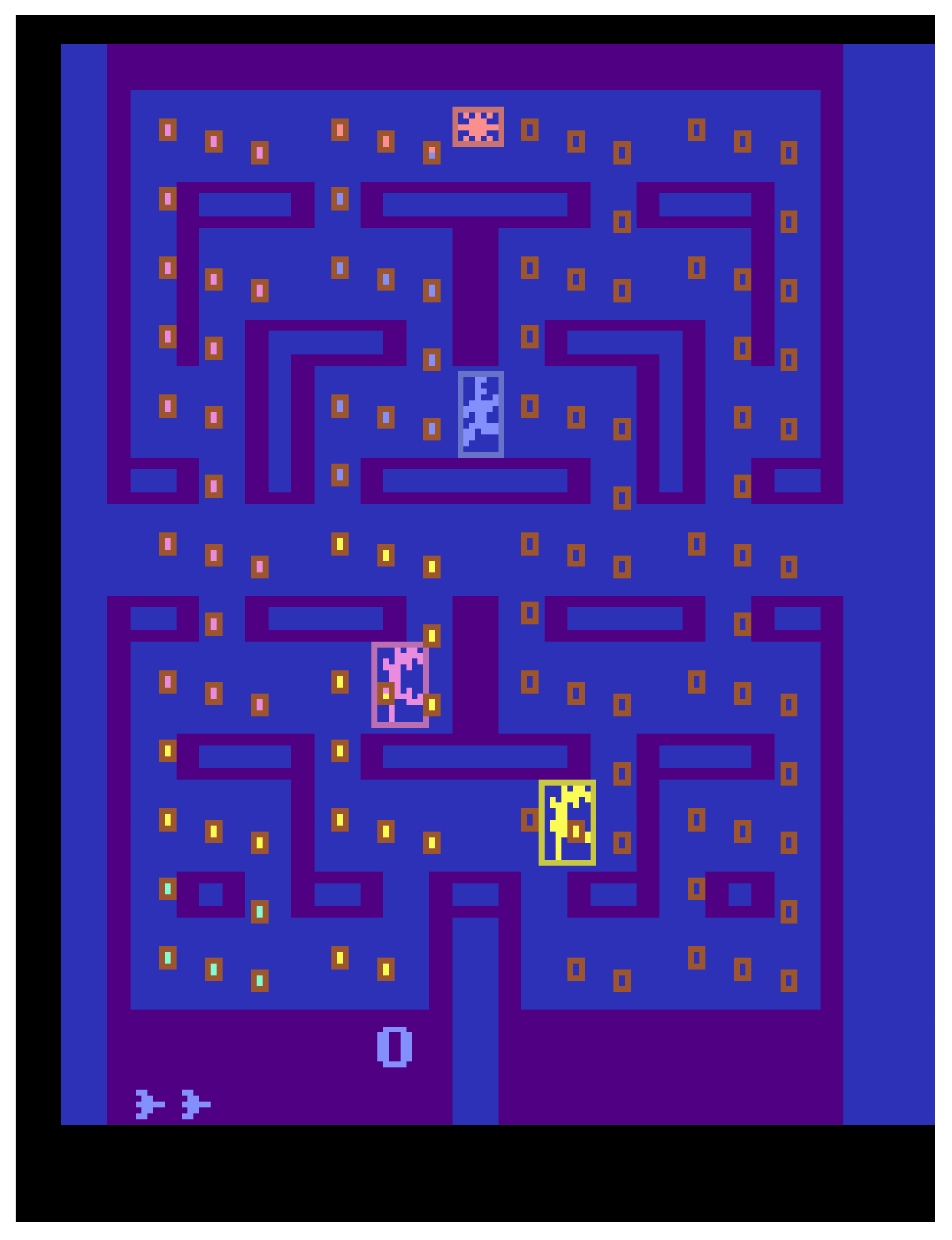}
    \end{minipage}
  }{}
  \IfFileExists{game_modifs/Alien.tex}{%
      \begin{table}[h]
\centering
\begin{tabular}{p{3cm} | p{10cm}} \toprule
        \textbf{Modification} & \textbf{Effect} \\ \midrule
        last\_egg & Removes all eggs but one. \\ \bottomrule
    \end{tabular}
\end{table}
  }{}
  
  \vspace{-1em} 

  \subsection{Amidar}
  \noindent
  \IfFileExists{environment_description/Amidar.tex}{%
    \begin{minipage}{0.80\textwidth}
     \textbf{Description:} \\
      This game is similar to Pac-Man: You are trying to visit all places on a 2-dimensional grid while simultaneously avoiding your enemies. You can turn the tables at one point in the game: Your enemies turn into chickens and you can catch them.
    \end{minipage}
  }{}
\IfFileExists{environment_images_new_pdfs/Amidar.pdf}{%
    \begin{minipage}{0.18\textwidth}
      \raggedleft
      \includegraphics[scale=0.10]{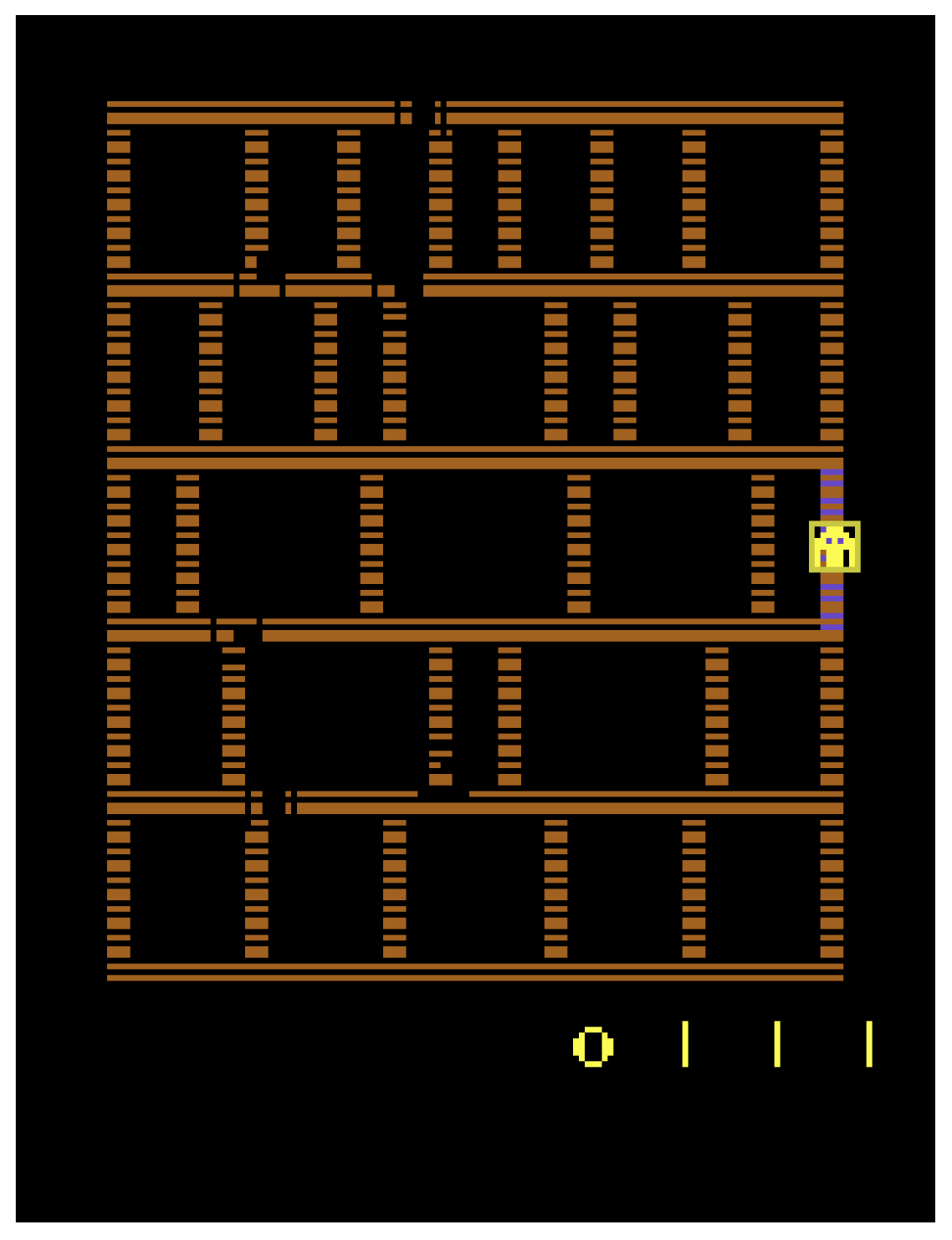}
    \end{minipage}
  }{}
  \IfFileExists{game_modifs/Amidar.tex}{%
      \begin{table}[h]
    \centering
    \begin{tabular}{p{3cm} | p{10cm}} \toprule
        \textbf{Modification} & \textbf{Effect} \\ \midrule
        pig\_enemies & Replaces all enemies with the pig enemies. \\
        paint\_roller\_player & Changes the player to the paint roller character. \\
        unlimited\_lives & Player has an unlimited amounts of lives. \\ \bottomrule
    \end{tabular}
\end{table}
  }{}
  
  \vspace{-1em} 

  \subsection{Asterix}
  \noindent
  \IfFileExists{environment_description/Asterix.tex}{%
    \begin{minipage}{0.80\textwidth}
     \textbf{Description:} \\
      You are Asterix and can move horizontally (continuously) and vertically (discretely). Objects move horizontally across the screen: lyres and other (more useful) objects. Your goal is to guide Asterix in such a way as to avoid lyres and collect as many other objects as possible. You score points by collecting objects and lose a life whenever you collect a lyre. You have three lives available at the beginning. If you score sufficiently many points, you will be awarded additional points.
    \end{minipage}
  }{}
\IfFileExists{environment_images_new_pdfs/Asterix.pdf}{%
    \begin{minipage}{0.18\textwidth}
      \raggedleft
      \includegraphics[scale=0.10]{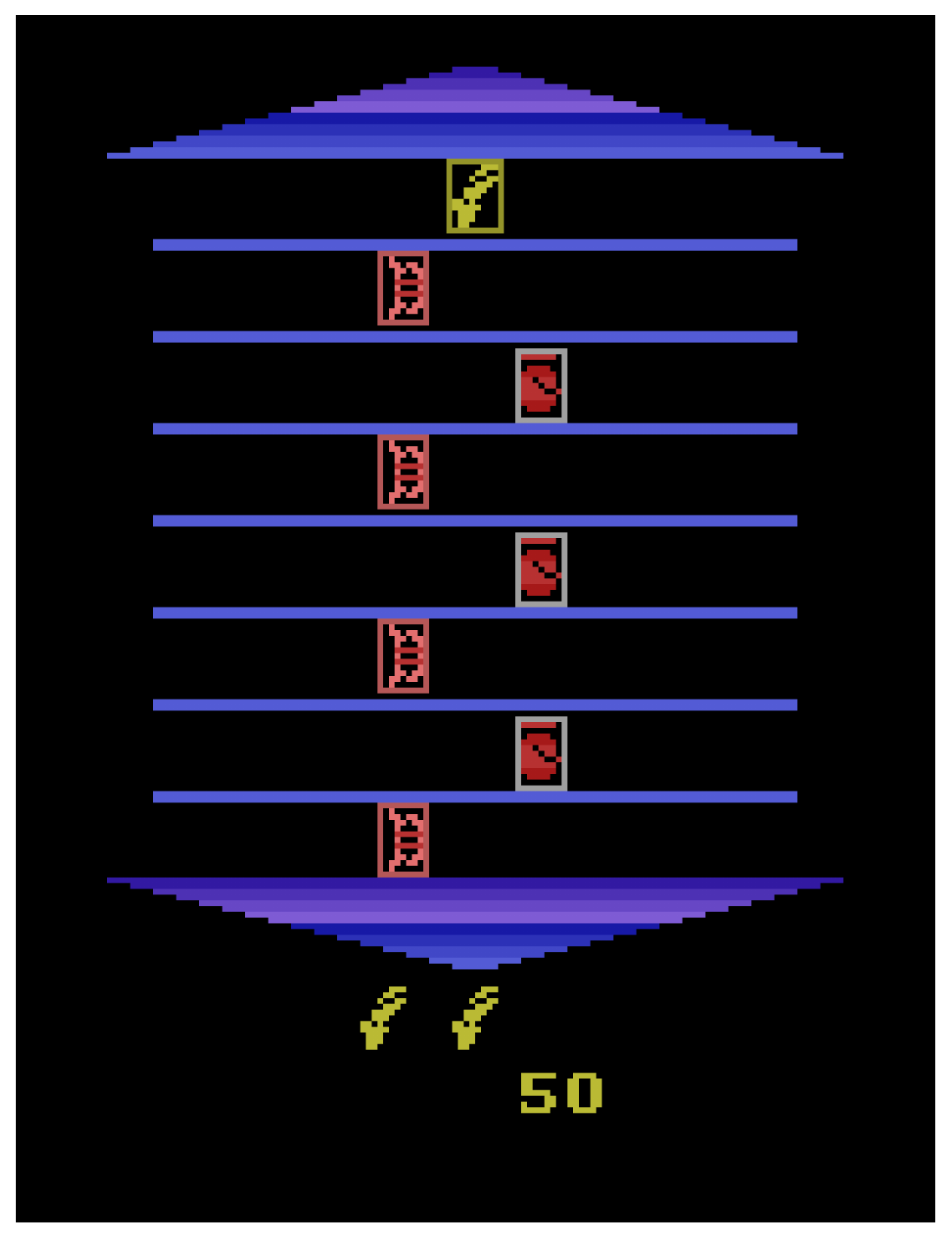}
    \end{minipage}
  }{}
  \IfFileExists{game_modifs/Asterix.tex}{%
      \begin{table}[h]
    \centering
    \begin{tabular}{p{3cm} | p{10cm}} \toprule
        \textbf{Modification} & \textbf{Effect} \\ \midrule
        obelix & Changes the playable character to obelix. \\
        set\_consumable\_1 & Changes all consumables to pink consumables (100points). \\
        set\_consumable\_2 & Changes all consumables to shields (200points). \\
        unlimited\_lives & Player has an unlimited amounts of lives. \\
        even\_lines\_free & Clears the even numbered lines (1 being the top most line). \\
        odd\_lines\_free & Clears the odd numbered lines (1 being the top most line). \\ \bottomrule
    \end{tabular}
\end{table}
  }{}
  
  \vspace{-1em} 

  \subsection{Atlantis}
  \noindent
  \IfFileExists{environment_description/Atlantis.tex}{%
    \begin{minipage}{0.80\textwidth}
     \textbf{Description:} \\
      Your job is to defend the submerged city of Atlantis. Your enemies slowly descend towards the city and you must destroy them before they reach striking distance. To this end, you control turrets stationed on three defense posts. You lose if your enemies manage to destroy all seven of Atlantis’ installations. You may rebuild installations after you have fought of a wave of enemies and scored a sufficient number of points.
    \end{minipage}
  }{}
\IfFileExists{environment_images_new_pdfs/Atlantis.pdf}{%
    \begin{minipage}{0.18\textwidth}
      \raggedleft
      \includegraphics[scale=0.10]{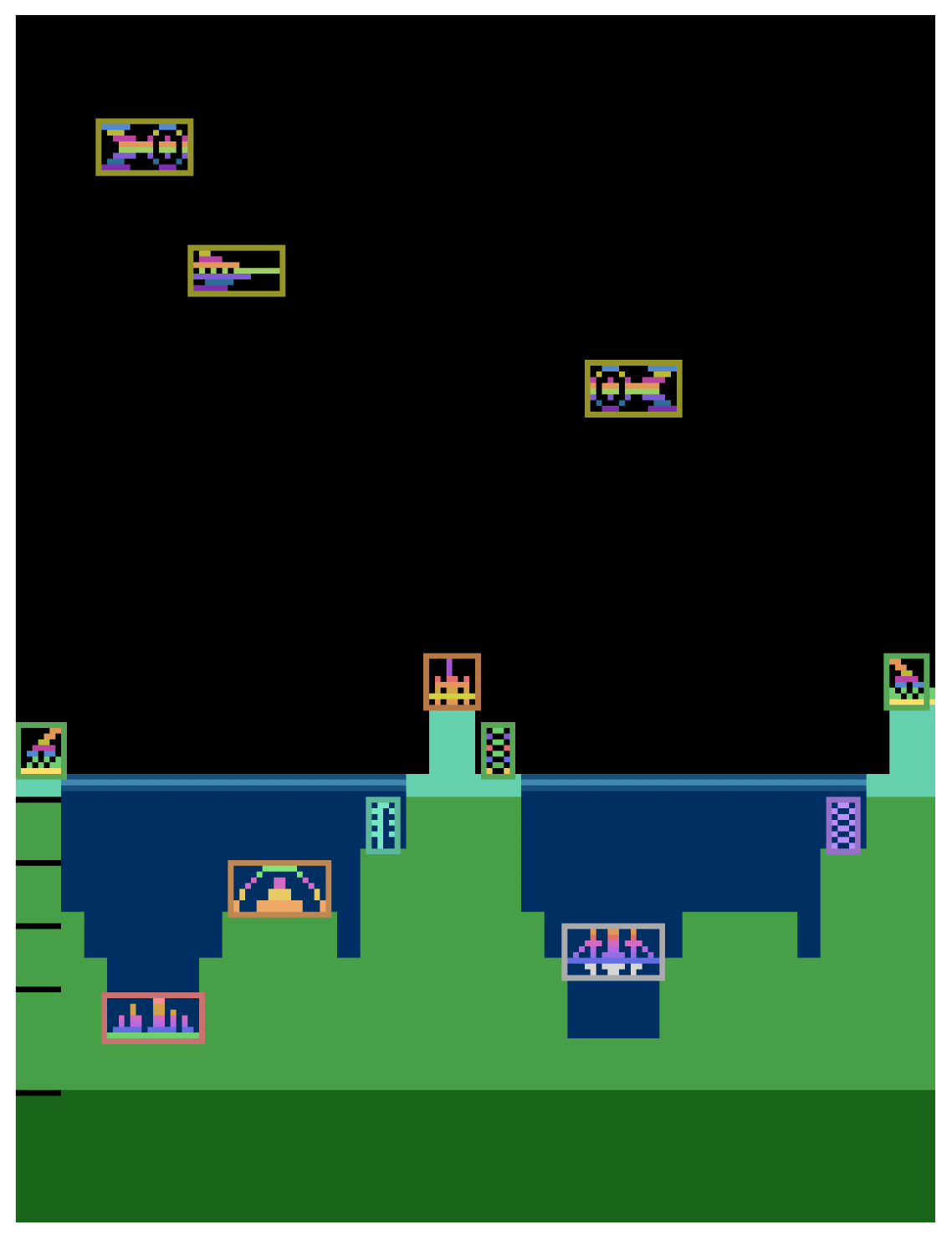}
    \end{minipage}
  }{}
  \IfFileExists{game_modifs/Atlantis.tex}{%
      \begin{table}[h]
    \centering
    \begin{tabular}{p{3.5cm} | p{9.5cm}} \toprule
        \textbf{Modification} & \textbf{Effect} \\ \midrule
        no\_last\_line & Removes enemies from the lowest line. \\
        jets\_only & All enemy ships are jets. \\
        random\_enemies & Randomly assigns enemy types, instead of following the standardized pattern. \\
        speed\_mode\_slow & Sets enemy speed to low. \\
        speed\_mode\_medium & Sets the enemy speed to medium. \\
        speed\_mode\_fast & Sets the enemy speed to fast. \\
        speed\_mode\_ultrafast & Sets the enemy speed to ultrafast. \\ \bottomrule
    \end{tabular}
\end{table}
  }{}
  
  \vspace{-1em} 

  \subsection{BankHeist}
  \noindent
  \IfFileExists{environment_description/BankHeist.tex}{%
    \begin{minipage}{0.80\textwidth}
     \textbf{Description:} \\
      You are a bank robber and (naturally) want to rob as many banks as possible. You control your getaway car and must navigate maze-like cities. The police chases you and will appear whenever you rob a bank. You may destroy police cars by dropping sticks of dynamite. You can fill up your gas tank by entering a new city. At the beginning of the game you have four lives. Lives are lost if you run out of gas, are caught by the police, or run over some dynamite you have previously dropped.
    \end{minipage}
  }{}
\IfFileExists{environment_images_new_pdfs/BankHeist.pdf}{%
    \begin{minipage}{0.18\textwidth}
      \raggedleft
      \includegraphics[scale=0.10]{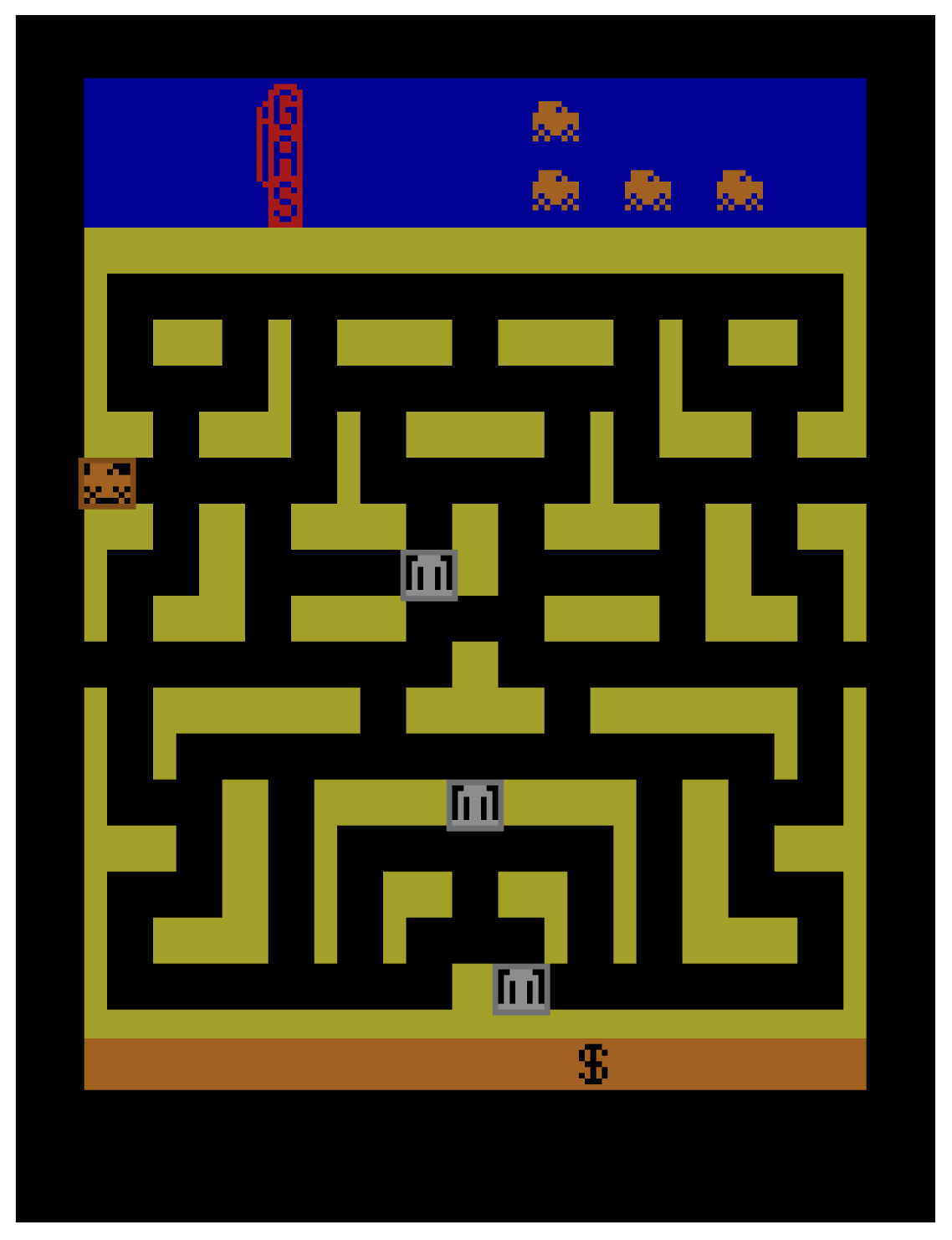}
    \end{minipage}
  }{}
  \IfFileExists{game_modifs/BankHeist.tex}{%
      \begin{table}[h]
    \centering
    \begin{tabular}{p{3cm} | p{10cm}} \toprule
        \textbf{Modification} & \textbf{Effect} \\ \midrule
        unlimited\_gas & Unlimited gas for the player. \\
        no\_police & Removes police from the game.\\
        only\_police & No banks only police. \\
        two\_police\_cars & Replaces 2 banks with police cars and robbed banks give 50 points. \\
        random\_city & Randomizes which city is entered next. \\
        revisit\_city & Allows player to go back one city. \\ \bottomrule
    \end{tabular}
\end{table}
  }{}
  
  \vspace{-1em} 

  \subsection{Bowling}
  \noindent
  \IfFileExists{environment_description/Bowling.tex}{%
    \begin{minipage}{0.80\textwidth}
     \textbf{Description:} \\
      Your goal is to score as many points as possible in the game of Bowling. A game consists of 10 frames and you have two tries per frame. Knocking down all pins on the first try is called a “strike”. Knocking down all pins on the second roll is called a “spar”. Otherwise, the frame is called “open”.
    \end{minipage}
  }{}
\IfFileExists{environment_images_new_pdfs/Bowling.pdf}{%
    \begin{minipage}{0.18\textwidth}
      \raggedleft
      \includegraphics[scale=0.10]{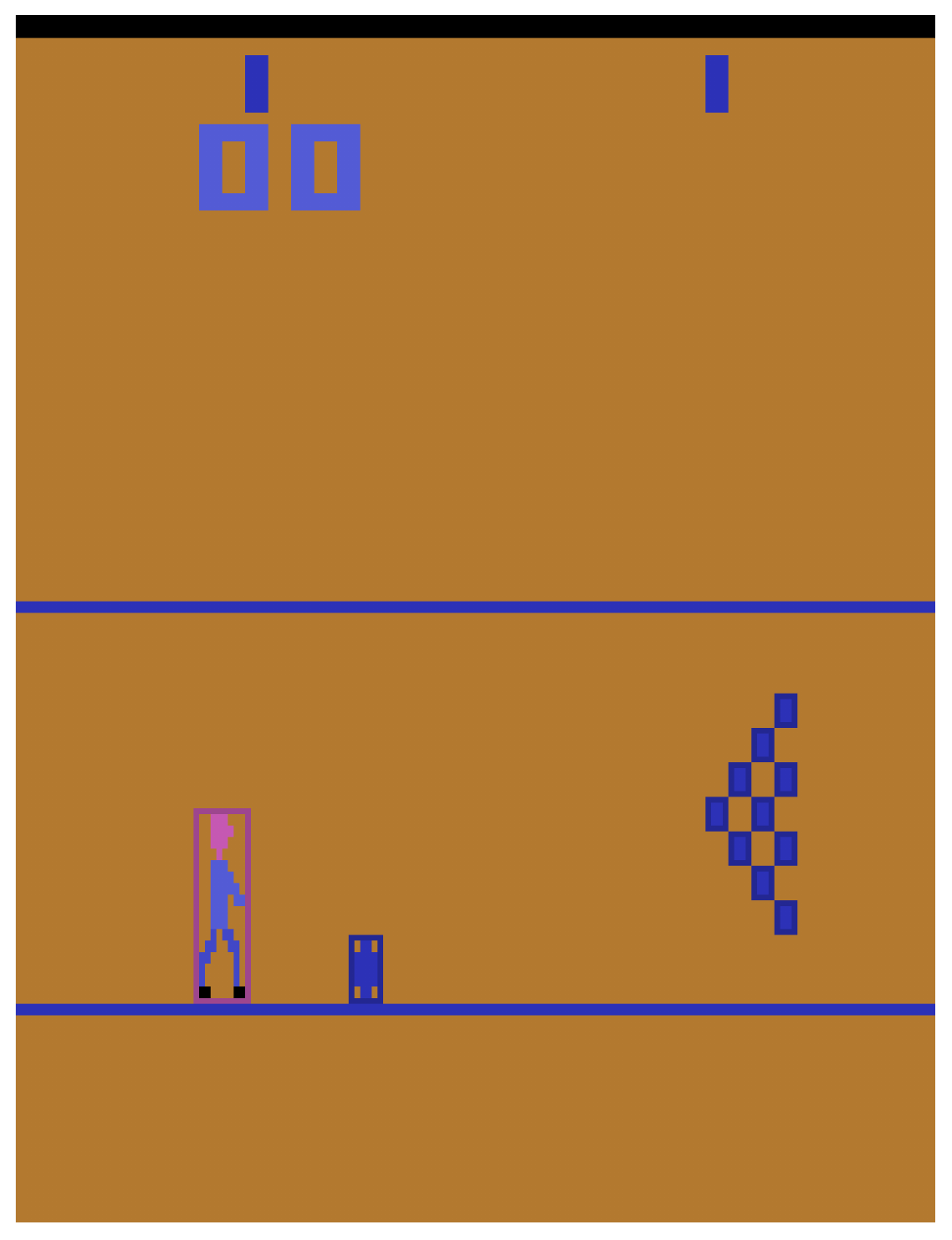}
    \end{minipage}
  }{}
  \IfFileExists{game_modifs/Bowling.tex}{%
      \begin{table}[h]
    \centering
    \begin{tabular}{p{3cm} | p{10cm}} \toprule
        \textbf{Modification} & \textbf{Effect} \\ \midrule
        shift\_player & Shifts the player to the right. \\
        horizontal\_pins & Draws the pins horizontally instead of vertically. \\
        small\_pins & Decreases the pin size from 2 pixels to 1 pixel. \\
        moving\_pins & Moves the pins up and down. \\
        top\_pins & The top two pins are in-game. \\
        middle\_pins & Two middle pins are in-game. \\
        bottom\_pins & The bottom two pins are in-game. \\ \bottomrule
    \end{tabular}
\end{table}
  }{}
  
  \vspace{-1em} 

  \subsection{Boxing}
  \noindent
  \IfFileExists{environment_description/Boxing.tex}{%
    \begin{minipage}{0.80\textwidth}
     \textbf{Description:} \\
      You fight an opponent in a boxing ring. You score points for hitting the opponent. If you score 100 points, your opponent is knocked out.
    \end{minipage}
  }{}
\IfFileExists{environment_images_new_pdfs/Boxing.pdf}{%
    \begin{minipage}{0.18\textwidth}
      \raggedleft
      \includegraphics[scale=0.10]{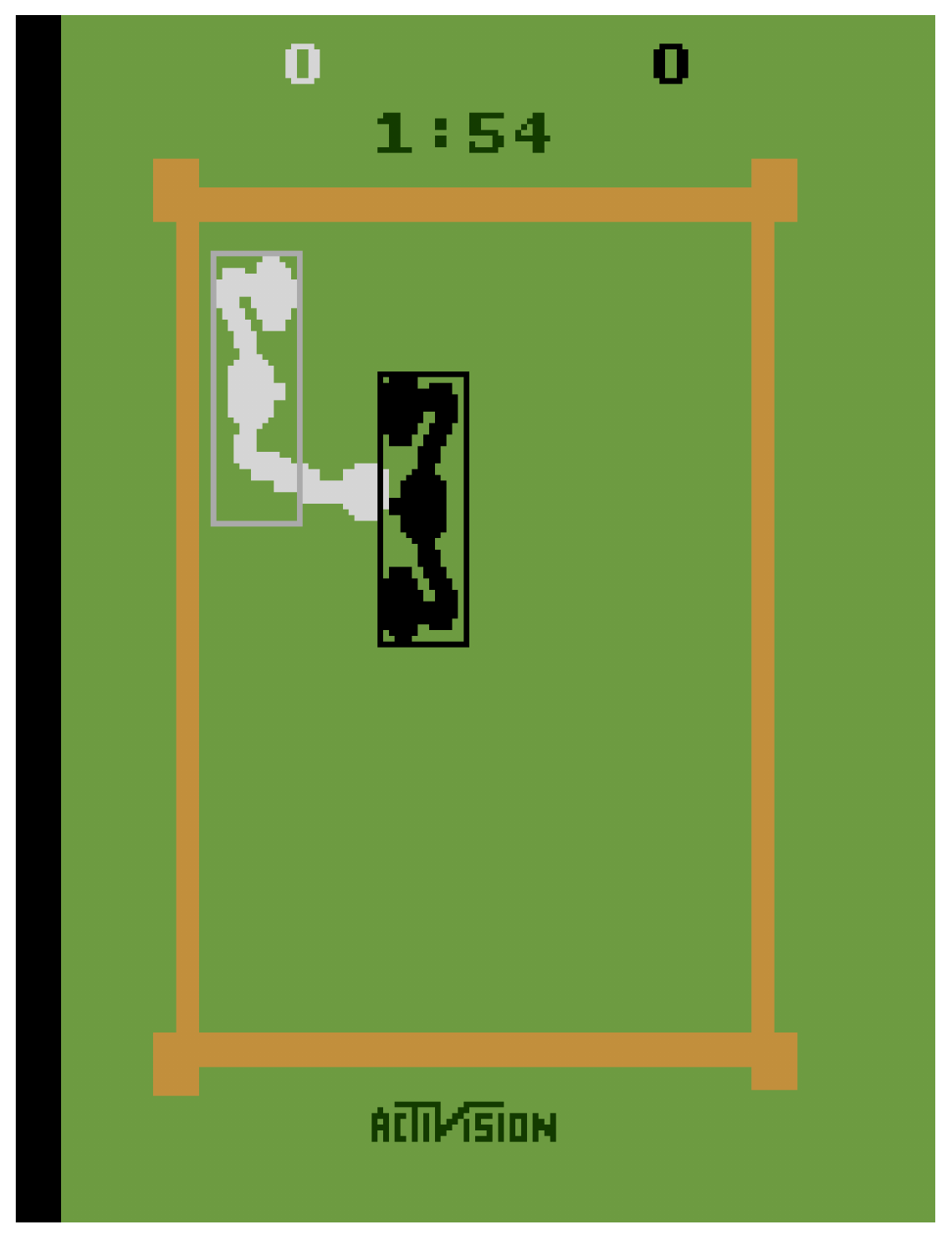}
    \end{minipage}
  }{}
  \IfFileExists{game_modifs/Boxing.tex}{%
      \begin{table}[h]
    \centering
    \begin{tabular}{p{3cm} | p{10cm}} \toprule
        \textbf{Modification} & \textbf{Effect} \\ \midrule
        gravity & Adds a permanent downwards movement. \\
        antigravity & Adds a permanent upwards movement. \\
        offensive & Moves the player character forward in the game environment. \\
        defensive & Moves the player character backward in the game environment. \\
        down & Moves the player character down in the game environment. \\
        one\_armed &  Disables the "hitting motion" with the right arm permanently \\
        drunken\_boxing & Applies random movements to the players input. \\
        color\_player\_$X$ &  Changes the color of the player to color $X \in$ [Black, White, Red, Blue, Green]\\
        color\_enemy\_$X$ &  Changes the color of the enemy to color $X \in$ [Black, White, Red, Blue, Green] by choosing a value 0-4 \\
        switch\_positions & Switches the position of player and enemy \\
        classic\_colors & Using a red player and a blue enemy \\
    \bottomrule
    \end{tabular}
\end{table}
  }{}
  
  \vspace{-1em} 

  \subsection{Breakout}
  \noindent
  \IfFileExists{environment_description/Breakout.tex}{%
    \begin{minipage}{0.80\textwidth}
     \textbf{Description:} \\
      Another famous Atari game. The dynamics are similar to pong: You move a paddle and hit the ball at a brick wall at the top of the screen. Your goal is to destroy the brick wall. You can try to break through the wall and let the ball wreak havoc on the other side, all on its own! If the ball falls below your padle you lose one of your five lives.
    \end{minipage}
  }{}
\IfFileExists{environment_images_new_pdfs/Breakout.pdf}{%
    \begin{minipage}{0.18\textwidth}
      \raggedleft
      \includegraphics[scale=0.10]{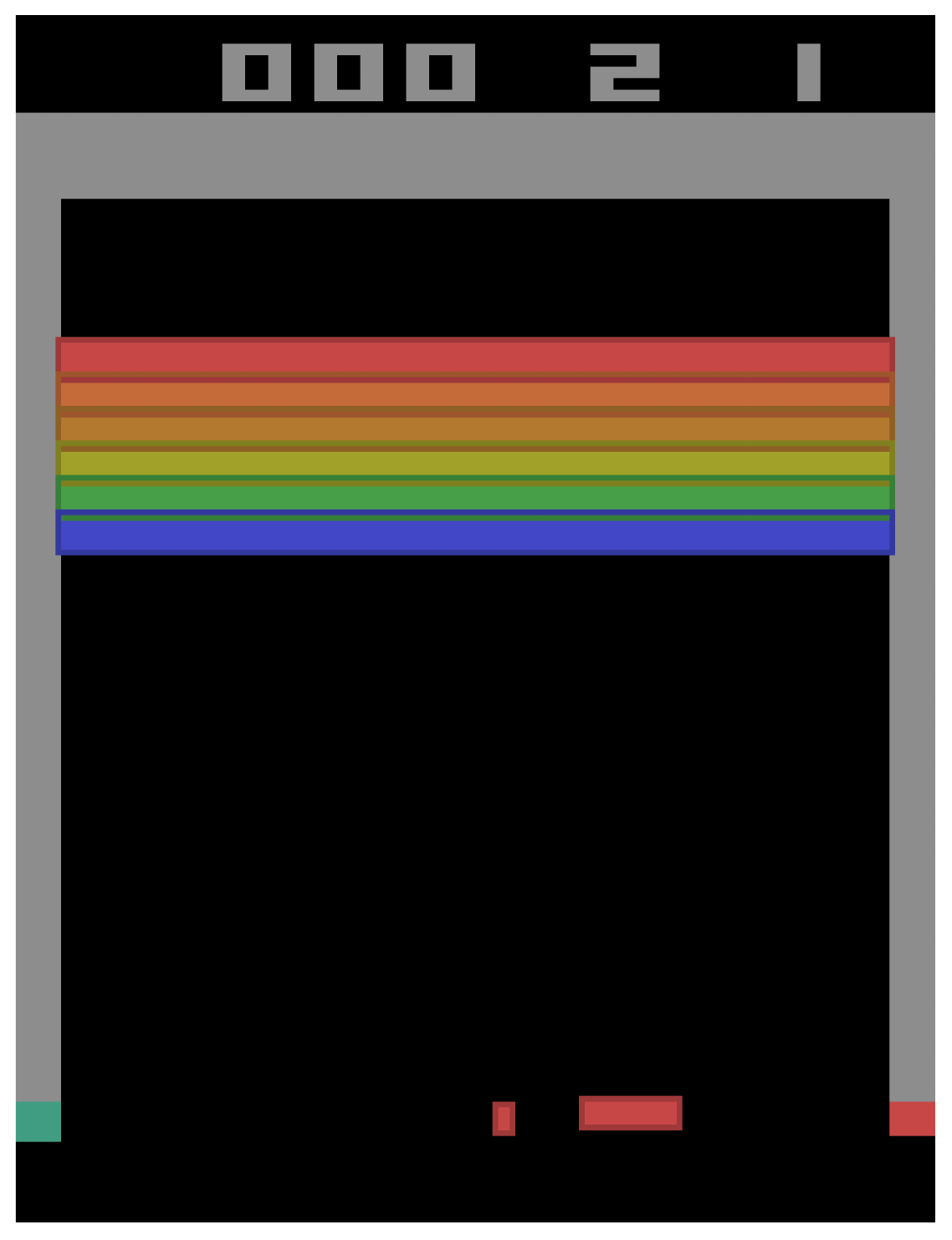}
    \end{minipage}
  }{}
  \IfFileExists{game_modifs/Breakout.tex}{%
      \begin{table}[h]
    \centering
    \begin{tabular}{p{3.75cm} | p{9.25cm}} \toprule
        \textbf{Modification} & \textbf{Effect} \\ \midrule
        right\_drift & Applies a right drift to the ball.\\
        left\_drift & Applies a left drift to the ball.\\
        gravity & Applies a downward force to the ball\\
        inverse\_gravity & Applies an upward force to the ball \\
        color\_player\_and\_ball\_$X$ &  Changes the color of the player to color $X \in$ [Black, White, Red, Blue, Green].\\
        color\_all\_blocks\_$X$ & Changes the color of all blocks to color $X \in$ [Black, White, Red, Blue, Green]. \\
    \bottomrule
    \end{tabular}
\end{table}
  }{}
  
  \vspace{-1em} 

  \subsection{Carnival}
  \noindent
  \IfFileExists{environment_description/Carnival.tex}{%
    \begin{minipage}{0.80\textwidth}
     \textbf{Description:} \\
      This is a “shoot ‘em up” game. Targets move horizontally across the screen and you must shoot them. You control the gun at the bottom of the screen that can be moved horizontally. The supply of ammunition is limited and chickens may steal some bullets from you if you don’t hit them in time.
    \end{minipage}
  }{}
\IfFileExists{environment_images_new_pdfs/Carnival.pdf}{%
    \begin{minipage}{0.18\textwidth}
      \raggedleft
      \includegraphics[scale=0.10]{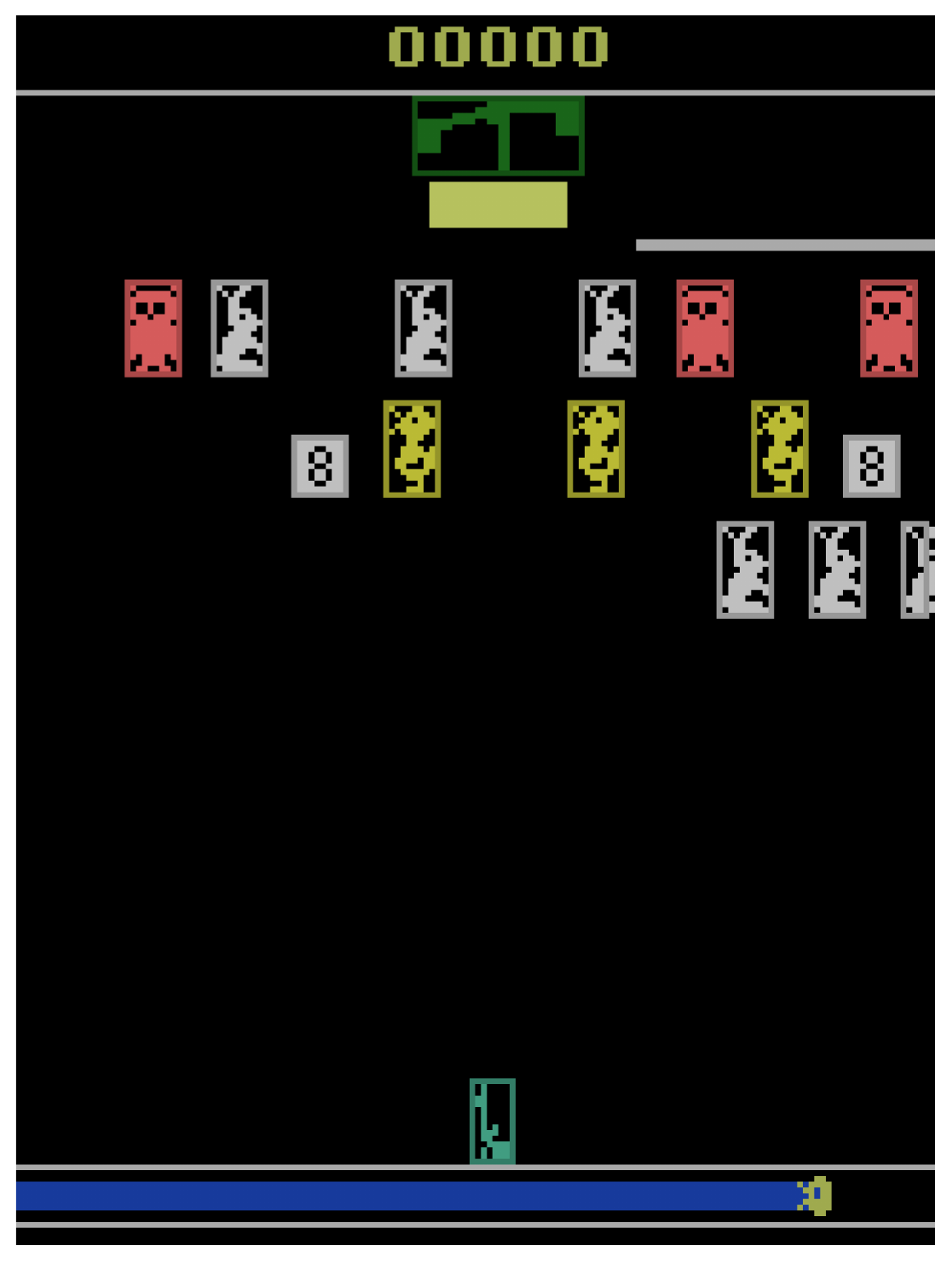}
    \end{minipage}
  }{}
  \IfFileExists{game_modifs/Carnival.tex}{%
      \begin{table}[h]
    \centering
    \begin{tabular}{p{5cm} |p{8cm}} \toprule
        \textbf{Modification} & \textbf{Effect} \\ \midrule
        no\_flying\_ducks & Ducks in the last row disappear instead of turning into flying ducks. \\
        unlimited\_ammo & Ammunition doesn't decrease. \\
        missile\_speed\_small\_increase & The projectiles fired from the players are slightly faster.\\
        missile\_speed\_medium\_increase & The projectiles fired from the players are notably faster.\\
        missile\_speed\_faster\_increase & The projectiles fired from the players are a lot faster.\\
    \bottomrule
    \end{tabular}
\end{table}
  }{}
  
  \vspace{-1em} 

  \subsection{ChopperCommand}
  \noindent
  \IfFileExists{environment_description/ChopperCommand.tex}{%
    \begin{minipage}{0.80\textwidth}
     \textbf{Description:} \\
      You control a helicopter and must protect truck convoys. To that end, you need to shoot down all enemy aircraft. A mini-map is displayed at the bottom of the screen that shows all truck and aircraft positions.
    \end{minipage}
  }{}
\IfFileExists{environment_images_new_pdfs/ChopperCommand.pdf}{%
    \begin{minipage}{0.18\textwidth}
      \raggedleft
      \includegraphics[scale=0.10]{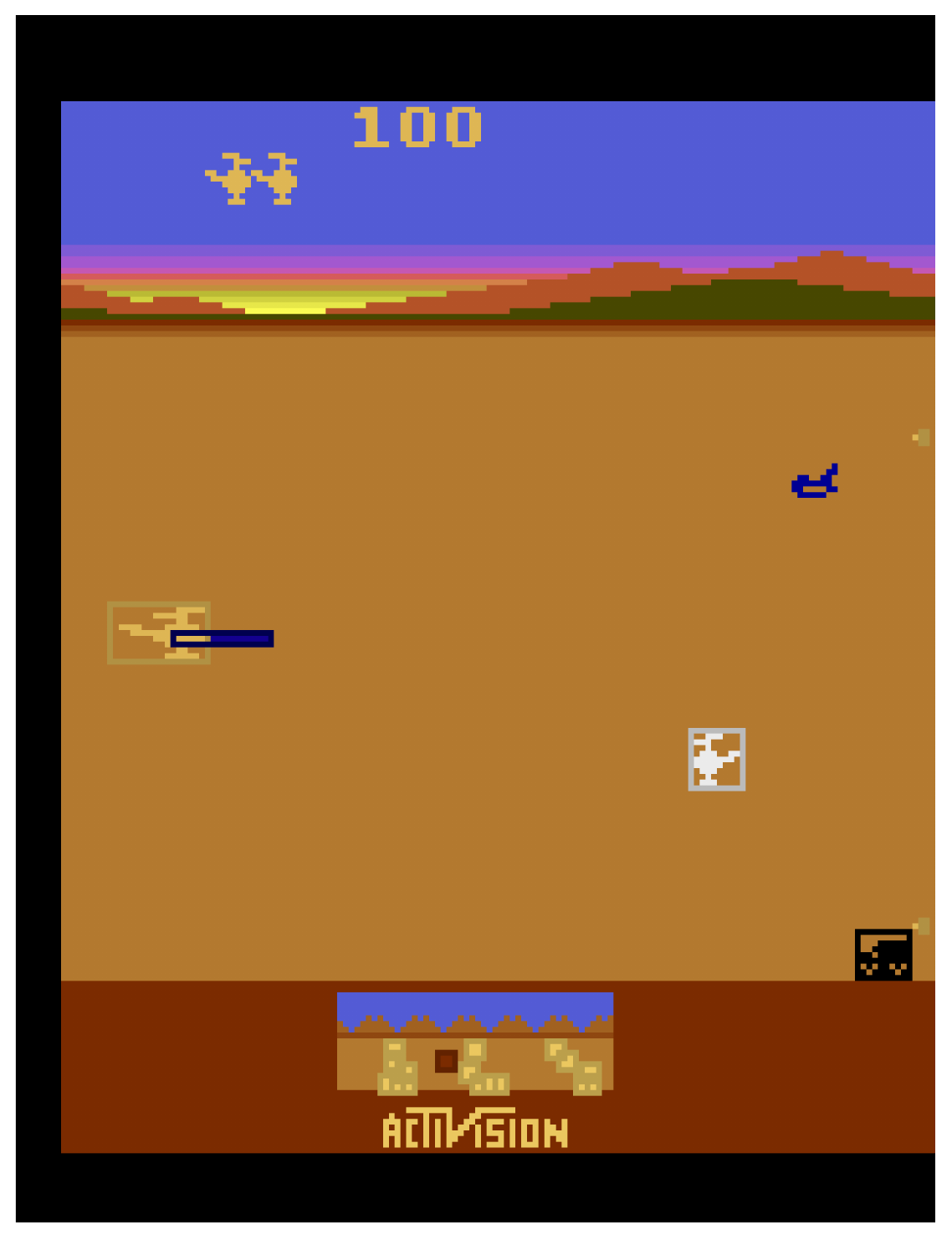}
    \end{minipage}
  }{}
  \IfFileExists{game_modifs/ChopperCommand.tex}{%
      \begin{table}[h]
    \centering
    \begin{tabular}{p{3cm} | p{10cm}} \toprule
        \textbf{Modification} & \textbf{Effect} \\ \midrule
        delay\_shots & Puts time delay between shots \\
        no\_enemies & Removes all Enemies from the game\\
        no\_radar & Removes the radar content\\
        invisible\_player & Makes the player invisible \\
        color\_$X$ & Changes the color of background to color $X \in$ [black, white, red, blue, green]. This also affects the enemies colors \\
        \bottomrule
    \end{tabular}
\end{table}
  }{}
  
  \vspace{-1em} 

  \subsection{DemonAttack}
  \noindent
  \IfFileExists{environment_description/DemonAttack.tex}{%
    \begin{minipage}{0.80\textwidth}
     \textbf{Description:} \\
      You are facing waves of demons in the ice planet of Krybor. Points are accumulated by destroying demons. You begin with 3 reserve bunkers, and can increase its number (up to 6) by avoiding enemy attacks. Each attack wave you survive without any hits, grants you a new bunker. Every time an enemy hits you, a bunker is destroyed. When the last bunker falls, the next enemy hit will destroy you and the game ends.
    \end{minipage}
  }{}
\IfFileExists{environment_images_new_pdfs/DemonAttack.pdf}{%
    \begin{minipage}{0.18\textwidth}
      \raggedleft
      \includegraphics[scale=0.10]{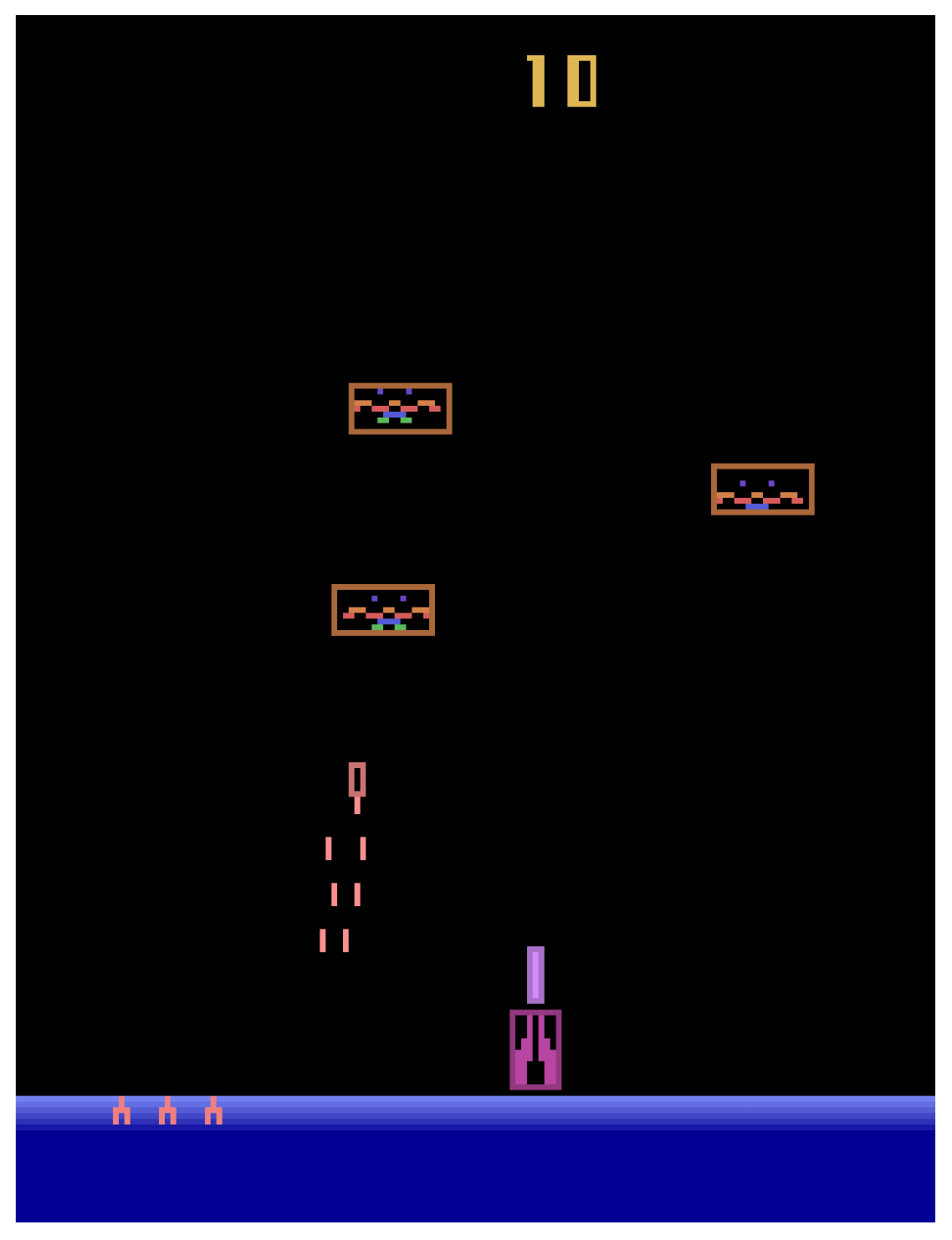}
    \end{minipage}
  }{}
  \IfFileExists{game_modifs/DemonAttack.tex}{%
      \begin{table}[h]
    \centering
    \begin{tabular}{p{3cm} | p{10cm}} \toprule
        \textbf{Modification} & \textbf{Effect} \\ \midrule
        static\_enemies & Makes the enemies horizontally static (static x). \\
        one\_missile & Enemies only shoot one missile, instead of three\\
        \bottomrule
    \end{tabular}
\end{table}
  }{}
  
  \vspace{-1em} 

  \subsection{DonkeyKong}
  \noindent
  \IfFileExists{environment_description/DonkeyKong.tex}{%
    \begin{minipage}{0.80\textwidth}
     \textbf{Description:} \\
      You play as Mario trying to save Pauline from the hands of Donkey Kong. Remove rivets and jump over the barrels, with a score that starts high and counts down throughout the game.
    \end{minipage}
  }{}
\IfFileExists{environment_images_new_pdfs/DonkeyKong.pdf}{%
    \begin{minipage}{0.18\textwidth}
      \raggedleft
      \includegraphics[scale=0.10]{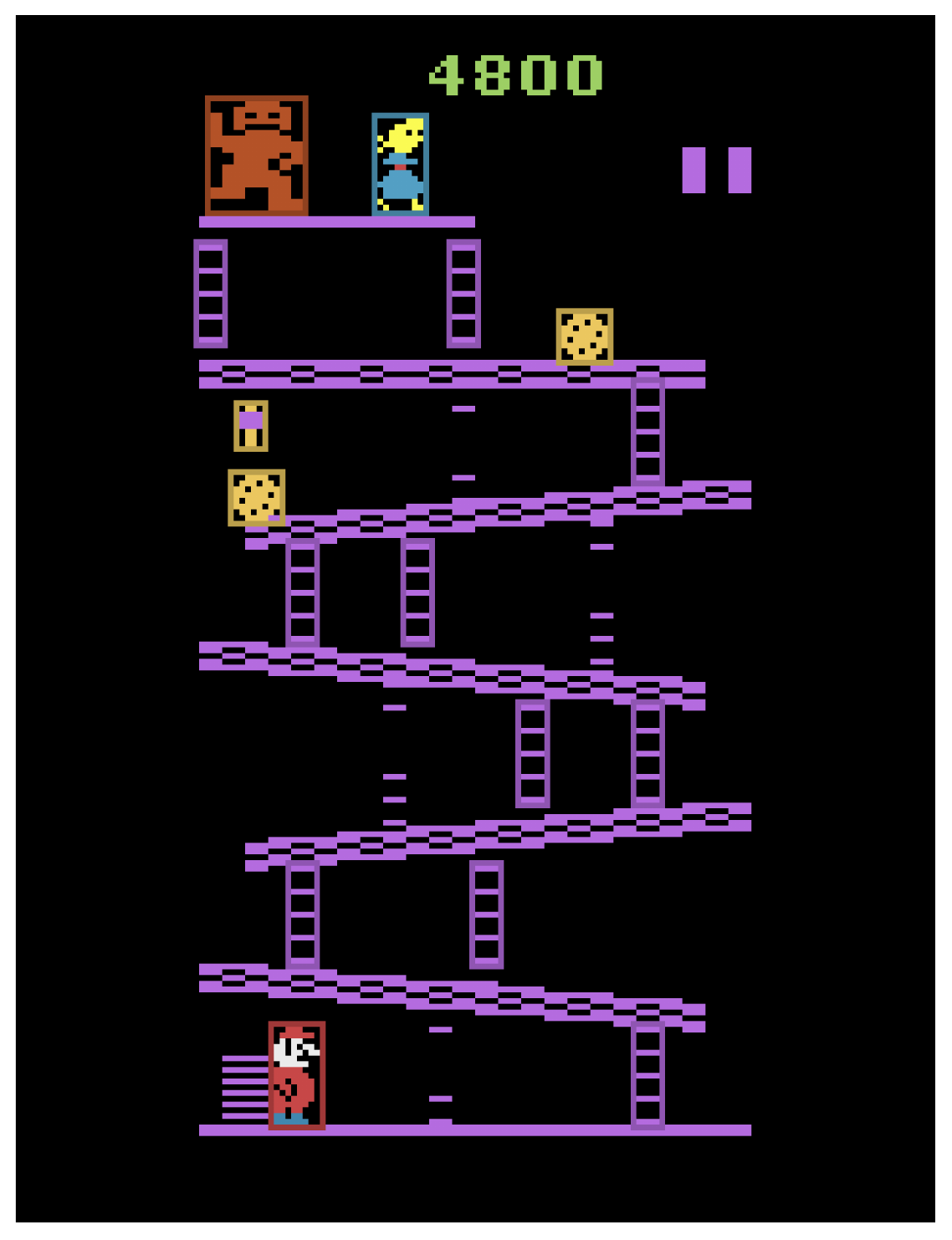}
    \end{minipage}
  }{}
  \IfFileExists{game_modifs/DonkeyKong.tex}{%
      \begin{table}[h]
    \centering
    \begin{tabular}{p{3cm} | p{10cm}} \toprule
        \textbf{Modification} & \textbf{Effect} \\ \midrule
        no\_barrel & Remove barrels from the game. \\
        unlimited\_time & Provides unlimited time for the player. \\
        random\_start & Set the players start position to a random pre-defined start position. \\
        \bottomrule
    \end{tabular}
\end{table}
  }{}
  
  \vspace{-1em} 

  \subsection{FishingDerby}
  \noindent
  \IfFileExists{environment_description/FishingDerby.tex}{%
    \begin{minipage}{0.80\textwidth}
     \textbf{Description:} \\
      Your objective is to catch more sunfish than your opponent. Watch out for the shark, as it tries to snatch the fish caught on your hook.
    \end{minipage}
  }{}
\IfFileExists{environment_images_new_pdfs/FishingDerby.pdf}{%
    \begin{minipage}{0.18\textwidth}
      \raggedleft
      \includegraphics[scale=0.10]{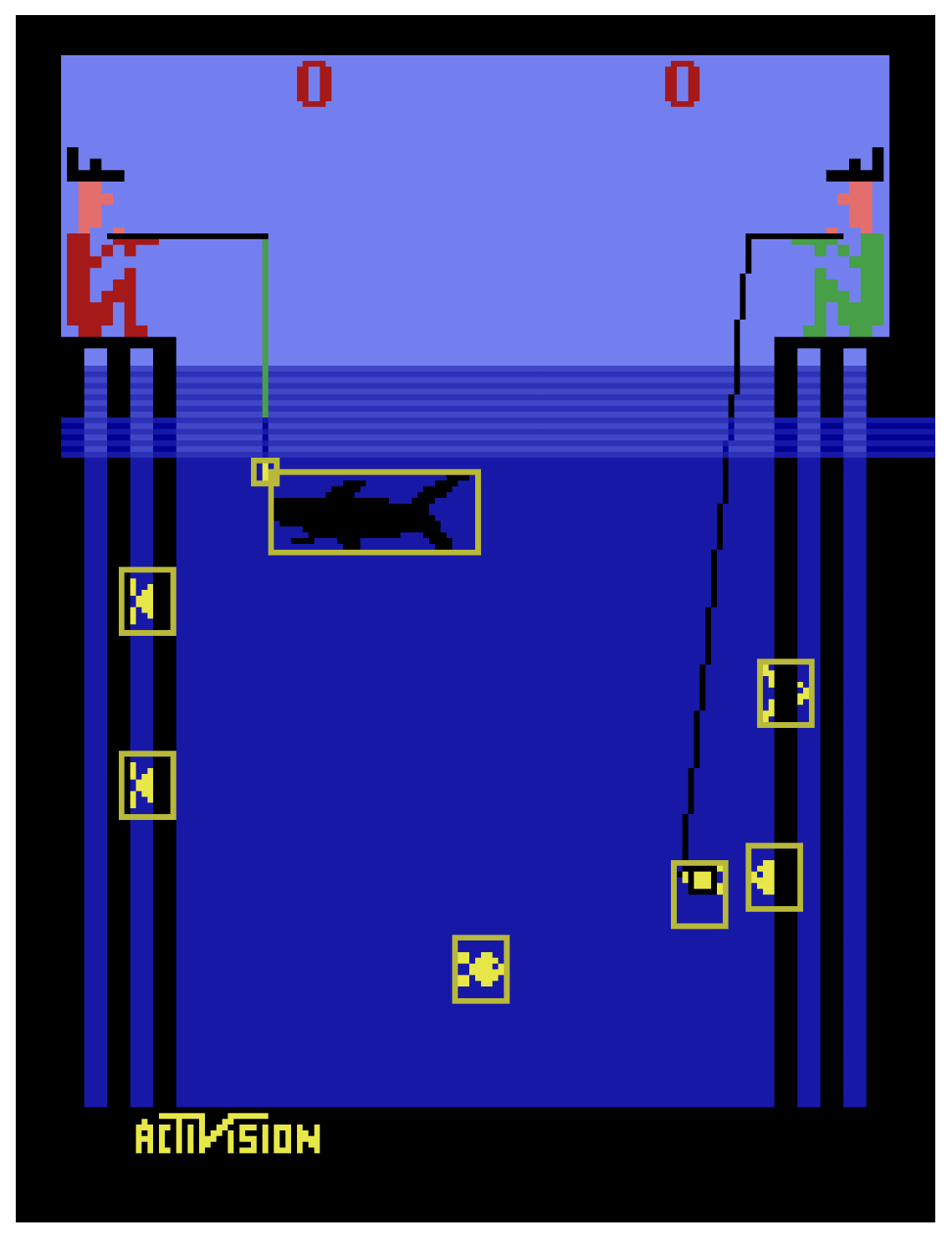}
    \end{minipage}
  }{}
  \IfFileExists{game_modifs/FishingDerby.tex}{%
      \begin{table}[h]
    \centering
    \begin{tabular}{p{4.25cm} | p{8.75cm}} \toprule
        \textbf{Modification} & \textbf{Effect} \\ \midrule
        fish\_on\_player\_side & Spawns all fish on the players side. \\
        fish\_in\_middle & Spawns all fish in the middle. \\
        fished\_on\_each\_sides1 & Swap fish sides, the fish that were on the player side are now on the enemy's one and vice versa. \\
        fished\_on\_each\_sides2 & Swap fish sides, the fish that were on the player side are now on the enemy's one and vice versa. \\
        shark\_no\_movement\_easy & The shark will not move and stay on the opponents side. \\
        shark\_no\_movement\_middle & The shark will not move and is placed in the middle. \\
        shark\_teleport & The shark will teleport between the player and the enemy side at a set interval. \\
        shark\_speed\_mode & Increases the sharks movement speed. \\
        \bottomrule
    \end{tabular}
\end{table}
  }{}
  
  \vspace{-1em} 

  \subsection{Freeway}
  \noindent
  \IfFileExists{environment_description/Freeway.tex}{%
    \begin{minipage}{0.80\textwidth}
     \textbf{Description:} \\
      Your objective is to guide your chicken across lane after lane of busy rush hour traffic. You receive a point for every chicken that makes it to the other side at the top of the screen.
    \end{minipage}
  }{}
\IfFileExists{environment_images_new_pdfs/Freeway.pdf}{%
    \begin{minipage}{0.18\textwidth}
      \raggedleft
      \includegraphics[scale=0.10]{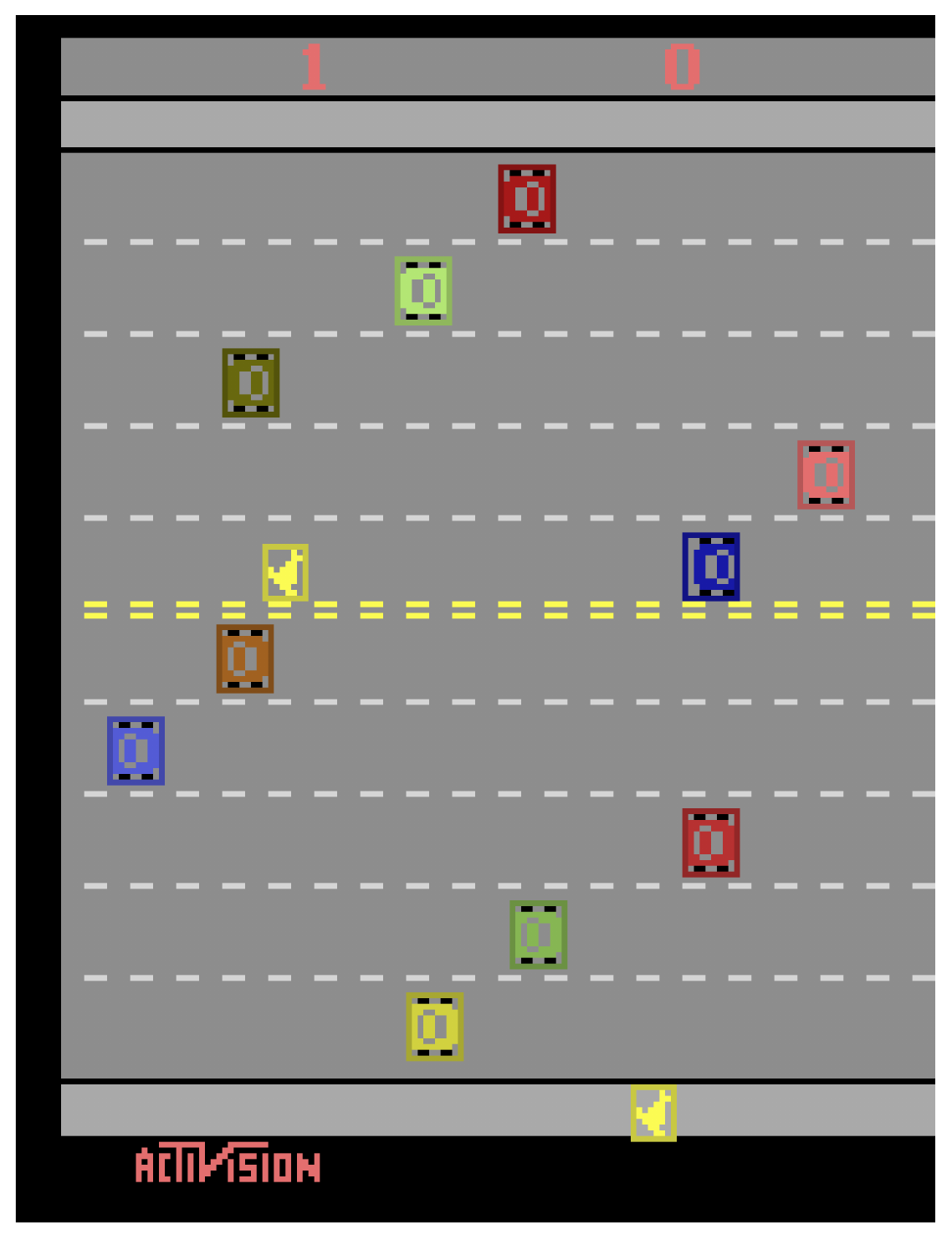}
    \end{minipage}
  }{}
  \IfFileExists{game_modifs/Freeway.tex}{%
      \begin{table}[h]
    \centering
    \begin{tabular}{p{4cm} | p{9cm}} \toprule
        \textbf{Modification} & \textbf{Effect} \\ \midrule
        stop\_random\_car & Stops a random car with a biased probability. \\
        stop\_all\_cars & Stops all cars and repositions some to predefined positions. \\
        align\_all\_cars & Aligns the x position of all cars. \\
        reverse\_car\_speed\_bottom & Reverses the speed order of the cars on the bottom road (the fastest becomes the slowest and vice versa). \\
        reverse\_car\_speed\_top & Reverses the speed order of the cars on the top road (the fastest becomes the slowest and vice versa). \\
        speed\_mode & Increases the speed of all cars. \\
        invisible\_mode & Makes the cars invisible. \\
        phantom\_mode & Each car changes color from black to invisible approximately every second. \\
        blinking\_mode & Each car changes color randomly approximately every second. \\
        strobo\_mode & Each car changes color randomly every timestep. \\
        all\_$X$\_cars & Set color of all cars to $X \in$ [black, white, red, blue, green]. \\
        \bottomrule
    \end{tabular}
\end{table}
  }{}
  
  \vspace{-1em} 

  \subsection{Frostbite}
  \noindent
  \IfFileExists{environment_description/Frostbite.tex}{%
    \begin{minipage}{0.80\textwidth}
     \textbf{Description:} \\
      In Frostbite, the player controls “Frostbite Bailey” who hops back and forth across an Arctic river, changing the color of the ice shards from white to blue. Each time he does so, a block is added to his igloo. Hurry before the polar bear gets you, but watch out for birds that throw you of the shards or crabs that snap you. Look for fish jumping out of the water to receive some bonus points.
    \end{minipage}
  }{}
\IfFileExists{environment_images_new_pdfs/Frostbite.pdf}{%
    \begin{minipage}{0.18\textwidth}
      \raggedleft
      \includegraphics[scale=0.10]{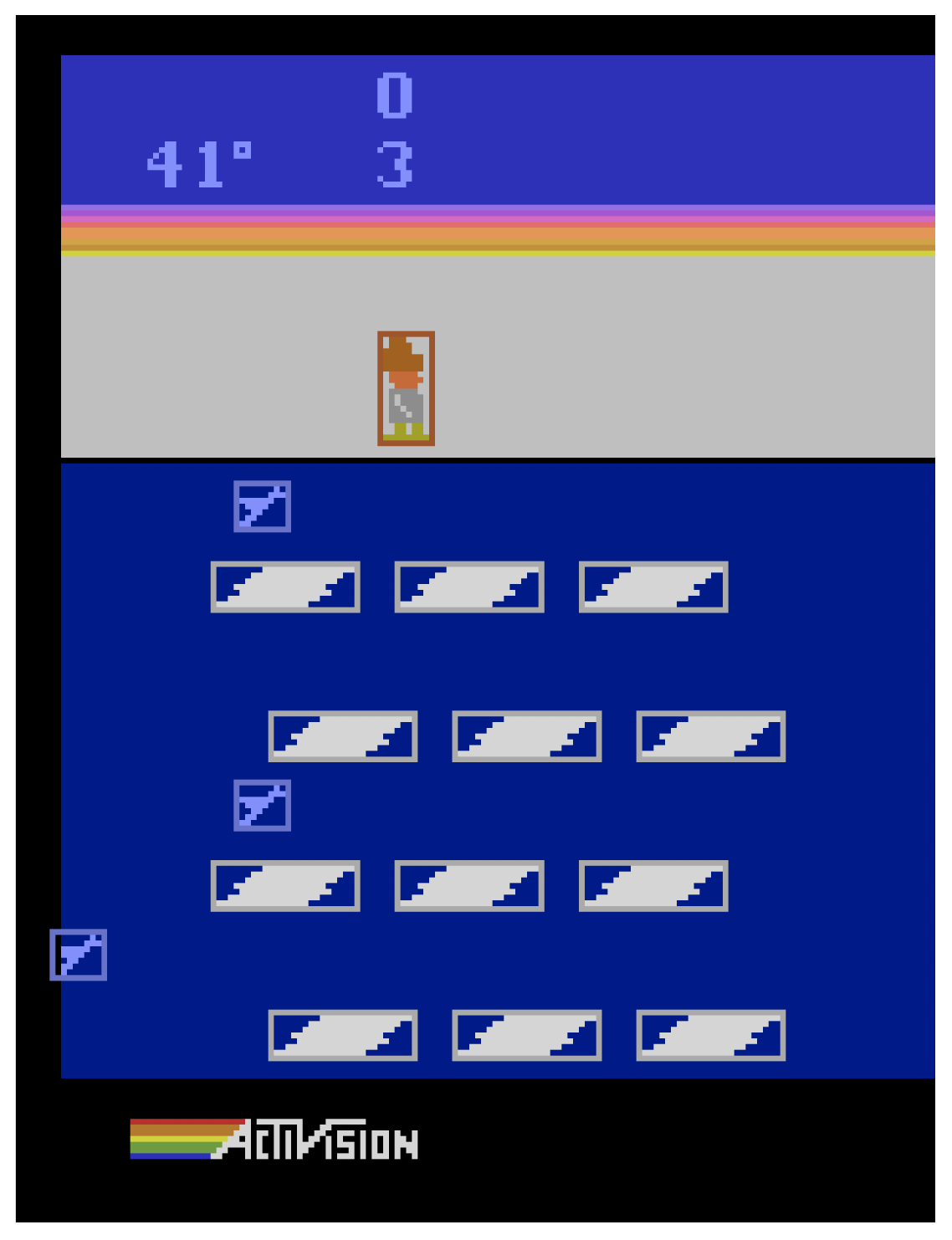}
    \end{minipage}
  }{}
  \IfFileExists{game_modifs/Frostbite.tex}{%
      \begin{table}[h]
    \centering
    \begin{tabular}{p{3.75cm} | p{9.25cm}} \toprule
        \textbf{Modification} & \textbf{Effect} \\ \midrule
        ui\_color\_$X$ &Adjust color of the UI to color $X \in$ [black, red] \\
        reposition\_floes\_easy & Adjusts the position of the ice floes to an easy layout. \\
        reposition\_floes\_medium & Adjusts the position of the ice floes to a medium layout. \\
        reposition\_floes\_hard & Adjusts the position of the ice floes to a hard layout. \\
        no\_birds & Removes the birds. \\
        few\_enemies & Reduces the amount of enemies spawning in. \\
        many\_enemies & Increases the amount of enemies spawning in. \\
        \bottomrule
    \end{tabular}
\end{table}
  }{}
  
  \vspace{-1em} 

  \subsection{Jamesbond}
  \noindent
  \IfFileExists{environment_description/Jamesbond.tex}{%
    \begin{minipage}{0.80\textwidth}
     \textbf{Description:} \\
      Your mission is to control Mr. Bond’s specially designed multipurpose craft to complete a variety of missions. The craft moves forward with a right input and slightly back with a left input. An up or down input causes the craft to jump or dive. You can also fire by either lobbing a bomb to the bottom of the screen or firing a fixed angle shot to the top of the screen.
    \end{minipage}
  }{}
\IfFileExists{environment_images_new_pdfs/Jamesbond.pdf}{%
    \begin{minipage}{0.18\textwidth}
      \raggedleft
      \includegraphics[scale=0.10]{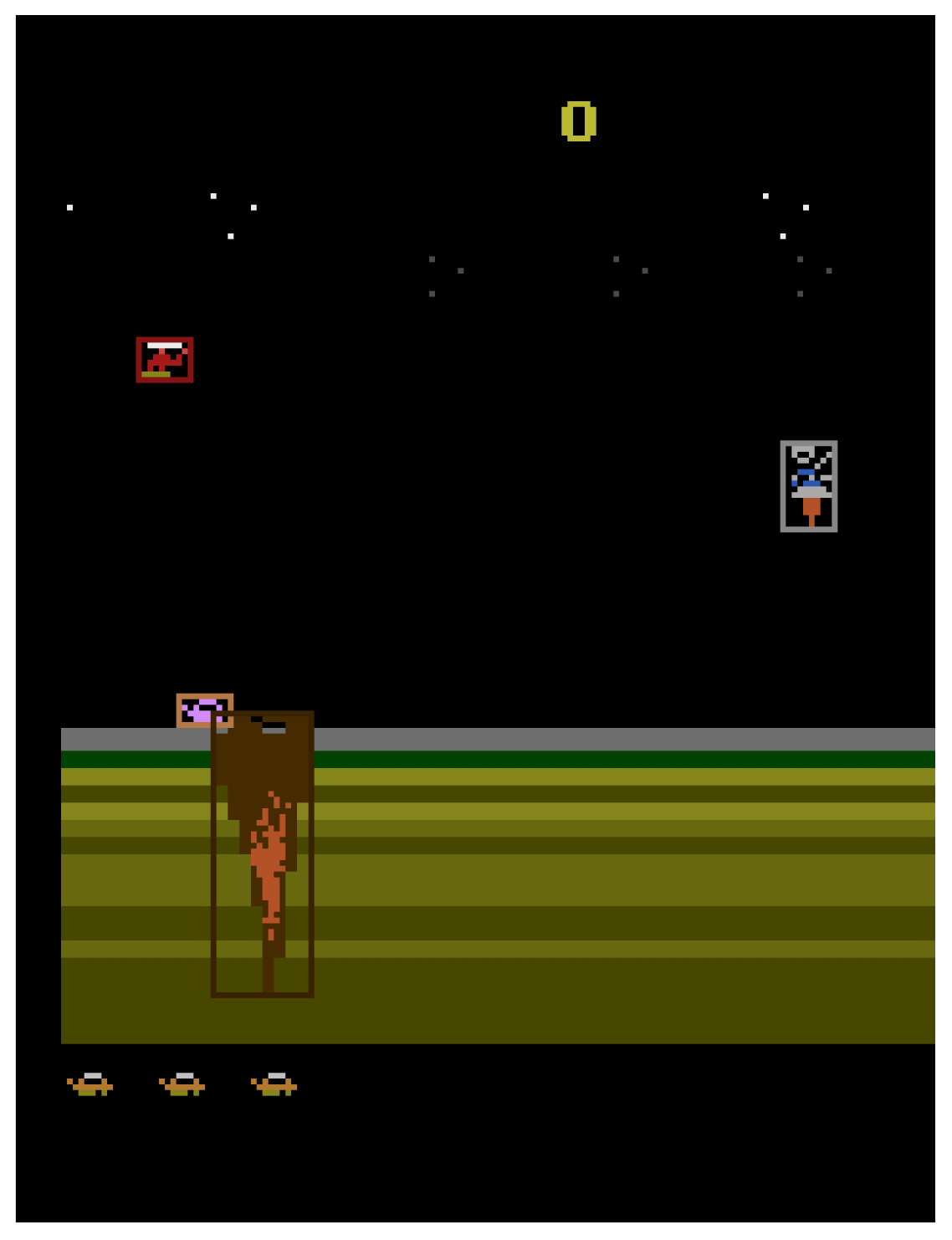}
    \end{minipage}
  }{}
  \IfFileExists{game_modifs/Jamesbond.tex}{%
      \begin{table}[h]
    \centering
    \begin{tabular}{p{3cm} | p{10cm}} \toprule
        \textbf{Modification} & \textbf{Effect} \\ \midrule
        constant\_jump & Makes the player character jump constantly. \\ 
        fast\_backward & Increases the reversing speed. \\
        mobile\_player & Makes the player character jump constantly and increases the reversing speed. \\
        straight\_shots & The player shots go straight up, instead of diagonal. \\
        unlimited\_lives & Player has an unlimited amounts of lives. \\
        \bottomrule
    \end{tabular}
\end{table}
  }{}
  
  \vspace{-1em} 

  \subsection{Kangaroo}
  \noindent
  \IfFileExists{environment_description/Kangaroo.tex}{%
    \begin{minipage}{0.80\textwidth}
     \textbf{Description:} \\
      The object of the game is to score as many points as you can while controlling Mother Kangaroo to rescue her precious baby. You start the game with three lives. During this rescue mission, Mother Kangaroo encounters many obstacles. You need to help her climb ladders, pick bonus fruit, and throw punches at monkeys, while avoiding the falling coconuts.
    \end{minipage}
  }{}
\IfFileExists{environment_images_new_pdfs/Kangaroo.pdf}{%
    \begin{minipage}{0.18\textwidth}
      \raggedleft
      \includegraphics[scale=0.10]{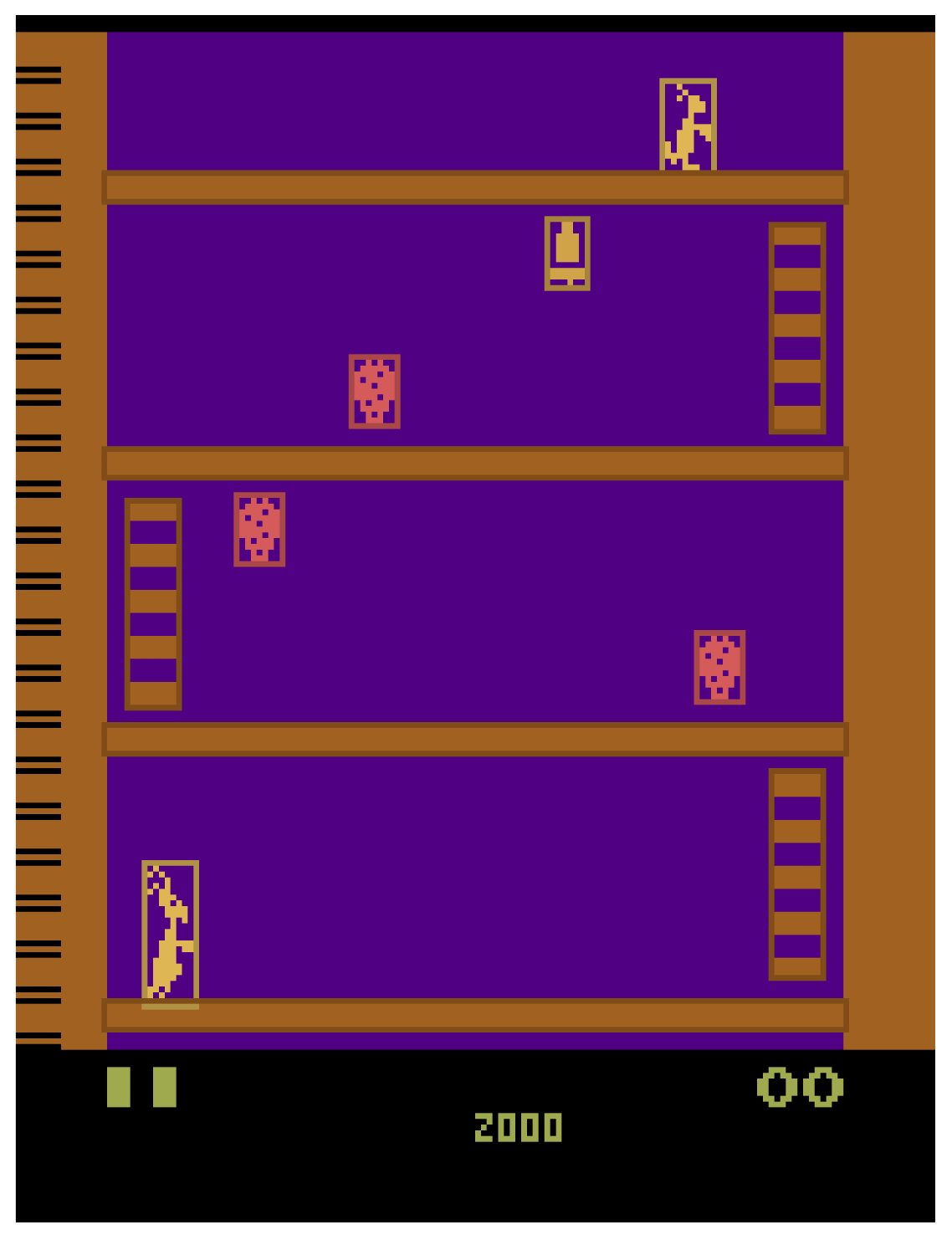}
    \end{minipage}
  }{}
  \IfFileExists{game_modifs/Kangaroo.tex}{%
      \begin{table}[h]
    \centering
    \begin{tabular}{p{4.5cm} | p{8.5cm}} \toprule
        \textbf{Modification} &  \textbf{Effect} \\ \midrule
        disable\_monkeys & Disables the monkeys in the game. \\
        disable\_coconut & Disables the coconuts in the game. \\
        disable\_thrown\_coconut & Disables the thrown coconut in the game. \\
        no\_danger & Combines the three modifications above, disabling all hazards in the game. \\
        set\_kangaroo\_position\_floor1 & Sets the kangaroo's position for floor 1.\\
        set\_kangaroo\_position\_floor2 & Sets the kangaroo's position for floor 2.\\
        randomize\_kangaroo\_position & Randomize the floor on which the kangaroo starts.\\
        change\_level1 & Changes the level to 1. \\
        change\_level2 & Changes the level to 2. \\
        change\_level3 & Changes the level to 3. \\
        unlimited\_time & Provides unlimited time to clear the level. \\
        \bottomrule
    \end{tabular}
\end{table}
  }{}
  
  \vspace{-1em} 

  \subsection{KungFuMaster}
  \noindent
  \IfFileExists{environment_description/KungFuMaster.tex}{%
    \begin{minipage}{0.80\textwidth}
     \textbf{Description:} \\
      You are a Kung-Fu master on the mission to rescue your girlfriend from the evil Mr. X. find your way through the castle fighting his lackeys and dodging traps.
    \end{minipage}
  }{}
\IfFileExists{environment_images_new_pdfs/KungFuMaster.pdf}{%
    \begin{minipage}{0.18\textwidth}
      \raggedleft
      \includegraphics[scale=0.10]{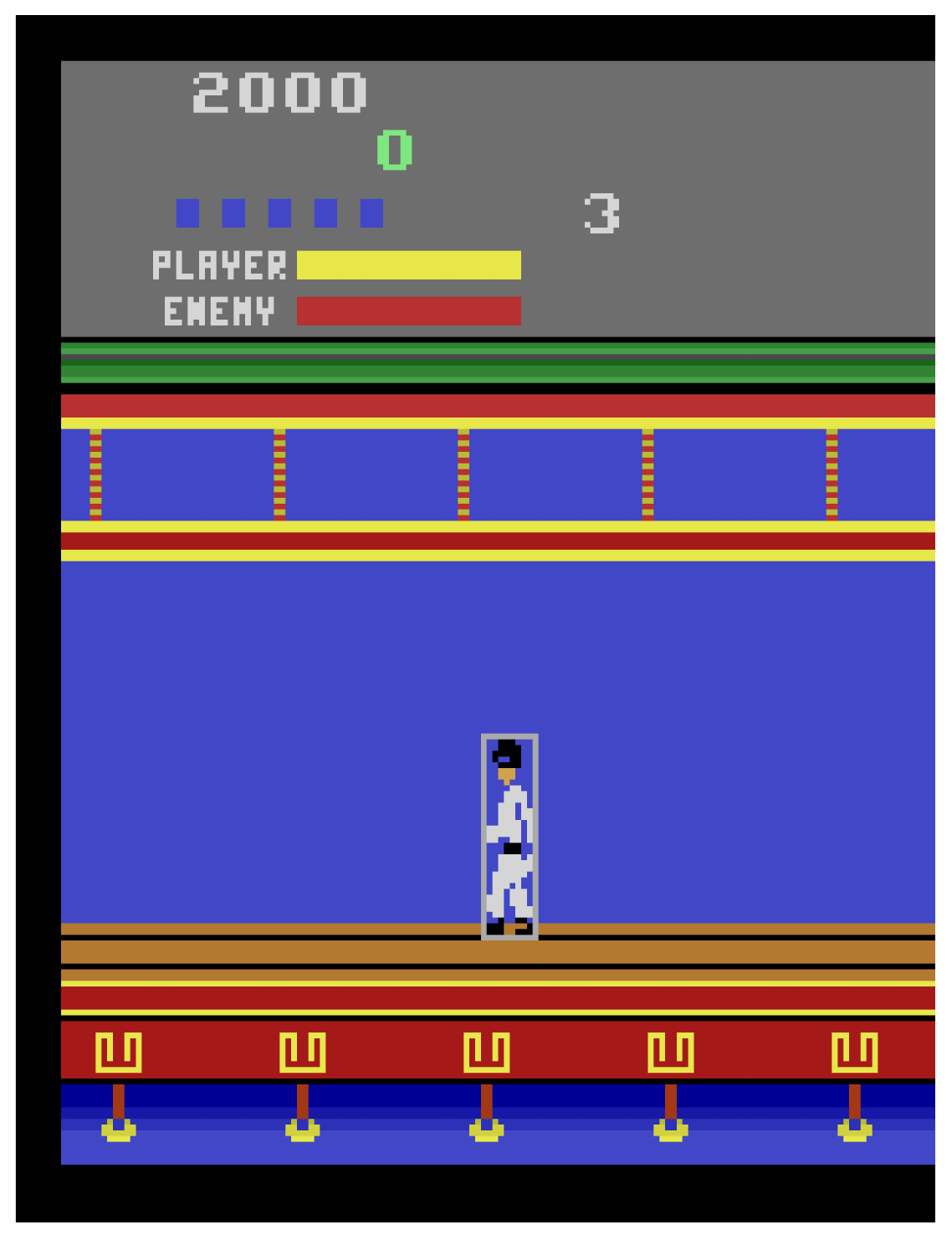}
    \end{minipage}
  }{}
  \IfFileExists{game_modifs/KungFuMaster.tex}{%
      \begin{table}[h]
    \centering
    \begin{tabular}{p{3cm} | p{10cm}} \toprule
        \textbf{Modification} & \textbf{Effect} \\ \midrule
        no\_damage & Player does not take damage. \\
        unlimited\_time & Provides unlimited time to clear the level. \\
        unlimited\_lives & Player has an unlimited amounts of lives. \\ \bottomrule
    \end{tabular}
\end{table}
  }{}
  
  \vspace{-1em} 

  \subsection{MontezumaRevenge}
  \noindent
  \IfFileExists{environment_description/MontezumaRevenge.tex}{%
    \begin{minipage}{0.80\textwidth}
     \textbf{Description:} \\
      Your goal is to acquire Montezuma’s treasure by making your way through a maze of chambers within the emperor’s fortress. You must avoid deadly creatures while collecting valuables and tools which can help you escape with the treasure.
    \end{minipage}
  }{}
\IfFileExists{environment_images_new_pdfs/MontezumaRevenge.pdf}{%
    \begin{minipage}{0.18\textwidth}
      \raggedleft
      \includegraphics[scale=0.10]{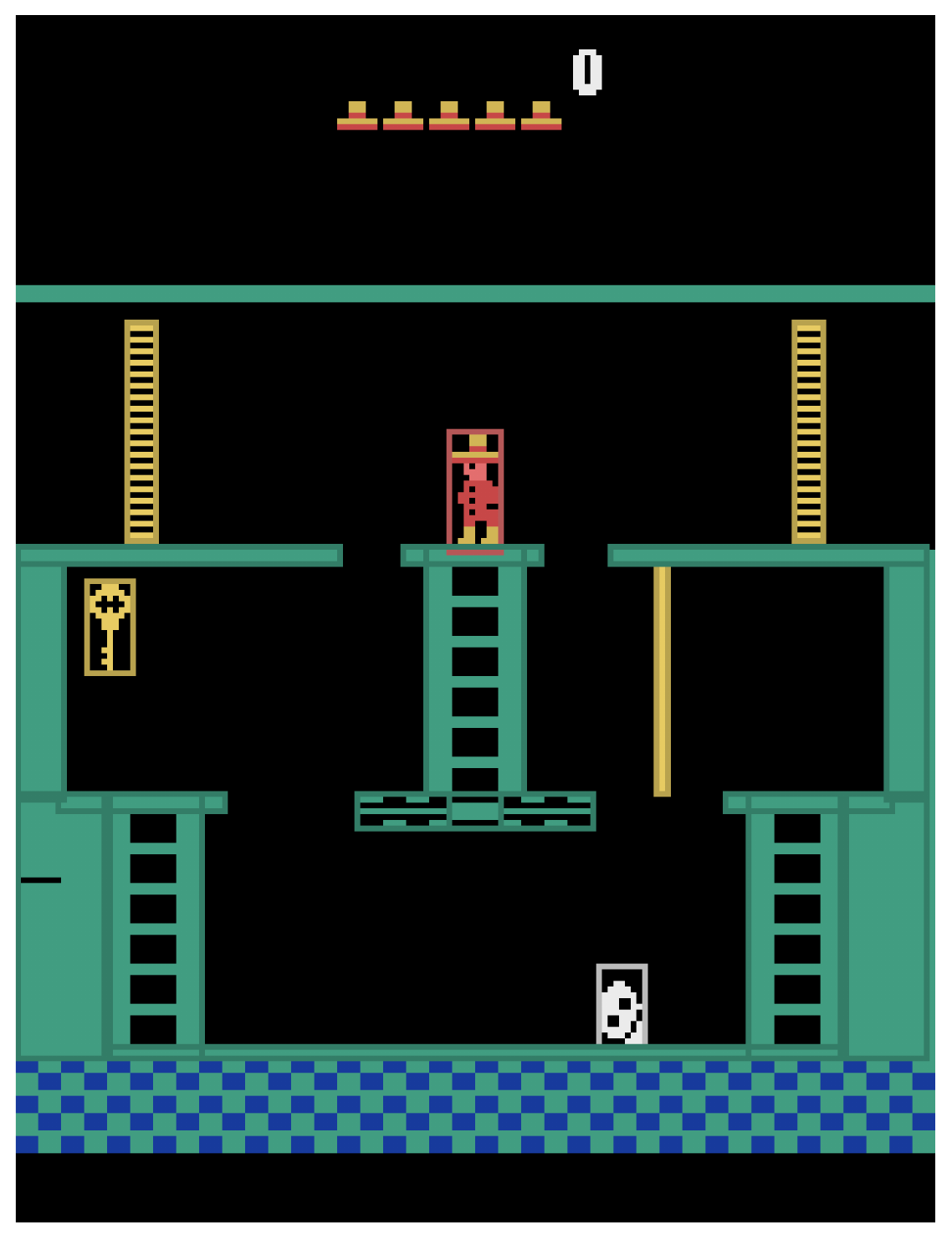}
    \end{minipage}
  }{}
  \IfFileExists{game_modifs/MontezumaRevenge.tex}{%
      \begin{table}[h]
    \centering
    \begin{tabular}{p{3cm} | p{10cm}} \toprule
        \textbf{Modification} & textbf{Effect} \\ \midrule
        random\_position\_start & Randomize the start position within the room \\
        set\_level\_0 & Sets the level to 0.\\
        set\_level\_1 & Sets the level to 1.\\
        set\_level\_2 & Sets the level to 2.\\
        randomize\_items & Randomize which item is found in which room. \\
        full\_inventory & Adds all items to inventory. \\      
        \bottomrule
    \end{tabular}
\end{table}
  }{}
  
  \vspace{-1em} 

  \subsection{MsPacman}
  \noindent
  \IfFileExists{environment_description/MsPacman.tex}{%
    \begin{minipage}{0.80\textwidth}
     \textbf{Description:} \\
      Your goal is to collect all of the pellets on the screen while avoiding the ghosts.
    \end{minipage}
  }{}
\IfFileExists{environment_images_new_pdfs/MsPacman.pdf}{%
    \begin{minipage}{0.18\textwidth}
      \raggedleft
      \includegraphics[scale=0.10]{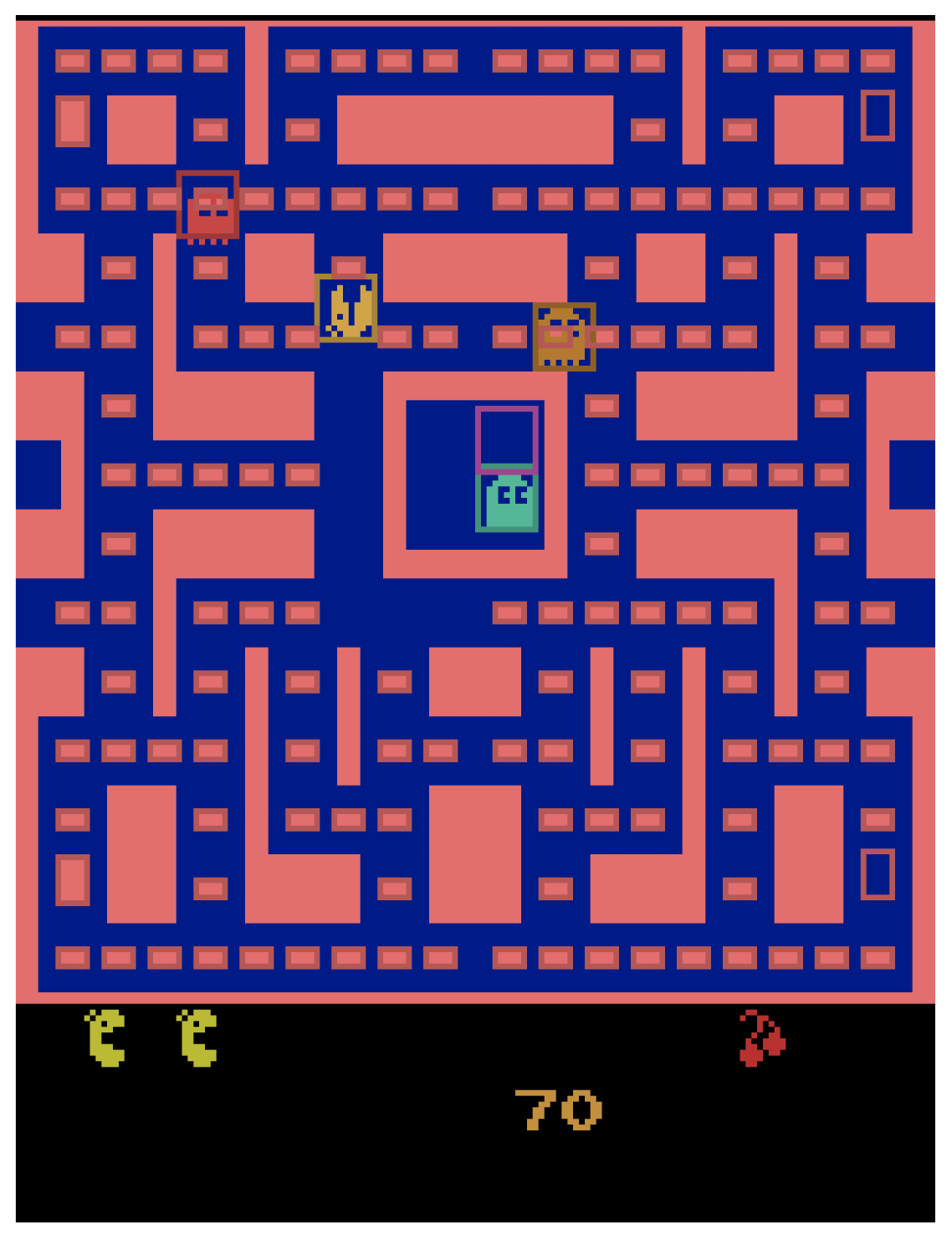}
    \end{minipage}
  }{}
  \IfFileExists{game_modifs/MsPacman.tex}{%
      \begin{table}[h]
    \centering
    \begin{tabular}{p{3cm} | p{10cm}} \toprule
        \textbf{Modification} &  \textbf{Effect} \\ \midrule
        caged\_ghosts & Fix the position of the ghost inside the square in the middle of the screen. \\
        disable\_orange & Fix the position of the orange ghost only. \\
        disable\_red & Fix the position of the red ghost only. \\
        disable\_cyan & Fix the position of the cyan ghost only. \\
        disable\_pink & Fix the position of the pink ghost only. \\
        caged\_ghosts & Fix the position of all the ghosts. \\
        edible\_ghosts & All ghost will be made edible the entire game \\
        set\_level\_$X$ & Changes the level to $X \in$ [0, 1, 2]. \\
        end\_game & Simulates an almost done game, with 15 pills remaining in the level. \\
        maze\_man & Changes the game to a maze solving task. Only one pill will spawn at a time. After the player collects it, a new pill will spawn. The game is won when the player has collected 20 pills. \\
        \bottomrule
    \end{tabular}
\end{table}
  }{}
  
  \vspace{-1em} 

  \subsection{NameThisGame}
  \noindent
  \IfFileExists{environment_description/NameThisGame.tex}{%
    \begin{minipage}{0.80\textwidth}
     \textbf{Description:} \\
      You, the diver need to defend your treasure from the giant octopus at the top of the screen. Defend yourself by shooting his tentacles and the shark that is looking to bite you. Remember to refill your oxygen via the the boats tube, before you run out.
    \end{minipage}
  }{}
\IfFileExists{environment_images_new_pdfs/NameThisGame.pdf}{%
    \begin{minipage}{0.18\textwidth}
      \raggedleft
      \includegraphics[scale=0.10]{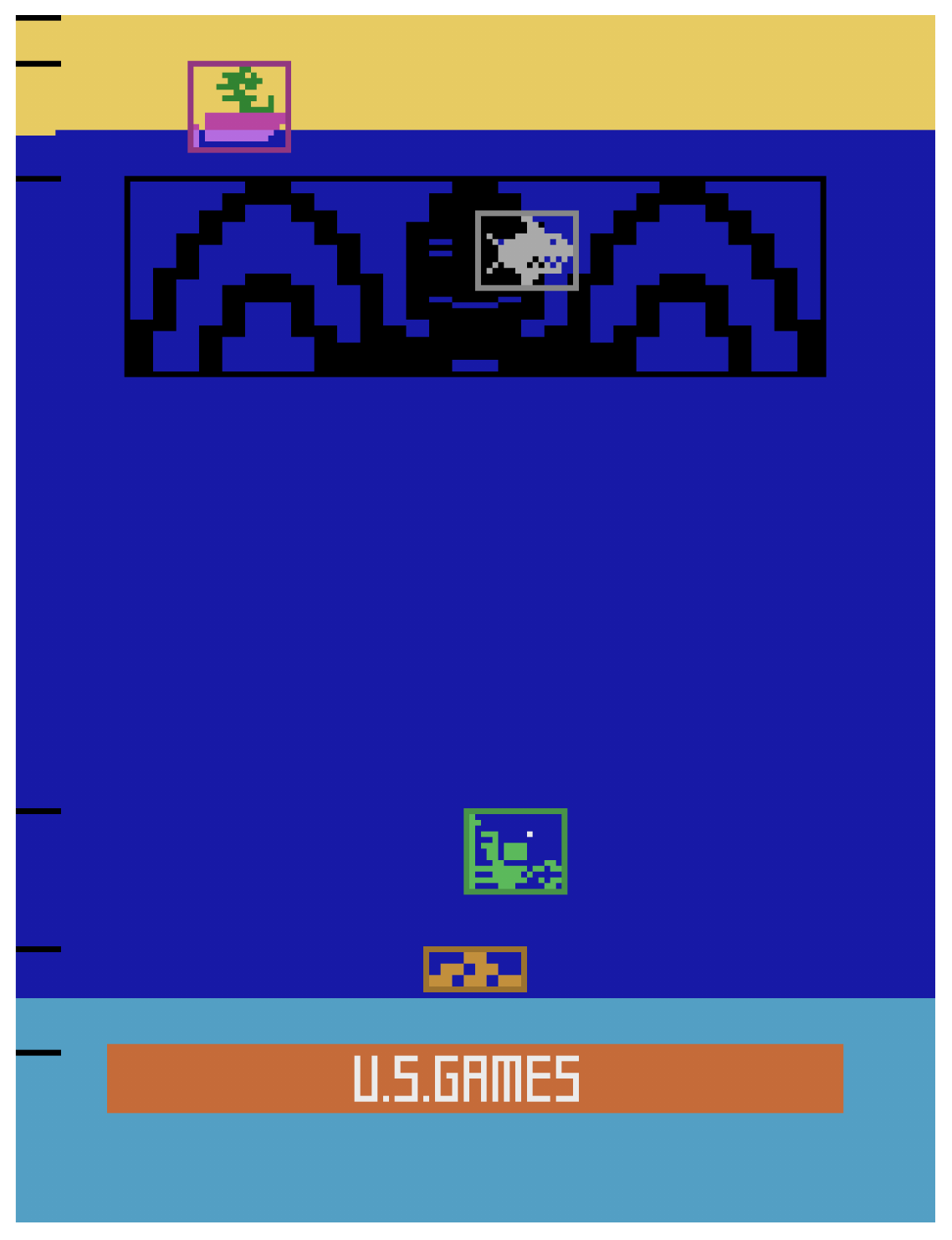}
    \end{minipage}
  }{}
  \IfFileExists{game_modifs/NameThisGame.tex}{%
      \foreach \n in {NameThisGame}{\input{game_modifs/\n.tex}}
  }{}
  
  \vspace{-1em} 

  \subsection{Pong}
  \noindent
  \IfFileExists{environment_description/Pong.tex}{%
    \begin{minipage}{0.80\textwidth}
     \textbf{Description:} \\
      You control the right paddle and compete against the left paddle controlled by the computer. You each try to keep deflecting the ball away from your goal and into your opponent’s goal.
    \end{minipage}
  }{}
\IfFileExists{environment_images_new_pdfs/Pong.pdf}{%
    \begin{minipage}{0.18\textwidth}
      \raggedleft
      \includegraphics[scale=0.10]{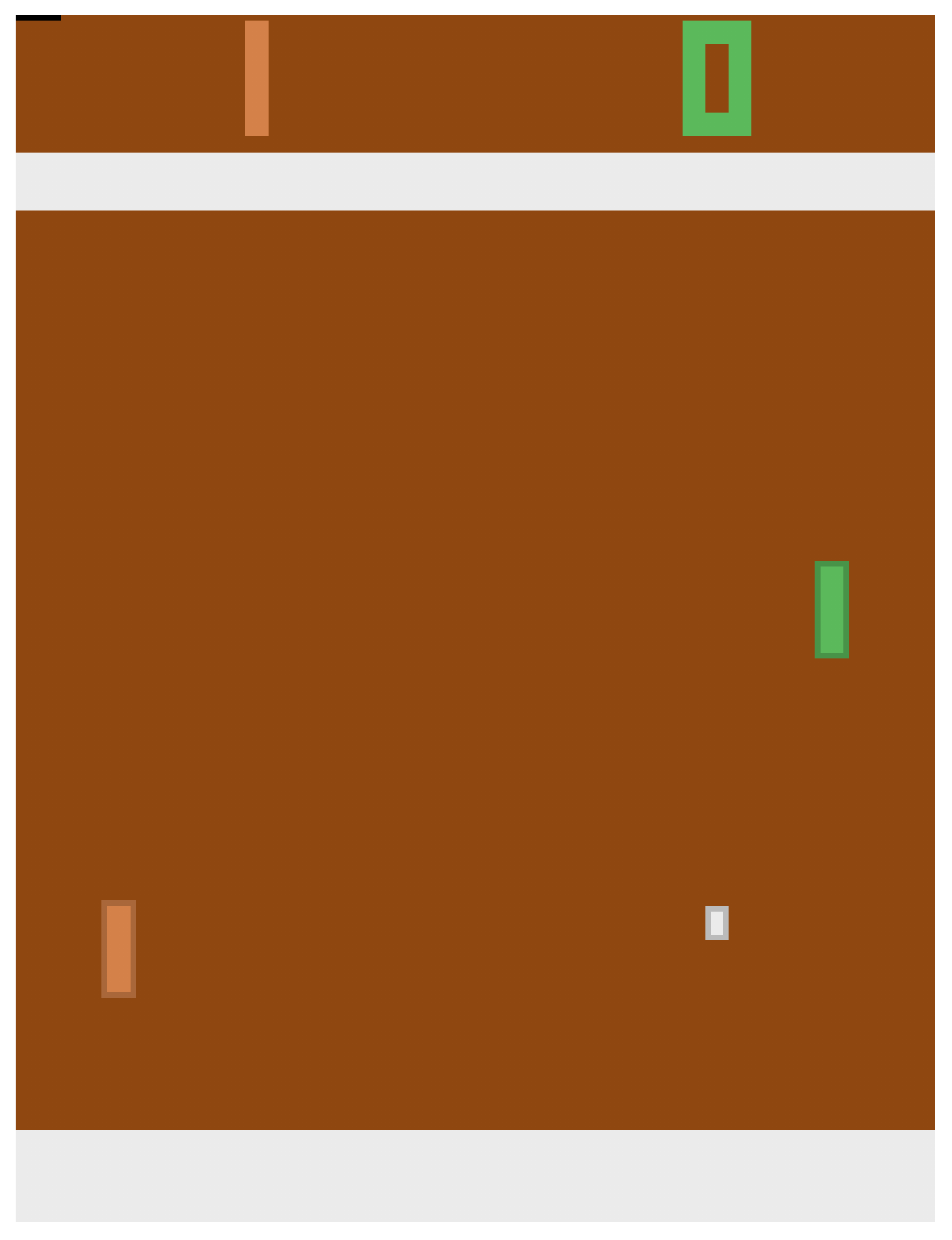}
    \end{minipage}
  }{}
  \IfFileExists{game_modifs/Pong.tex}{%
      \begin{table}[h]
    \centering
    \begin{tabular}{p{3cm} | p{10cm}} \toprule
        \textbf{Modification} & \textbf{Effect} \\ \midrule
        lazy\_enemy & Enemy does not move after returning the shot until player hits ball.\\
        hidden\_enemy & Removes the enemy from the OCAtari object list (Makes it invisible to object based agents).\\
        up\_drift &  Makes the ball drift upwards. \\
        down\_drift&  Makes the ball drift down.\\
        left\_drift &  Makes the ball drift to the left.\\
        right\_drift &  Makes the ball drift to the right.\\
        \bottomrule
    \end{tabular}
\end{table}
  }{}
  
  \vspace{-1em} 

  \subsection{RiverRaid}
  \noindent
  \IfFileExists{environment_description/RiverRaid.tex}{%
    \begin{minipage}{0.80\textwidth}
     \textbf{Description:} \\
      \begin{table}[h]
    \centering
    \begin{tabular}{p{3cm} | p{10cm}} \toprule
        \textbf{Modification} & \textbf{Effect} \\ \midrule
        no\_fuel & Removes the fuel deposits from the game. \\
        red\_river & Turns the river red. \\
        linear\_river & Makes the river straight, however objects still spwan at their normal position making them unreachable in the worst case. \\
        exploding\_fuels & Shooting the fuel deposits will now provides -80 points (instead of 20). \\
        unlimited\_lives & Player has an unlimited amounts of lives. \\
        restricted\_firing & The player is only able to shoot in critical situation, facing a bridge or in a corridor. \\
        unlimited\_lives & Player has an unlimited amounts of lives. \\
        restricted\_firing & The player is only able to shoot in critical situation, facing a bridge or in a corridor. \\
        game\_color\_change\_$X$ &  Changes the color of the player to color $X \in$ [01, 02, 03]\\
        object\_color\_change\_$X$ &  Changes the color of the player to color $X \in$ [01, 02, 03]\\
        \bottomrule
    \end{tabular}
\end{table}
    \end{minipage}
  }{}
\IfFileExists{environment_images_new_pdfs/RiverRaid.pdf}{%
    \begin{minipage}{0.18\textwidth}
      \raggedleft
      \includegraphics[scale=0.10]{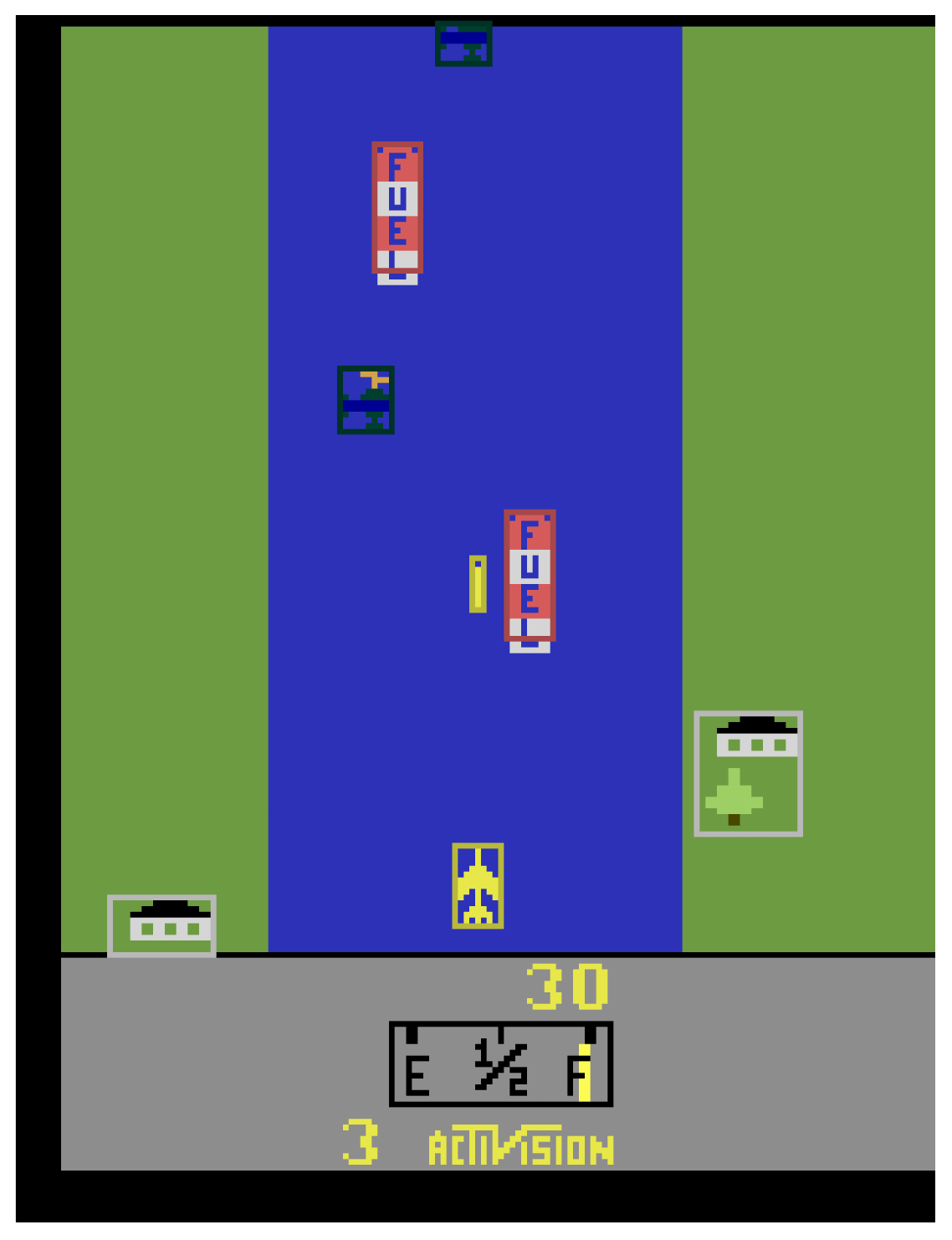}
    \end{minipage}
  }{}
  \IfFileExists{game_modifs/RiverRaid.tex}{%
      
  }{}
  
  \vspace{-1em} 

  \subsection{RoboTank}
  \noindent
  \IfFileExists{environment_description/RoboTank.tex}{%
    \begin{minipage}{0.80\textwidth}
     \textbf{Description:} \\
      You are playing a tank looking to defeat the enemy squadrons of tanks. Each squadron consists of 12 tanks. Use the radar to find enemy tanks even in the night or during foggy weather. Being hit by the enemy requires you to switch to one of your reserve tanks. Enemy stray shots may destroy vital sensors.
    \end{minipage}
  }{}
\IfFileExists{environment_images_new_pdfs/RoboTank.pdf}{%
    \begin{minipage}{0.18\textwidth}
      \raggedleft
      \includegraphics[scale=0.10]{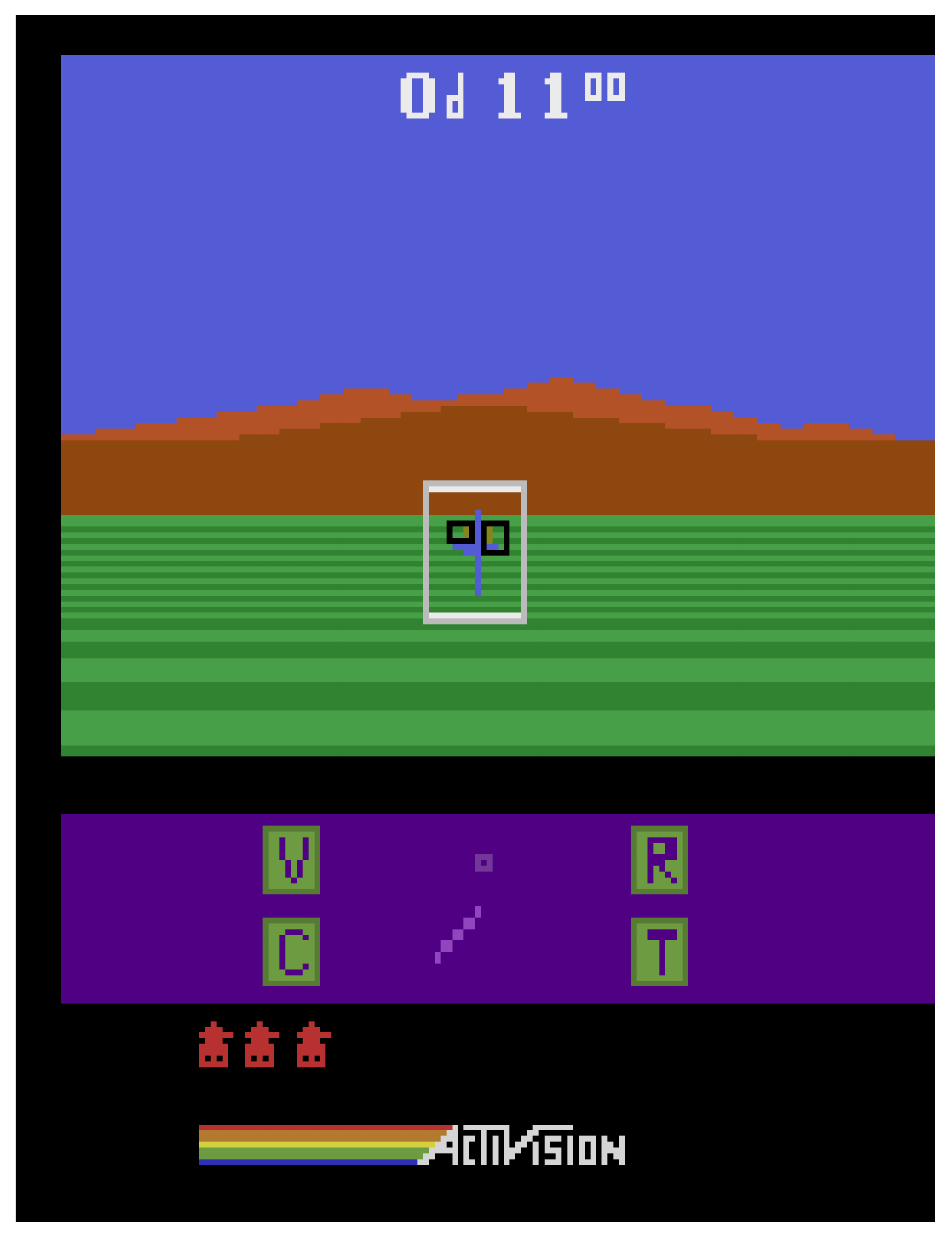}
    \end{minipage}
  }{}
  \IfFileExists{game_modifs/RoboTank.tex}{%
      \begin{table}[h]
    \centering
    \begin{tabular}{p{3cm} | p{10cm}} \toprule
        \textbf{Modification} & \textbf{Effect} \\ \midrule
        fog & Weather condition is always set to fog. \\
        snow & Weather condition is always set to snow. \\
        rain & Weather condition is always set to rain. \\
        tread\_damage & Tread sensor is damaged. \\ 
        canon\_damage & Canon is damaged. \\
        vision\_damage & Vision sensor is damaged. \\ 
        no\_radar & Disables the radar. \\ 
        \bottomrule
    \end{tabular}
\end{table}
  }{}
  
  \vspace{-1em} 

  \subsection{Seaquest}
  \noindent
  \IfFileExists{environment_description/Seaquest.tex}{%
    \begin{minipage}{0.80\textwidth}
     \textbf{Description:} \\
      You control a submarine able to move in all directions and fire torpedoes. The goal is to rescue as many divers as you can, while dodging and blasting enemy subs and killer sharks; points will be awarded accordingly. The game begins with one sub and three waiting on the horizon. Each time you score 10,000 points, an extra sub will be delivered to your base. You can only have six reserve subs at one time. Your sub will explode if it collides with anything except your own divers. The sub has a limited amount of oxygen that decreases at a constant rate during the game. When the oxygen tank is almost empty, you need to surface and if you don’t do it in time, your sub will blow up and you’ll lose one diver. Each time you’re forced to surface, with less than six divers, you lose one diver as well.
    \end{minipage}
  }{}
\IfFileExists{environment_images_new_pdfs/Seaquest.pdf}{%
    \begin{minipage}{0.18\textwidth}
      \raggedleft
      \includegraphics[scale=0.10]{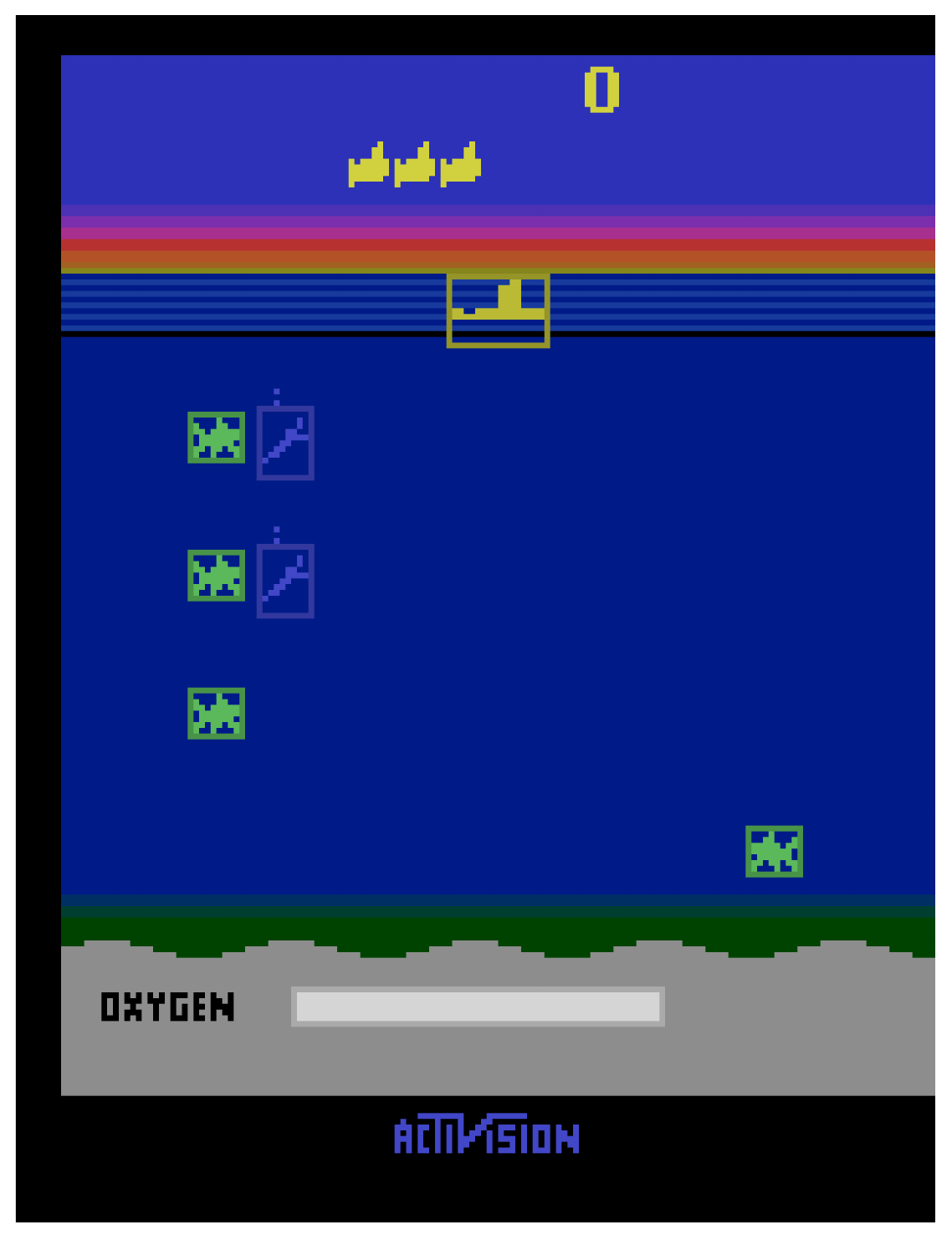}
    \end{minipage}
  }{}
  \IfFileExists{game_modifs/Seaquest.tex}{%
      \foreach \n in {Seaquest}{\input{game_modifs/\n.tex}}
  }{}
  
  \vspace{-1em} 

  \subsection{Skiing}
  \noindent
  \IfFileExists{environment_description/Skiing.tex}{%
    \begin{minipage}{0.80\textwidth}
     \textbf{Description:} \\
      You control a skier who can move sideways. The goal is to pass through all the gates (between the poles) in the fastest time. You are penalized five seconds for each gate you miss. If you hit a gate or a tree, your skier will jump back up and keep going.
    \end{minipage}
  }{}
\IfFileExists{environment_images_new_pdfs/Skiing.pdf}{%
    \begin{minipage}{0.18\textwidth}
      \raggedleft
      \includegraphics[scale=0.10]{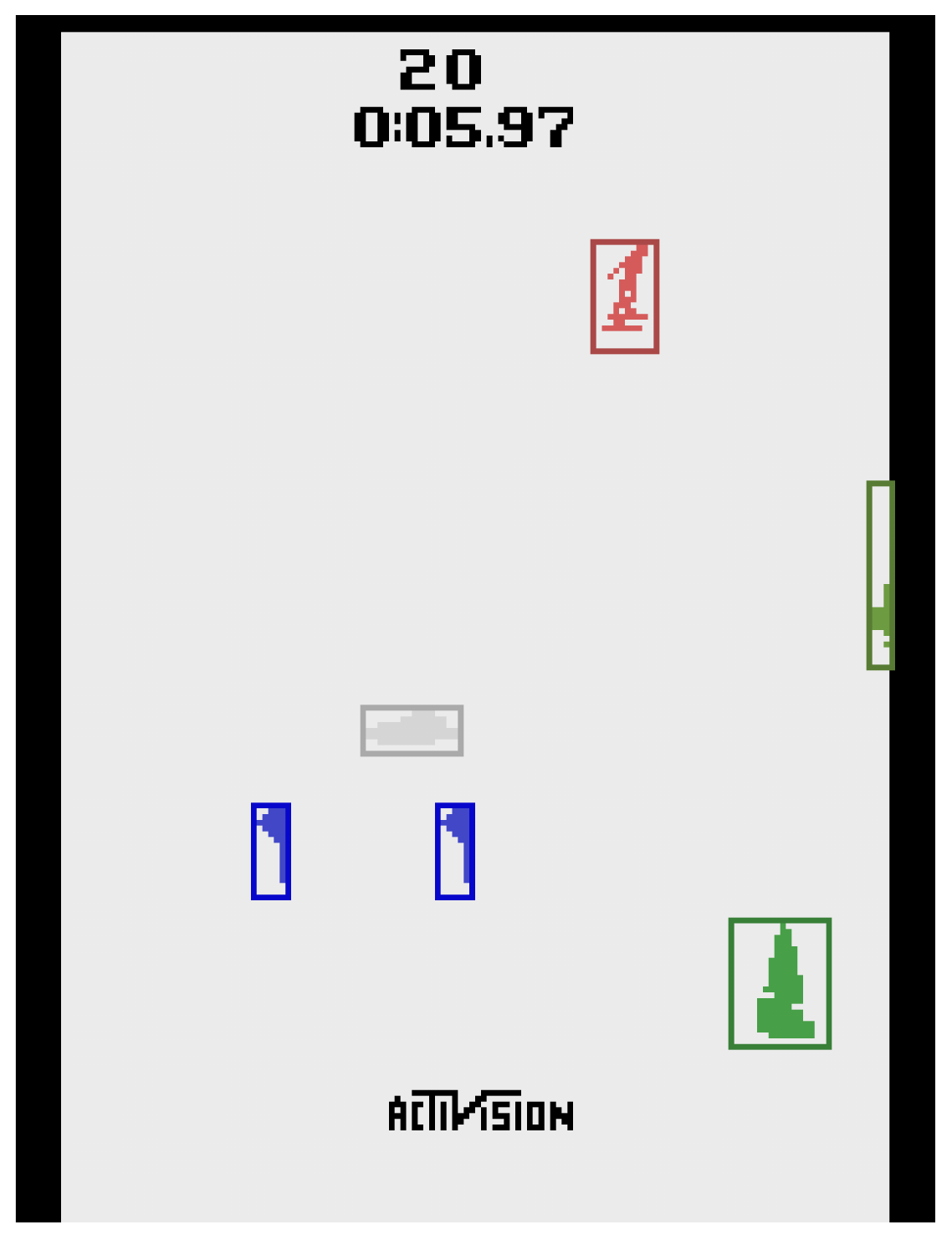}
    \end{minipage}
  }{}
  \IfFileExists{game_modifs/Skiing.tex}{%
      \begin{table}[h]
    \centering
    \begin{tabular}{p{3cm} | p{10cm}} \toprule
        \textbf{Modification} & \textbf{Effect} \\ \midrule
       invert\_flags & Switches the flag color from blue to red. \\
       moguls\_to\_trees & Replaces all moguls with trees. \\
       moving\_flags & Flags move to the left and right. \\
        \bottomrule
    \end{tabular}
\end{table}
  }{}
  
  \vspace{-1em} 

  \subsection{SpaceInvaders}
  \noindent
  \IfFileExists{environment_description/SpaceInvaders.tex}{%
    \begin{minipage}{0.80\textwidth}
     \textbf{Description:} \\
      Your objective is to destroy the space invaders by shooting your laser cannon at them before they reach the Earth. The game ends when all your lives are lost after taking enemy fire, or when they reach the earth.
    \end{minipage}
  }{}
\IfFileExists{environment_images_new_pdfs/SpaceInvaders.pdf}{%
    \begin{minipage}{0.18\textwidth}
      \raggedleft
      \includegraphics[scale=0.10]{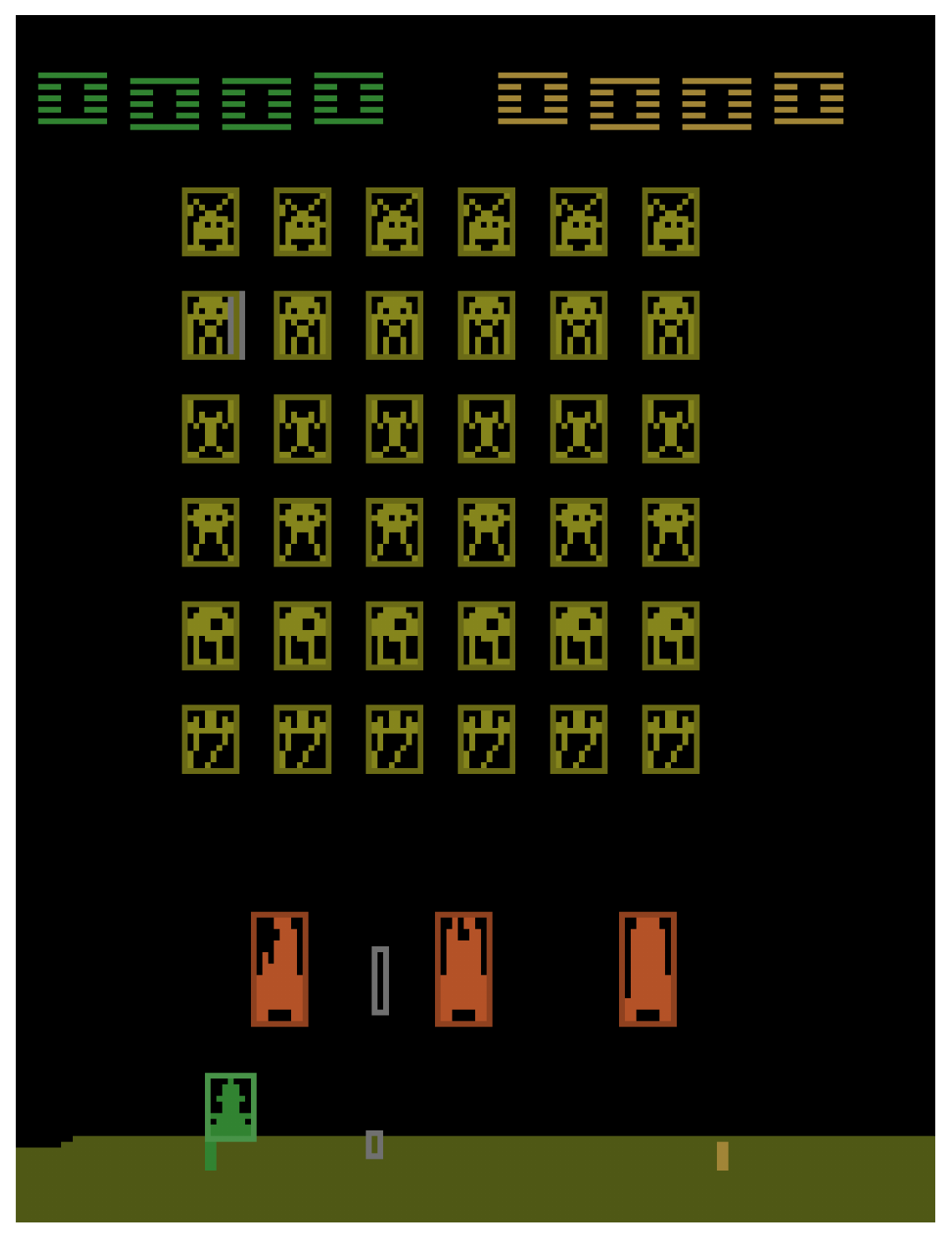}
    \end{minipage}
  }{}
  \IfFileExists{game_modifs/SpaceInvaders.tex}{%
      \begin{table}[h]
    \centering
    \begin{tabular}{p{5cm} |p{8cm}} \toprule
        \textbf{Modification} & \textbf{Effect} \\ \midrule
        disable\_shield\_left & Disables the left shield.\\
        disable\_shield\_middle & Disables the middle shield.\\
        disable\_shield\_right & Disables the right shield.\\
        disable\_shields & Disables all shields.\\
        relocate\_shields\_right & Relocates the shields to a more right position.\\
        relocate\_shields\_slight\_left & Relocates the shields to a slightly more left position.\\
        relocate\_shields\_off\_by\_one & Relocates the shields to to the right by 1 pixel.\\
        relocate\_shields\_off\_by\_three & Relocates the shields to to the right by 3 pixels.\\
        controlable\_missile & The player can control the missile by moving left and right.\\
        no\_danger & Removes disables the enemies shots and removes the shields.\\
        
        \bottomrule
    \end{tabular}
\end{table}
  }{}
  
  \vspace{-1em} 

  \subsection{StarGunner}
  \noindent
  \IfFileExists{environment_description/StarGunner.tex}{%
    \begin{minipage}{0.80\textwidth}
     \textbf{Description:} \\
      You are playing as a stargunner from the Yarthae Empire. The Empire is being invaded by the Sphyzygi. You, the stargunner, must destroy their invading ships while dodging the bombs dropped by the Sphyzygi droid. The droid cannot be destroyed.
    \end{minipage}
  }{}
\IfFileExists{environment_images_new_pdfs/StarGunner.pdf}{%
    \begin{minipage}{0.18\textwidth}
      \raggedleft
      \includegraphics[scale=0.10]{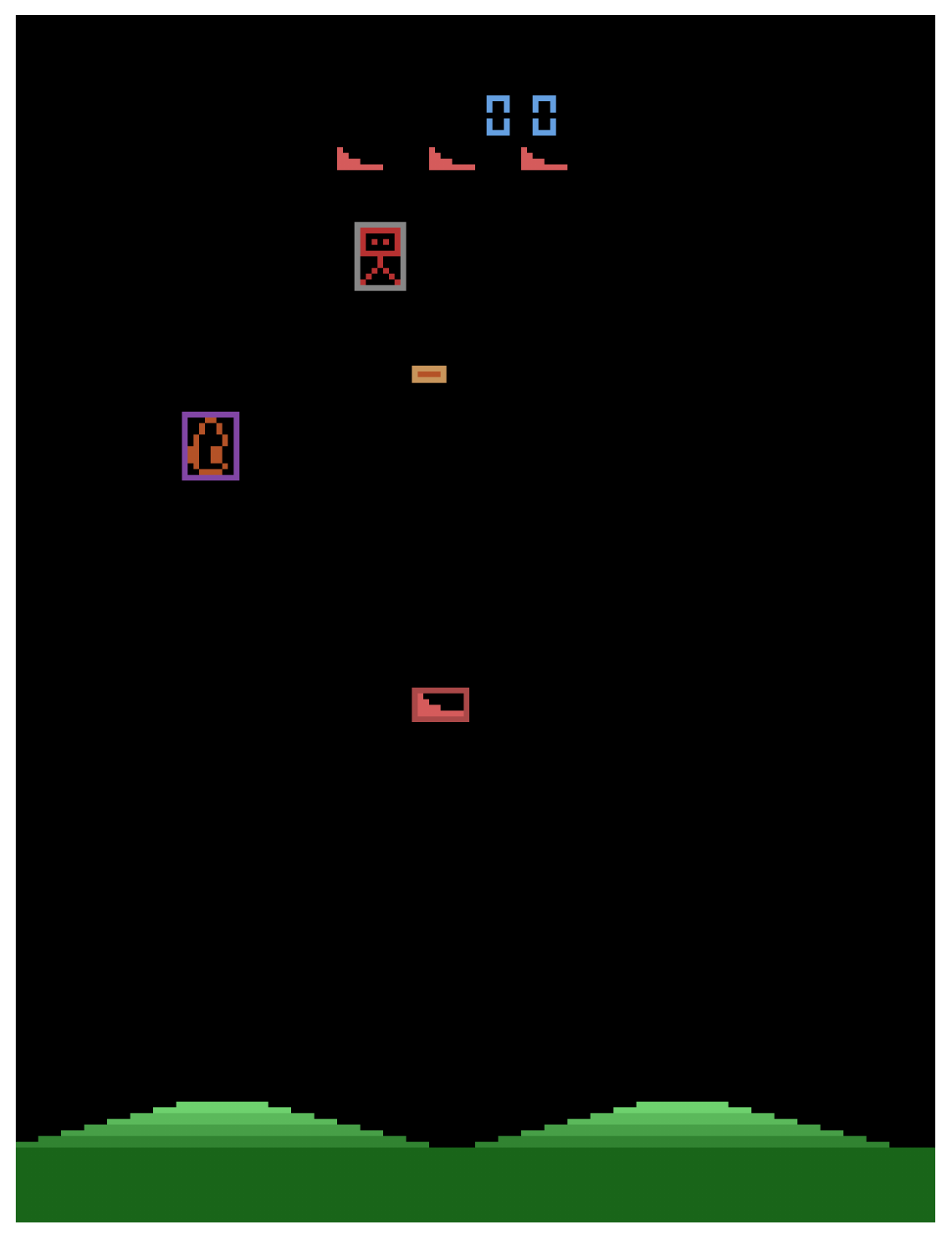}
    \end{minipage}
  }{}
  \IfFileExists{game_modifs/StarGunner.tex}{%
      \foreach \n in {StarGunner}{\input{game_modifs/\n.tex}}
  }{}
  
  \vspace{-1em} 

  \subsection{Tennis}
  \noindent
  \IfFileExists{environment_description/Tennis.tex}{%
    \begin{minipage}{0.80\textwidth}
     \textbf{Description:} \\
      You control the orange player playing against a computer-controlled blue player. The game follows the rules of tennis. The first player to win at least 6 games with a margin of at least two games wins the match. If the score is tied at 6-6, the first player to go 2 games up wins the match.
    \end{minipage}
  }{}
\IfFileExists{environment_images_new_pdfs/Tennis.pdf}{%
    \begin{minipage}{0.18\textwidth}
      \raggedleft
      \includegraphics[scale=0.10]{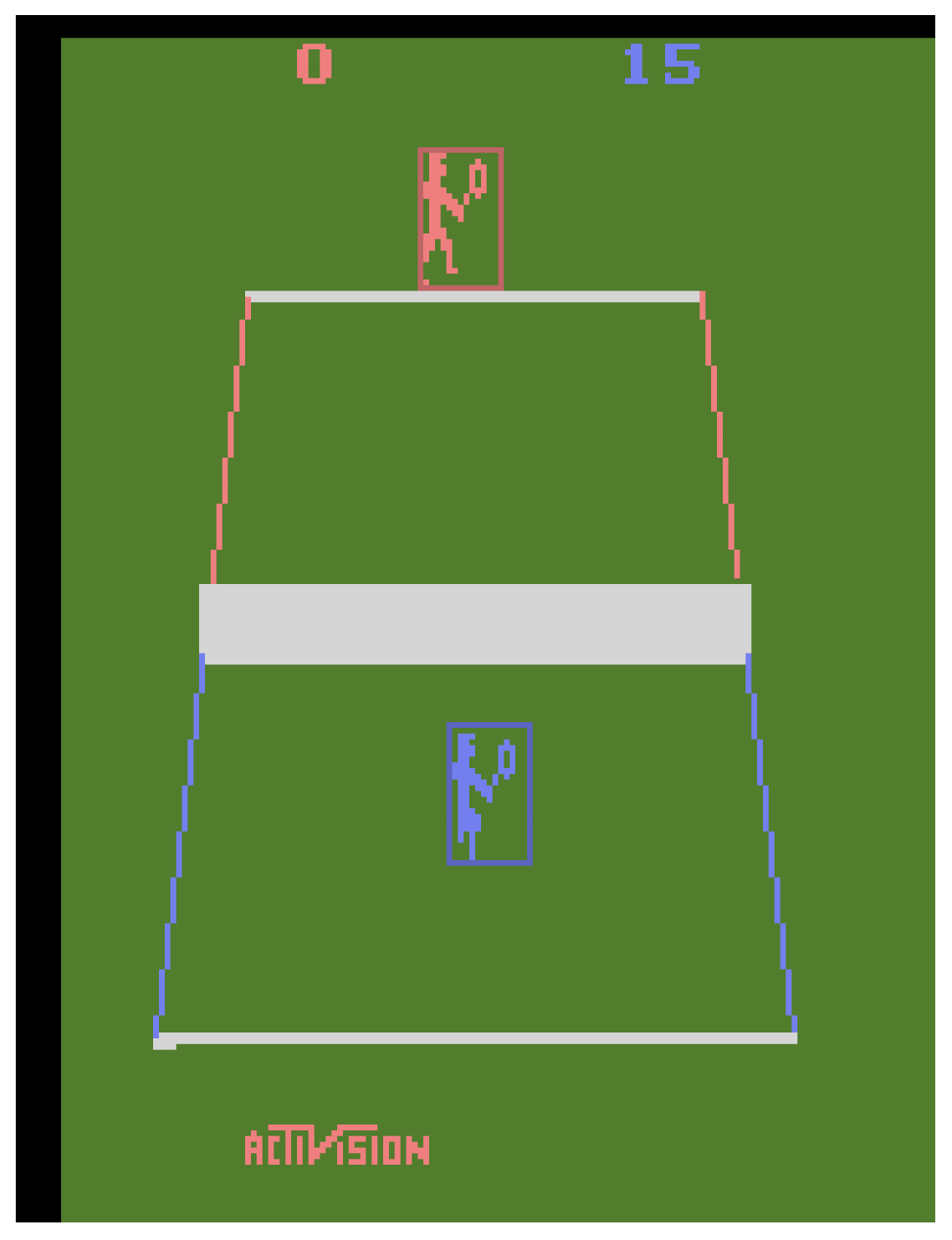}
    \end{minipage}
  }{}
  \IfFileExists{game_modifs/Tennis.tex}{%
      \begin{table}[h]
    \centering
    \begin{tabular}{p{3cm} | p{10cm}} \toprule
        \textbf{Modification} & \textbf{Effect} \\ \midrule
        wind\_effect & Sets the ball in the up and right direction by 3 pixles every single ram step to simulate the effect of wind \\
        upper\_pitches & Changes the ram so that it is always the upper persons turn to pitch. \\
        lower\_pitches & Changes the ram so that it is always the lower persons turn to pitch. \\
        upper\_player & Changes the ram so that the player is always in the upper field \\
        lower\_player & Changes the ram so that the player is always in the lower field \\
        \bottomrule
    \end{tabular}
\end{table}
  }{}
  
  \vspace{-1em} 

  \subsection{TimePilot}
  \noindent
  \IfFileExists{environment_description/TimePilot.tex}{%
    \begin{minipage}{0.80\textwidth}
     \textbf{Description:} \\
      You control an aircraft. Destroy your enemies by shooting them down. As you progress through the game, you will encounter enemies with increasingly advanced technology.
    \end{minipage}
  }{}
\IfFileExists{environment_images_new_pdfs/TimePilot.pdf}{%
    \begin{minipage}{0.18\textwidth}
      \raggedleft
      \includegraphics[scale=0.10]{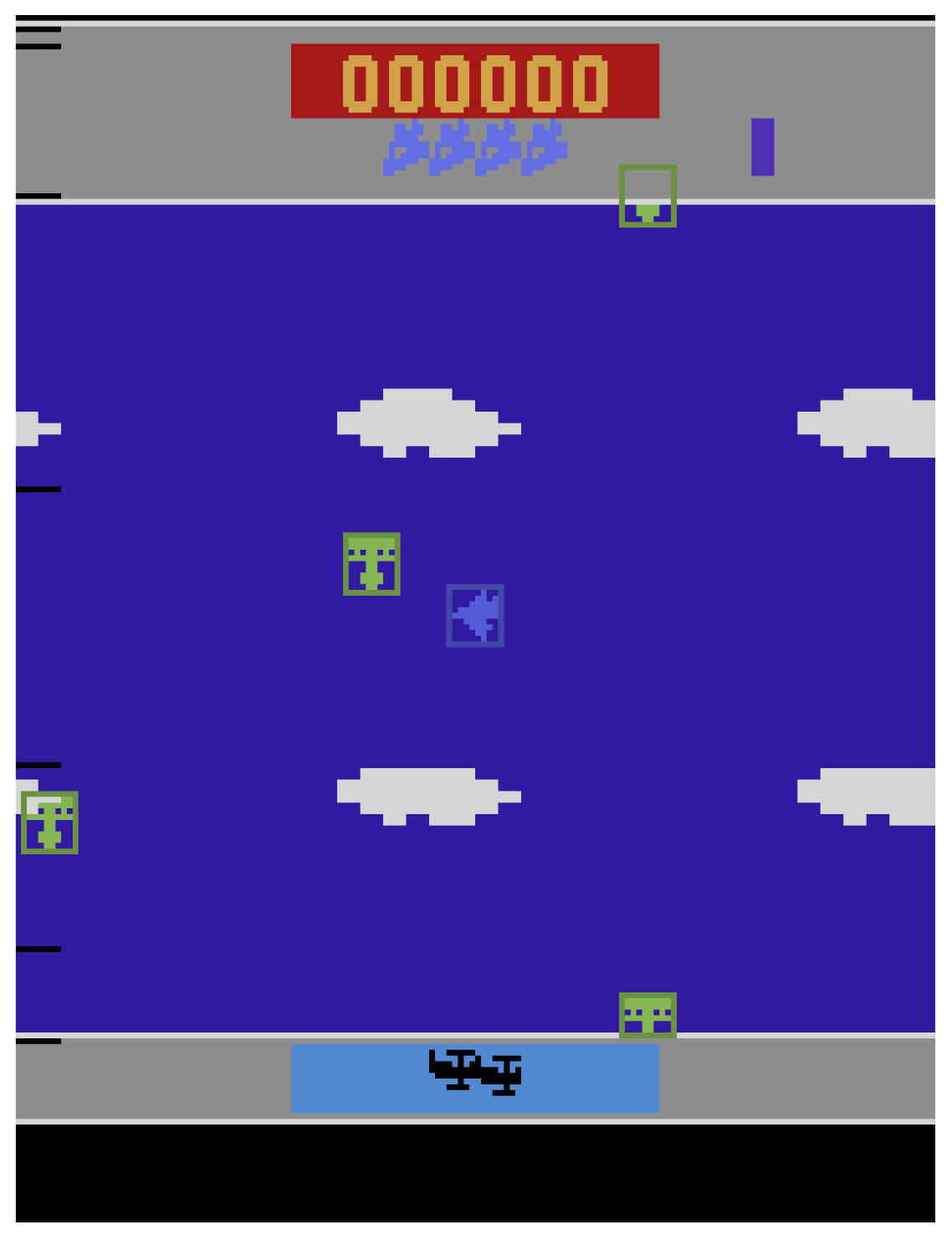}
    \end{minipage}
  }{}
  \IfFileExists{game_modifs/TimePilot.tex}{%
      \begin{table}[h]
    \centering
    \begin{tabular}{p{3cm} | p{10cm}} \toprule
        \textbf{Modification} & \textbf{Effect} \\ \midrule
        level\_$X$ & Changes the level to $X \in$ [1, 2, 3, 4, 5]. \\
        random\_orientation & Randomizes orientation of enemies. They are no longer aligned. \\ \bottomrule
    \end{tabular}
\end{table}
  }{}
  
  \vspace{-1em} 

  \subsection{Venture}
  \noindent
  \IfFileExists{environment_description/Venture.tex}{%
    \begin{minipage}{0.80\textwidth}
     \textbf{Description:} \\
      Your goal is to capture the treasure in every chamber of the dungeon while avoiding the monsters.
    \end{minipage}
  }{}
\IfFileExists{environment_images_new_pdfs/Venture.pdf}{%
    \begin{minipage}{0.18\textwidth}
      \raggedleft
      \includegraphics[scale=0.10]{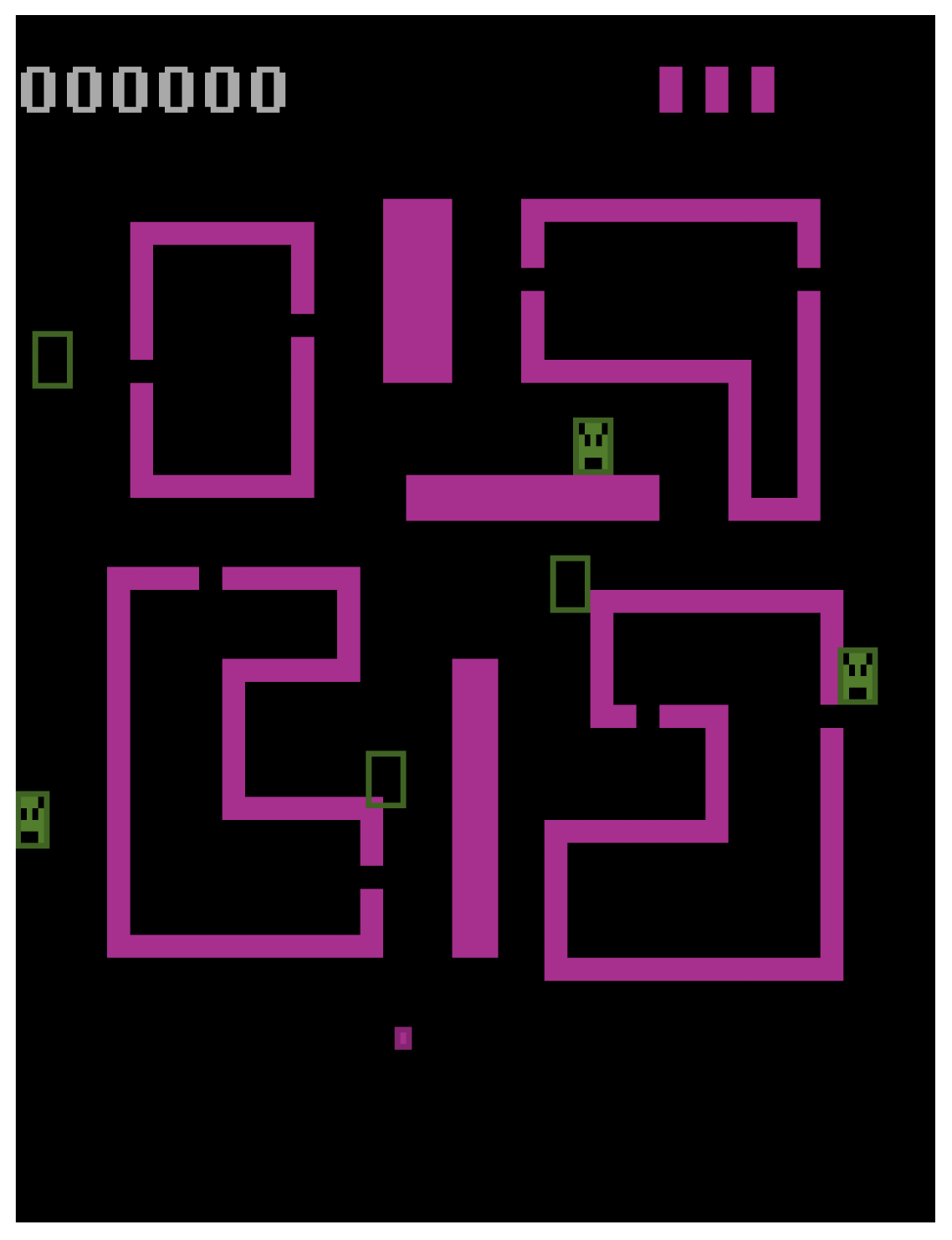}
    \end{minipage}
  }{}
  \IfFileExists{game_modifs/Venture.tex}{%
      \begin{table}[h]
    \centering
    \begin{tabular}{p{3.25cm} | p{9.75cm}} \toprule
        \textbf{Modification} & \textbf{Effect} \\ \midrule
        enemy\_color\_$X$ & Changes the color of all enemies to the color $X \in$ [black, white, red, blue, green]. \\
        random\_enemy\_colors & Changes the color of all enemies to a random color.\\
        \bottomrule
    \end{tabular}
\end{table}
  }{}
  
  \vspace{-1em} 

  \subsection{YarsRevenge}
  \noindent
  \IfFileExists{environment_description/YarsRevenge.tex}{%
    \begin{minipage}{0.80\textwidth}
     \textbf{Description:} \\
      The objective is to break a path through the shield and destroy the Qotile with a blast from the Zorlon Cannon.
    \end{minipage}
  }{}
\IfFileExists{environment_images_new_pdfs/YarsRevenge.pdf}{%
    \begin{minipage}{0.18\textwidth}
      \raggedleft
      \includegraphics[scale=0.10]{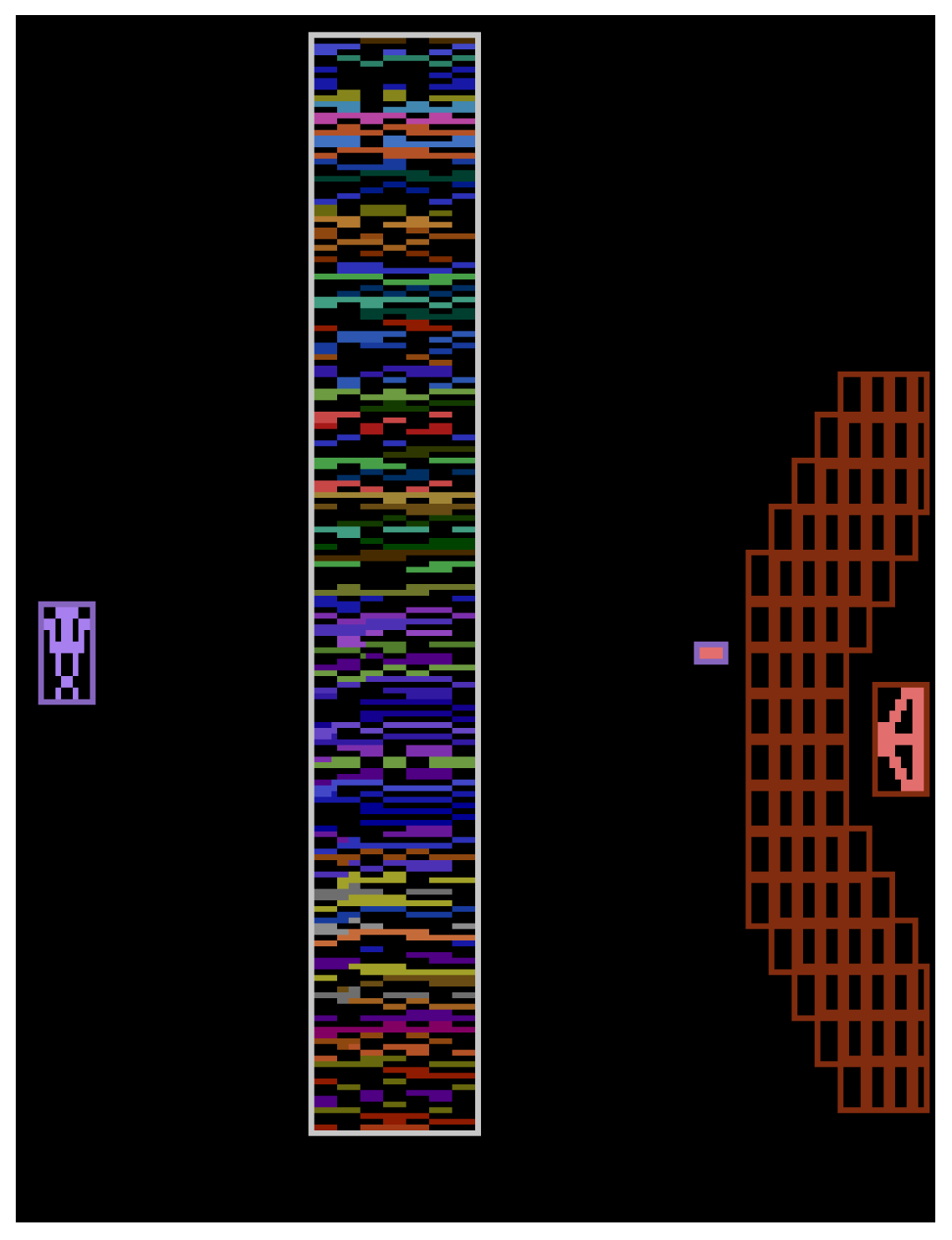}
    \end{minipage}
  }{}
  \IfFileExists{game_modifs/YarsRevenge.tex}{%
      \begin{table}[h]
    \centering
    \begin{tabular}{p{4cm} | p{10cm}} \toprule
        \textbf{Modification} & \textbf{Effect} \\ \midrule
        disable\_enemy\_movement & Disables enemy movement. \\
        disable\_block\_movement & Stops blocks in place. \\
        static & Disables enemy movement and stops blocks in place. \\
        \bottomrule
    \end{tabular}
\end{table}
  }{}
  
  \vspace{-1em} 

\end{document}